%% file: main.tex
\documentclass{article}

\usepackage{PRIMEarxiv}

\usepackage[utf8]{inputenc} 
\usepackage[T1]{fontenc}    
\usepackage{hyperref}       
\usepackage{url}            
\usepackage{booktabs}       
\usepackage{amsfonts}       
\usepackage{nicefrac}       
\usepackage{microtype}      
\usepackage{lipsum}
\usepackage{fancyhdr}       
\usepackage{graphicx}       
\graphicspath{{media/}}     

\usepackage{amsmath}
\usepackage{amssymb}
\usepackage{booktabs}
\usepackage{times}
\usepackage{epsfig}
\usepackage{enumitem}

\usepackage{wrapfig}
\usepackage{algorithm}
\usepackage{algpseudocode}
\usepackage{caption}
\usepackage{subcaption}
\usepackage{amsthm}
\usepackage{tabularx}
\usepackage{enumitem}
\usepackage{multirow}
\usepackage{wrapfig}
\usepackage{threeparttable}
\usepackage{makecell}

\usepackage[capitalize]{cleveref}
\crefname{section}{Sec.}{Secs.}
\crefname{table}{Tab.}{Tabs.}
\crefname{assumption}{Assump.}{Assumps.}
\crefname{problem}{Prob.}{Probs.}
\crefname{definition}{Defn.}{Defns.}

\Crefname{section}{Section}{Sections}
\Crefname{table}{Table}{Tables}
\Crefname{assumption}{Assumption}{Assumptions}

\newtheorem{proposition}{Proposition}
\newtheorem{definition}{Definition}
\newtheorem{lemma}{Lemma}
\newtheorem{theorem}{Theorem}
\newtheorem{assumption}{Assumption}
\newtheorem{problem}{Problem}

\newcommand{\sysname}{FEDORA}
\newcommand{\sysnameabs}{w/o $E_a$}
\newcommand{\sysnameabss}{w/o $T$}
\newcommand{\sysnameabsss}{w/o $\mathcal{L}_{fair}$}

\makeatletter
\newcommand{\multiline}[1]{%
  \begin{tabularx}{\dimexpr\linewidth-\ALG@thistlm}[t]{@{}X@{}}
    #1
  \end{tabularx}
}
\makeatother

\title{Algorithmic Fairness Generalization under Covariate and Dependence Shifts Simultaneously
\thanks{This paper is accepted by KDD 2024 research track.} 
}

\author{
    Chen Zhao\thanks{equal contribution}\\
    Baylor University\\
    Waco, Texas\\
    chen\_zhao@baylor.edu
    \And
    Kai Jiang$^\dagger$\\
    The University of Texas at Dallas \\
    Richardson, Texas\\
    kai.jiang@utdallas.edu \\
    \And
    Xintao Wu\\
    University of Arkansas \\
    Fayetteville, Arkansas\\
    xintaowu@uark.edu \\
    \And
    Haoliang Wang\\
    The University of Texas at Dallas \\
    Richardson, Texas\\
    haoliang.wang@utdallas.edu \\
    \And
    Latifur Khan \\
    The University of Texas at Dallas \\
    Richardson, Texas\\
    lkhan@utdallas.edu \\
    \And
    Christan Grant\\
    University of Florida\\
    Gainesville, Florida\\
    christan@ufl.edu\\
    \And
    Feng Chen\\
    The University of Texas at Dallas \\
    Richardson, Texas\\
    feng.chen@utdallas.edu
}

\begin{document}
\maketitle

\begin{abstract}
    \input{abstract}
\end{abstract}


\section{Introduction}
\label{sec:intro}
    \input{introduction.tex}

\section{Related Works}
\label{app:relatedworks}
    \input{app_relatedworks}

\section{Preliminaries}
\label{sec:preliminaries}
    \input{preliminaries.tex}

\section{Methodology}
\label{sec:method}
    \input{method.tex}

\section{Analysis}
\label{sec:theory}

\input{theory}

\input{algorithm}
\section{Experiments}
\label{sec:experiments}

\input{experiments}
\section{Conclusion}
\label{sec:conclusion}
    \input{conclusions.tex}


\clearpage
\bibliographystyle{unsrt}  
\bibliography{references}

\clearpage
\appendix
\begin{center}
    \textbf{{\Large Supplementary Materials}}
\end{center}

\section{Notations}
\label{app:notations}
    \input{app_notations.tex}

\section{Experimental Settings}
\label{app:expdetails}

\input{app_expdetails.tex}

\section{Ablation Studies}
\label{app:abs}
    \input{app_abs.tex}

\section{Proofs}
\label{app:proofs}

\input{app_proofs.tex}


\section{Additional Results}
\label{app:addtional-results}
    \input{app_additional_results.tex}

\section{Limitations}
\label{sec:limitations}
    \input{app_limitations}

\end{document}

%% file: abstract.tex
The endeavor to preserve the generalization of a fair and invariant classifier across domains, especially in the presence of distribution shifts, becomes a significant and intricate challenge in machine learning.
In response to this challenge, numerous effective algorithms have been developed with a focus on addressing the problem of fairness-aware domain generalization. 
These algorithms are designed to navigate various types of distribution shifts, with a particular emphasis on covariate and dependence shifts.
In this context, covariate shift pertains to changes in the marginal distribution of input features, while dependence shift involves alterations in the joint distribution of the label variable and sensitive attributes.
In this paper, we introduce a simple but effective approach that aims to learn a fair and invariant classifier by simultaneously addressing both covariate and dependence shifts across domains.
We assert the existence of an underlying transformation model can transform data from one domain to another, while preserving the semantics related to non-sensitive attributes and classes.
By augmenting various synthetic data domains through the model, we learn a fair and invariant classifier in source domains.
This classifier can then be generalized to unknown target domains, maintaining both model prediction and fairness concerns.
Extensive empirical studies on four benchmark datasets demonstrate that our approach surpasses state-of-the-art methods. 
Code repository is available at \url{https://github.com/jk-kaijiang/FDDG}.

%% file: introduction.tex
\begin{figure}[t!]
    \centering
    \begin{subfigure}[b]{0.485\linewidth}
        \includegraphics[width=\linewidth]{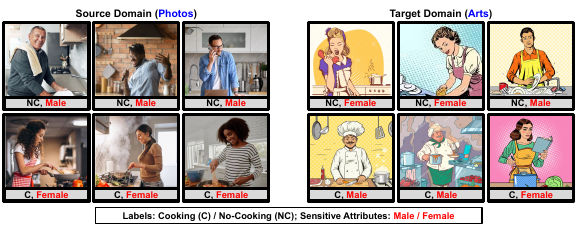}
    \end{subfigure}
    \hfill
    \begin{subfigure}[b]{0.485\linewidth}
        \includegraphics[width=\linewidth]{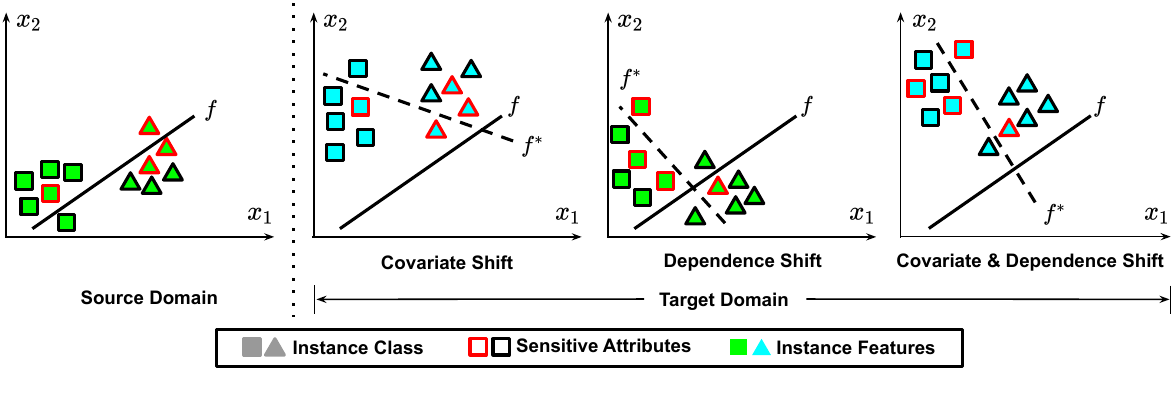}
    \end{subfigure}
    \caption{
      Illustration of the problem in generalizing fair classifiers across different data domains under covariate and dependence shifts simultaneously. 
      (Left)
      Images in source and target domains have different styles (Photos and Arts). 
      Each data domain is linked to a distinct correlation between class labels (NC and C) and sensitive attributes (Male and Female). 
      (Right) We consider $\mathbf{x}=[x_1,x_2]^T$ a simple example of a two-dimensional feature vector. A fair classifier $f$ learned using source data is applied to data sampled from various types of shifted target domains, resulting in misclassification and unfairness. $f^*$ represents the true classifier in the target domain.
      }
    \label{fig:example}
\end{figure}

While modern fairness-aware machine learning techniques have demonstrated significant success in various applications \cite{Zemel-ICML-2013, zhao-KDD-2021, Wu-2019-WWW,zhao-KDD-2022, Zhao-ICDM-2019, Zhao-ICKG-1-2020, Zhao-ICKG-2-2020,zhao2023towards, Zhao-TKDD-2014, lin2024supervised}, their primary objective is to facilitate equitable decision-making, ensuring algorithmic fairness across all demographic groups characterized by sensitive attributes, such as race and gender. 
Nevertheless, the generalization of a fair classifier learned in the source domain to a target domain during inference often demonstrates severe limitations in many state-of-the-art methods.
The poor generalization can be attributed to the data distribution shifts from source to target domains, resulting in catastrophic failures.



There are two main lines of data distribution shifts \cite{roh2023improving}: general and fairness-specific shifts. 
The former focuses on shifts involving input features and labels. 
Specifically, covariate shift \cite{shimodaira2000improving} and label shift \cite{wang2003mining} refer to variations due to different marginal distributions over feature and class variables, respectively. 
Concept shift \cite{conceptshift} indicates "functional relation change" due to the change amongst the instance-conditional distributions \cite{robey2021model}. 
Moreover, fairness-specific shifts consider additional sensitive attributes and hence place a greater emphasis on ensuring algorithmic fairness.
Demographic shift\footnote{Dependence shift is named as correlation shift in \cite{giguere2022fairness}.} \cite{giguere2022fairness} refers to certain sensitive population subgroups becoming more or less probable during inference. 
Dependence shift \cite{roh2023improving} captures the correlation change between the class variable and sensitive attributes. 
Within these distribution shifts, a trained fair classifier from source domains is directly influenced and may degrade when adapted to target domains.

To simplify, we narrow the scope of distribution shifts to two prominent ones: covariate shift, which has been extensively investigated in the context of out-of-distribution (OOD) generalization \cite{robey2021model,zhang2022towards}, and dependence shift, a topic that has gained attention in recent research.
In the illustrative example shown in \cref{fig:example}, the source and target domains exhibit variations stemming from different image styles (Photos and Arts) and correlations between labels (No-cooking and Cooking) and sensitive attributes (Male and Female). 
Specifically, in the source domain, most males in the kitchen are not cooking, whereas in the target domain, a distinct correlation is observed with most males engaging in cooking.
To learn a classifier that is both fair and accurate under such hybrid shifts, a variety of domain generalization approaches have been explored. 
Predominantly, these methods often exhibit two specific limitations: they (1) address either covariate shift \cite{robey2021model,krueger2021out,zhang2022towards} or dependence shift \cite{oh2022learning,creager2021environment}, or (2) solely focus on covariate shift but not explicitly indicate the existence of dependence shift \cite{pham2023fairness}.
Therefore, there is a need for research that explores the problem of \textit{fairness-aware domain generalization} (FDG), considering both covariate and dependence shifts simultaneously across source and target domains.

In this paper, we introduce a novel framework, namely \textit{Fair disEntangled DOmain geneRAlization} (\sysname{}). 
The key idea in our framework revolves around learning a fair and accurate classifier that can generalize from given source domains to target domains, which remain unknown and inaccessible during training.
The variations in these domains result from the concurrent presence of covariate and dependence shifts.
Notice that, unlike the settings in some works involving covariate shift \cite{taskesen2020distributionally,rezaei2020fairness,lin2023pursuing}, we assert each domain possesses a distinct data style (Photos and Arts), resulting in an alternation in feature spaces.
Technically, we assert the existence of a transformation model that can disentangle input data to a semantic factor that remains invariant across domains, a style factor that characterizes covariate-related information, and a sensitive factor that captures attributes of a sensitive nature.
To enhance the generalization of the training classifier and adapt it to unknown target domains, we augment the data by generating them through the transformation model. It utilizes semantic factors associated with various style and sensitive factors sampled from their respective prior distributions.
Furthermore, we leverage this framework to systematically define the FDG problem as a semi-infinite constrained optimization problem. 
Theoretically, we apply this re-formulation to demonstrate that a tight approximation of the problem can be achieved by solving the empirical, parameterized dual for this problem. 
Moreover, we develop a novel interpretable bound focusing on fairness within a target domain, considering the domain generalization arising from both covariate and dependence shifts.
Finally, extensive experimental results on the proposed new algorithm show that our algorithm significantly outperforms state-of-the-art baselines on several benchmarks.
Our main contributions are summarized. 
\begin{itemize}[leftmargin=*]
    
    \item We introduce a fairness-aware domain generalization problem within a framework that accommodates inter-domain variations arising from covariate and dependence shifts simultaneously. We also give a brief survey by comparing the setting of related works.

    \item We reformulate the problem to a novel constrained learning problem. We further establish duality gap bounds for the empirically parameterized dual of this problem and develop a novel upper bound that specifically addresses fairness within a target domain while accounting for the domain generalization stemming from both covariate and dependence shifts.

    \item We present a novel algorithm, \sysname{}, that enforces invariance across unseen target domains by utilizing generative models derived from the observed source domains.
    
        
    \item Comprehensive experiments are conducted to verify the effectiveness of \sysname{}. We empirically show that it significantly outperforms state-of-the-art baselines on four benchmarks.
\end{itemize}


%% file: app_relatedworks.tex
\begin{table}[t]
    \centering
    \footnotesize
    \caption{Different Types of Distribution Shifts.}
    \begin{tabular}{ll}
        \toprule
         \textbf{Type of Shifts} & \textbf{Notations}, $\forall s\in\mathcal{E}_{s}$ \\
        \midrule
        Covariate Shift (Cov.) \cite{shimodaira2000improving} & $\mathbb{P}_{X}^s\neq\mathbb{P}_{X}^t$ \\
        Label Shift (Lab.) \cite{wang2003mining} & $\mathbb{P}_{Y}^s\neq\mathbb{P}_{Y}^t$ \\
        Concept Shift (Con.) \cite{conceptshift} & $\mathbb{P}_{Y|X}^s\neq\mathbb{P}_{Y|X}^t$ \\
        Demographic Shift (Dem.) \cite{giguere2022fairness} & $\mathbb{P}_{Z}^s\neq\mathbb{P}_{Z}^t$ \\
        Dependence Shift (Dep.) \cite{roh2023improving} & $\mathbb{P}_{Y|Z}^s\neq\mathbb{P}_{Y|Z}^t$ and $\mathbb{P}_{Z}^s=\mathbb{P}_{Z}^t$; or, $\mathbb{P}_{Z|Y}^s\neq\mathbb{P}_{Z|Y}^t$ and $\mathbb{P}_{Y}^s=\mathbb{P}_{Y}^t$\\
        Hybrid Shift  & Any combination of the shifts above.\\
        \bottomrule
    \end{tabular}
    \label{tab:shifts}
\end{table}

\begin{table}[t]
    \footnotesize
    \centering
    \caption{An overview of different settings of existing approaches in mitigating unfairness under distribution shifts.}
    \begin{threeparttable}
    \begin{tabular}{lcccccccccc}
        \toprule
        \multirow{2}{*}{\textbf{Refs.}}  & \multicolumn{5}{c}{\textbf{Distribution Shifts}} & \multicolumn{3}{c}{\textbf{Spaces Change}\tnote{*}, \: $\forall s\in\mathcal{E}^{s}$} & \multirow{2}{*}{$|\mathcal{E}^s|$} & \multirow{2}{*}{\makecell[c]{\textbf{Access}\\\textbf{to Target} }} \\
        \cmidrule(lr){2-6} \cmidrule(lr){7-9}
         & \textbf{Cov.} & \textbf{Lab.} & \textbf{Con.} & \textbf{Dem.} & \textbf{Dep.} & $\mathcal{X}^s\neq\mathcal{X}^t$ & $\mathcal{Y}^s\neq\mathcal{Y}^t$ & $\mathcal{Z}^s\neq\mathcal{Z}^t$ & &  \\
        \midrule
        \cite{taskesen2020distributionally,rezaei2020fairness,lin2023pursuing} & $\bullet$ & & & & & & & & 1 & No\\
        \cite{creager2020causal} & $\bullet$ & & & & & & & & M & No\\
        \cite{rezaei2021robust,du2021fair} & $\bullet$ & & & & & & & & 1 & Yes\\
        \cite{pham2023fairness} & $\bullet$ & & & & & $\bullet$ & & & M & No\\
        \cite{biswas2021ensuring} & & $\bullet$ & & & & & & & 1 & Yes\\
        \cite{iosifidis2020online,iosifidis2019fairness} & & & $\bullet$ & & & & & & 1 & Yes\\
        \cite{schumann2019transfer,giguere2022fairness} & & & & $\bullet$ & & & & $\bullet$ & 1 & Yes\\
        \cite{creager2021environment} & & & & & $\bullet$ & & & & M & No\\
        \cite{oh2022learning} & & & & & $\bullet$ & & & & 1 & No\\
        \cite{roh2023improving} & & & & & $\bullet$ & & & & 1 & Yes\\
        \cite{kallus2018residual} & $\bullet$ & & $\bullet$ & & & & & & 1 & Yes\\
        \cite{singh2021fairness} & $\bullet$ & & & $\bullet$ & & & & & 1 & Yes\\
        \cite{schrouff2022diagnosing} & $\bullet$ & $\bullet$ & & $\bullet$ & & & & & 1 & Yes\\
        \cite{han2023achieving} & $\bullet$ & $\bullet$ & &  &  & & $\bullet$ & & 1 & No\\
        \cite{chen2022fairness} & $\bullet$ & $\bullet$ & & & & $\bullet$ & & & 1 & Yes\\
        \midrule
        \sysname{} & $\bullet$ & & & & $\bullet$ & $\bullet$ & & & M & No\\
        \bottomrule
    \end{tabular}
    \begin{tablenotes}
        \item[*] $\mathcal{Y}^s\neq\mathcal{Y}^t$ and $\mathcal{Z}^s\neq\mathcal{Z}^t$ indicate the introduction of new labels and new sensitive attributes. A change in $\mathcal{X}$ denotes a shift in feature variation, such as transitioning from photo images to arts.
    \end{tablenotes}
    \end{threeparttable}
    \label{tab:relatedwork}
\end{table}

\textbf{Domain generalization.} 
Addressing the challenge of domain shift and the absence of OOD data has led to the introduction of several state-of-the-art methods in the domain generalization field \cite{vapnik1999nature,arjovsky2019invariant,zhang2022towards,robey2021model}. These methods are designed to enable deep learning models to possess intrinsic generalizability, allowing them to adapt effectively from one or multiple source domains to target domains characterized by unknown distributions \cite{volpi2021continual}.
They encompass various techniques, such as aligning source domain distributions to facilitate domain-invariant representation learning \cite{li2018domain}, subjecting the model to domain shift during training through meta-learning \cite{li2018learning}, and augmenting data with domain analysis, among others \cite{zhou2020learning}, and so on.
In the context of the number of source domains, a significant portion of research \cite{zhang2022towards,robey2021model,blanchard2011generalizing} has focused on the multi-source setting. This setting assumes the availability of multiple distinct but relevant domains for the generalization task.
As mentioned in \cite{blanchard2011generalizing}, the primary motivation for studying domain generalization is to harness data from multiple sources in order to unveil stable patterns. This entails learning representations invariant to the marginal distributions of data features, all while lacking access to the target data.
Nevertheless, existing domain generalization methods tend to overlook the aspect of learning with fairness, where group fairness dependence patterns may not change domains.

\textbf{Fairness learning for changing environments.} 
Two primary research directions aim to tackle fairness-aware machine learning in dynamic or changing environments.
The first approach involves equality-aware monitoring methods \cite{kirton2019unions,alonso2021ordering,pham2023fairness,rezaei2021robust,singh2021fairness,giguere2022fairness,chen2022fairness}, which strive to identify and mitigate unfairness in a model's behavior by continuously monitoring its predictions. These methods adapt the model's parameters or structure when unfairness is detected. However, a significant limitation of such approaches is their assumption of invariant fairness levels across domains, which may not hold in real-world applications.
The second approach \cite{oh2022learning,creager2021environment} focuses on assessing a model's fairness in a dynamic environment exclusively under dependence shifts. However, it does not consider other types of distribution shifts.

In response to these limitations, this paper adopts a novel approach by attributing the distribution shift from source to target domains to both covariate shift and fairness dependence shift simultaneously. 
The objective is to train a fairness-aware invariant classifier capable of effective generalization across domains, ensuring robust performance in terms of both model accuracy and the preservation of fair dependence between predicted outcomes and sensitive attributes under both shifts.


%% file: preliminaries.tex
\textbf{Notations.} 
Let $\mathcal{X}\subseteq\mathbb{R}^d$ denote a feature space, $\mathcal{Z}=\{-1,1\}$ is a sensitive space, and $\mathcal{Y}=\{0,1\}$ is a label space for classification. 
Let $\mathcal{C}\subseteq\mathbb{R}^c$, $\mathcal{A}\subseteq\mathbb{R}^a$, and $\mathcal{S}\subseteq\mathbb{R}^s$ be the semantic, sensitive and style latent spaces, respectively, induced from $\mathcal{X}$ and $\mathcal{A}$ by an underlying transformation model $T:\mathcal{X}\times\mathcal{Z}\times\mathcal{E}\rightarrow\mathcal{X}\times\mathcal{Z}$. 
We use $X,Z,Y, C,A,S$ to denote random variables that take values in $\mathcal{X},\mathcal{Z},\mathcal{Y},\mathcal{C},\mathcal{A},\mathcal{S}$ and $\mathbf{x}, z, y, \mathbf{c},\mathbf{a},\mathbf{s}$ the realizations. 
A domain $e\in\mathcal{E}$ is defined as a joint distribution $\mathbb{P}^e_{XZY}=\mathbb{P}(X^e,Z^e,Y^e):\mathcal{X}\times\mathcal{Z}\times\mathcal{Y}\rightarrow[0,1]$. A classifier $f$ in a class space $\mathcal{F}$ denotes $f\in\mathcal{F}:\mathcal{X}\rightarrow\mathcal{Y}$. 
We denote $\mathcal{E}$ and $\mathcal{E}^s\subset\mathcal{E}$ as the set of domain labels for all domains and source domains, respectively.
Superscripts in the samples denote their domain labels, while subscripts specify the indices of encoders. 
For example, $E_s(\mathbf{x}^s)$ denotes a sample $\mathbf{x}$ drawn from the $s$ domain and encoded by a style encoder $E_s$.

\textbf{Fairness notions.} 
When learning a fair classifier $f\in\mathcal{F}$ that focuses on statistical parity across different sensitive subgroups, the fairness criteria require the independence between the sensitive random variables $Z$ and the predicted model outcome $f(X)$ \cite{Dwork-2011-CoRR}. 
Addressing the issue of preventing group unfairness can be framed as the formulation of a constraint. This constraint mitigates bias by ensuring that $f(X)$ aligns with the ground truth $Y$, fostering equitable outcomes.

\begin{definition}[Group Fairness Notion \cite{Wu-2019-WWW,Lohaus-2020-ICML}]
\label{def:fairNotion}
    Given a dataset $\mathcal{D}=\{(\mathbf{x}_i,z_i,y_i)\}_{i=1}^{|\mathcal{D}|}$ sampled \textit{i.i.d.} from $\mathbb{P}_{XZY}$, a classifier $f\in\mathcal{F}:\mathcal{X}\rightarrow\mathcal{Y}$ is fair when the prediction $\hat{Y}=f(X)$ is independent of the sensitive random variable $Z$. To get rid of the indicator function and relax the exact values, a linear approximated form of the difference between sensitive subgroups is defined as 
    \begin{align}
    \label{eq:fairnotion}
        \rho(\hat{Y},Z)=\big|\mathbb{E}_{\mathbb{P}_{XZY}}g(\hat{Y},Z)\big| 
        \quad \text{where} \quad g(\hat{Y},Z) = \frac{1}{p_1(1-p_1)}\Big(\frac{Z+1}{2}-p_1\Big)\hat{Y} 
    \end{align}
    $p_1$ and $1-p_1$ are the proportion of samples in the subgroup $Z=1$ and $Z=-1$, respectively.
\end{definition}
Specifically, when $p_1=\mathbb{P}(Z=1)$ and $p_1=\mathbb{P}(Z=1, Y=1)$, the fairness notion $\rho(\hat{Y},Z)$ is defined as the difference of demographic parity and the difference of equalized opportunity, respectively \cite{Lohaus-2020-ICML}. 
In this paper, we will present the results under demographic parity (and then the expectation in \cref{eq:fairnotion} is over $XZ$), while the framework can be generalized to multi-class, multi-sensitive attributes and other fairness notions. Strictly speaking, a classifier $f$ is fair over subgroups if it satisfies $\rho(\hat{Y},Z)=0$.

\begin{figure*}[t]
    \centering
    \includegraphics[width=\linewidth]{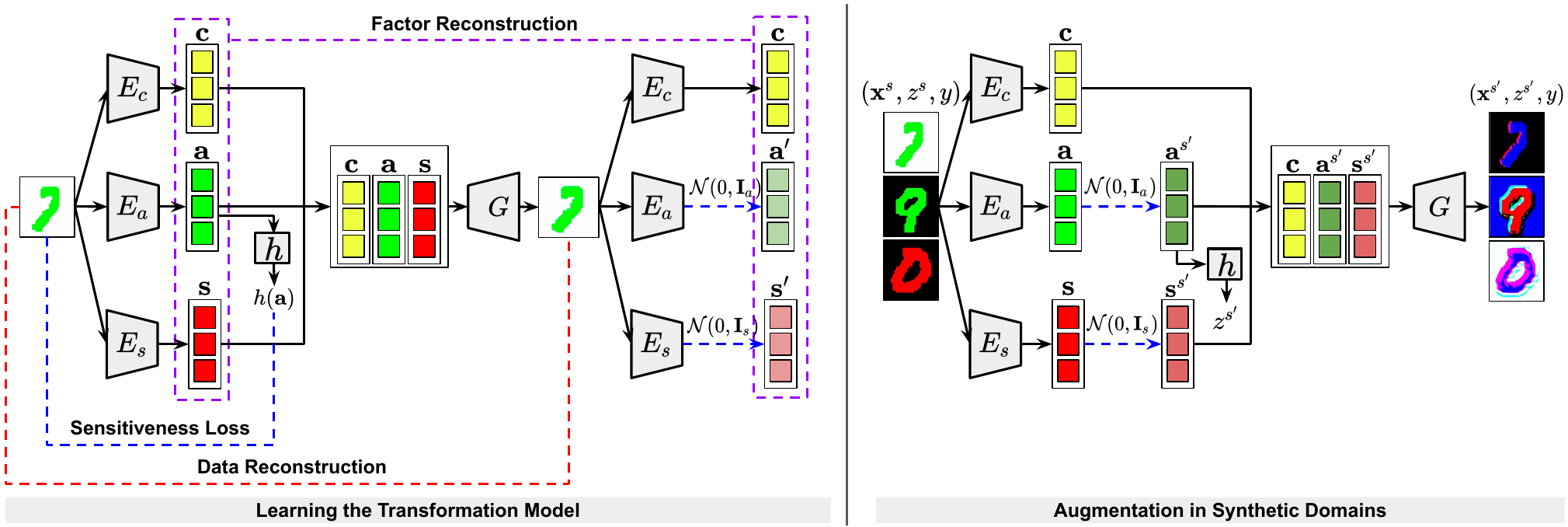}
    \caption{
    (Left) A transformation model $T$ is trained using a bi-directional reconstruction loss (data reconstruction and factor reconstruction) and a sensitiveness loss.
    (Right) To enhance the generalization of the classifier $f$ to unseen target domains, the transformation model $T$ is used for augmentation in synthetic domains by generating data based on invariant semantic factors and randomly sampled sensitive and style factors that encode synthetic domains. 
    We demonstrate the concept using the \texttt{ccMNIST} dataset, where the domains are distinguished by different digit colors and fair dependencies between class labels and sensitive attributes. Here, sensitive attributes are defined by image background colors. 
    }
    \label{fig:framework}
\end{figure*}

\textbf{Problem setting.}
Given a dataset $\mathcal{D}=\{\mathcal{D}^e\}_{e=1}^{|\mathcal{E}|}$, where each $\mathcal{D}^{e}=\{(\mathbf{x}^{e}_i, z^{e}_i, y^{e}_i)\}_{i=1}^{|\mathcal{D}^{e}|}$ is \textit{i.i.d.} sampled from a domain $\mathbb{P}^e_{XZY}$ and $e\in\mathcal{E}$, we consider multiple source domains $\{\mathbb{P}^s_{XZY}\}_{s=1}^{|\mathcal{E}^s|}$ and a distinct target domain $\mathbb{P}^t_{XZY}, t\neq s,\forall s\in\mathcal{E}^s\subset\mathcal{E}$ and $t\in\mathcal{E}\backslash\mathcal{E}^s$, which is unknown and inaccessible during training.
Given samples $\{\mathcal{D}^s\}_{s=1}^{|\mathcal{E}^s|}$ from finite source domains, the goal of fairness-aware domain generalization problems is to learn a classifier $f\in\mathcal{F}$ that is generalizable across all possible domains.
\begin{problem}[Fairness-aware Domain Generalization]
\label{prob:FDG}
    Let $\{\mathbb{P}^s_{XZY}\}_{s=1}^{|\mathcal{E}^s|}$ be a finite subset of source domains and assume that, for each $s\in \mathcal{E}^{s}$, we have access to its corresponding dataset $\mathcal{D}^{s}=\{(\mathbf{x}^{s}_i, z^{s}_i, y^{s}_i)\}_{i=1}^{|\mathcal{D}^{s}|}$ sampled \textit{i.i.d} from $\mathbb{P}^s_{XZY}$. Given a classifier set $\mathcal{F}$ and a loss function $\ell:\mathcal{Y\times Y}\rightarrow\mathbb{R}$, the goal is to learn a fair classifier $f\in\mathcal{F}$ for any $\mathcal{D}^{s}$ that minimizes the worst-case risk over all domains in $\{\mathbb{P}^e_{XZY}\}_{e=1}^{|\mathcal{E}|}$ satisfying a group fairness constraint:
    \begin{align}
    \label{eq:FDG}
        \min_{f\in\mathcal{F}}\: \max_{e\in\mathcal{E}}\: \mathbb{E}_{\mathbb{P}^s_{XZY}} \ell(f(X^{s}), Y^{s}), \:\:\:
        \text{s.t.}\:\:\rho(f(X^s),Z^s)=0
    \end{align}
\end{problem}
The goal of \cref{prob:FDG} is to seek a fair classifier $f$ that generalizes from the given finite set of source domains to give a good generalization performance on all domains. Since we do not assume data from a target domain is accessible, it makes \cref{prob:FDG} challenging to solve.

Another challenge is how closely the data distributions in unknown target domains match those in the observed source domains. 
As discussed in \cref{sec:intro} and \cref{tab:shifts}, there are five different types of distribution shifts. In this paper, we narrow the scope and claim the shift between source and target domains is solely due to covariate and dependence shifts.

\begin{definition}[Covariate Shift \cite{robey2021model} and Dependence Shift\cite{roh2023improving}] 
\label{def:CDshifts}
    In \cref{prob:FDG}, covariate shift occurs when domain variation is attributed to disparities in the marginal distributions over input features $\mathbb{P}^s_X\neq\mathbb{P}^t_X,\forall s$. On the other hand, \cref{prob:FDG} exhibits a dependence shift when domain variation arises from alterations in the joint distribution between $Y$ and $Z$, denoted $\mathbb{P}^s_{YZ}\neq\mathbb{P}^t_{YZ},\forall s$ where $\mathbb{P}_{Y|Z}^s\neq\mathbb{P}_{Y|Z}^t$ and $\mathbb{P}_{Z}^s=\mathbb{P}_{Z}^t$; or $\mathbb{P}_{Z|Y}^s\neq\mathbb{P}_{Z|Y}^t$ and $\mathbb{P}_{Y}^s=\mathbb{P}_{Y}^t$.
\end{definition}

\textbf{Underlying transformation models.} 
Inspired by existing domain generalization endeavors \cite{robey2021model,zhang2022towards,huang2018multimodal}, distribution shifts can characterize generalization tasks across domains through an underlying transformation model $T$. 
The motivation behind using $T$ lies in bolstering the robustness and adaptability of the classifier $f$ across diverse domains.
By learning a transformation model, the objective is twofold: (1) to enable the model to adapt domain-invariant data representations (factors) from the input data by disentangling domain-specific variations and (2) to generate augmented data in new domains by perturbing existing samples with various variations. 
This augmentation enhances the diversity of the source data and thereby improves the ability to generalize to unseen target domains.

%% file: method.tex
\subsection{Learning the Transformation Model}


One goal of the transformation model $T = \{E,G\}$ is to disentangle an input sample from source domains into three factors in latent spaces by learning a set of encoder $E=\{E_c,E_a,E_s\}$ and a decoder $G:\mathcal{C}\times\mathcal{A}\times\mathcal{S}\rightarrow\mathcal{X}$, where $E_c:\mathcal{X}\rightarrow\mathcal{C}$, $E_a:\mathcal{X}\rightarrow\mathcal{A}$, and $E_s:\mathcal{X}\rightarrow\mathcal{S}$ represent semantic, sensitive and style encoders, respectively.

\begin{assumption}[Multiple Latent Factors]
\label{assump:multiple-latent-spaces}
    Given dataset $\mathcal{D}^{e}=\{(\mathbf{x}^{e}_i, z^{e}_i, y^{e}_i)\}_{i=1}^{|\mathcal{D}^{e}|}$ sampled \textit{i.i.d.} from $\mathbb{P}^e_{XZY}$ domain $e\in\mathcal{E}$, we assume that each instance $\mathbf{x}^e_i$ is generated from 
    (1) a latent semantic factor $\mathbf{c}\in\mathcal{C}$, where $\mathcal{C}=\{\mathbf{c}_{y=0},\mathbf{c}_{y=1}\}$;
    (2) a latent sensitive factor $\mathbf{a}\in\mathcal{A}$, where $\mathcal{A}=\{\mathbf{a}_{z=1},\mathbf{a}_{z=-1}\}$;
    and (3) a latent style factor $\mathbf{s}^{e}$, where $\mathbf{s}^{e}$ is specific to the individual domain $e$.
    We assume that the semantic and sensitive factors in $\mathcal{C}$ and $\mathcal{A}$ do not change across domains. Each domain $\mathbb{P}^e_{XZY}$ is represented by a style factor $\mathbf{s}^{e}$ and the dependence score $\rho^e=\rho(Y^e,Z^e)$\footnote{Here, $\rho$ functions equivalently as it does in \cref{eq:fairnotion}, by substituting $\hat{Y}$ to $Y$.}, denoted $e:= (\mathbf{s}^{e}, \rho^{e})$, where $\mathbf{s}^{e}$ and $\rho^e$ are unique to the domain $\mathbb{P}^e_{XZY}$. 
\end{assumption}

Note that \cref{assump:multiple-latent-spaces} is similarly related to the one made in \cite{zhang2022towards,robey2021model,huang2018multimodal,liu2017unsupervised}. In our paper, with a focus on group fairness, we expand upon the assumptions of existing works by introducing three latent factors. 
Under \cref{assump:multiple-latent-spaces}, if two instances $(\mathbf{x}^{e_i}, z^{e_i},y)$ and $(\mathbf{x}^{e_j}, z^{e_j},y)$ where $ e_i,e_j\in\mathcal{E}, i\neq j$ share the same class label, then the latter instance can be reconstructed by decoder $G$ from the former using $\mathbf{c}=E_c(\mathbf{x}^{e_i})$, $\mathbf{s}=E_s(\mathbf{x}^{e_j})$, and $\mathbf{a}=E_a(\mathbf{x}^{e_j})$ through $T$, denoted $(\mathbf{x}^{e_j},z^{e_j})=T(\mathbf{x}^{e_i},z^{e_i},e_j)$.

To enhance the effectiveness of the transformation model $T$, our overall learning loss for these encoders and decoders consists of two main components: a bidirectional reconstruction loss and a sensitiveness loss.

\textbf{Data reconstruction loss} encourages learning reconstruction in the direction of data$\rightarrow$latent$\rightarrow$data.
As for it, a data sample $\mathbf{x}^s$ from $\mathbb{P}^s_X,\forall s\in\mathcal{E}^s$ is required to be reconstructed by its encoded factors.
\begin{align*}
    \mathcal{L}_{recon}^{data} = \mathbb{E}_{\mathbf{x}^s\sim\mathbb{P}^s_X} [\lVert G(E_c(\mathbf{x}^s),E_a(\mathbf{x}^s),E_s(\mathbf{x}^s))-\mathbf{x}^s\rVert_1]
\end{align*}

\textbf{Factor reconstruction loss.} Given latent factors $\mathbf{c}$, $\mathbf{a}$, and $\mathbf{s}^s$ encoded from a sample $\mathbf{x}^s$, they are encouraged to be reconstructed through some latent factors randomly sampled from the prior Gaussian distributions.
\begin{align*}
    \mathcal{L}_{recon}^{factor} 
    = \:&\mathbb{E}_{\mathbf{c}\sim\mathbb{P}_C,\mathbf{a}\sim\mathbb{P}_A,\mathbf{s}^s\sim\mathbb{P}_S}[\lVert E_c(G(\mathbf{c,a},\mathbf{s}^s))-\mathbf{c}\rVert_1] \\
    &+ \:\mathbb{E}_{\mathbf{c}\sim\mathbb{P}_C,\mathbf{a}\sim\mathcal{N}(0,\mathbf{I}_a),\mathbf{s}^s\sim\mathbb{P}_S} [\lVert E_a(G(\mathbf{c,a},\mathbf{s}^s))-\mathbf{a}\rVert_1]\\
    &+ \:\mathbb{E}_{\mathbf{c}\sim\mathbb{P}_C,\mathbf{a}\sim\mathbb{P}_A,\mathbf{s}^s\sim\mathcal{N}(0,\mathbf{I}_s)} [\lVert E_s(G(\mathbf{c,a},\mathbf{s}^s))-\mathbf{s}\rVert_1] 
\end{align*}
where $\mathbb{P}_C$, $\mathbb{P}_A$, $\mathbb{P}_S$ are given by $E_c(\mathbf{x}^s)$, $E_a(\mathbf{x}^s)$, and $E_s(\mathbf{x}^s)$, respectively.

\textbf{Sensitiveness loss.} Since a sensitive factor is causally dependent on the sensitive attribute of data ($\mathbf{x}^s,z^s,y^s$), a simple classifier $h:\mathcal{A}\rightarrow\mathcal{Z}$ is learned, and further it is used to label the sensitive attribute in augmented data when learning $f$.
\begin{align*}
    \mathcal{L}_{sens} = CrossEntropy(z^s,h(E_a(\mathbf{x}^s)))
\end{align*}

\textbf{Total loss.} We jointly train the encoders and the decoder to optimize the transformation model $T$ with a weighted sum loss of the three terms.
\begin{align}
    \min_{E_c,E_a,E_s,G} \: \beta_1\mathcal{L}_{recon}^{data} + \beta_2\mathcal{L}_{recon}^{factor} + \beta_3\mathcal{L}_{sens}
\end{align}
where $\beta_1,\beta_2,\beta_3>0$ are hyperparameters that control the importance of each loss term. 

\subsection{Fair Disentangled Domain Generalization}
Furthermore, with a trained transformation model $T$, to learn the fairness-aware invariant classifier $f$ across domains, we make the following assumption. 

\begin{assumption}[Fairness-aware Domain Shift]
\label{assump:FDS}
We assume that inter-domain variation is characterized by covariate and dependence shifts. As a consequence, we assume that the conditional distribution $\mathbb{P}^e_{Y|XZ}$ is stable across domains, $\forall e\in\mathcal{E}$. Given a transformation model $T$, it holds that $\mathbb{P}^{e_i}_{Y|XZ} = \mathbb{P}^{e_j}_{Y|XZ} ,\: \forall e_i,e_j\in\mathcal{E}, i\neq j$, where $(X^{e_j},Z^{e_j})=T(\mathbf{X}^{e_i},Z^{e_i},e_j)$.
\end{assumption}

In \cref{assump:FDS}, the domain shift captured by $T$ would characterize the mapping from the marginal distributions $\mathbb{P}^{e_i}_X$ and $\rho(Y^{e_i},Z^{e_i})$ over $\mathcal{D}^{e_i}$ to the distribution $\mathbb{P}^{e_j}_X$ and $\rho(Y^{e_j},Z^{e_j})$ over $\mathcal{D}^{e_j}$ sampled from a different data domain $\mathbb{P}^{e_j}_{XZY}$, respectively. 
With this in mind and under \cref{assump:FDS}, we introduce a new definition of fairness-aware invariance with respect to the variation captured by $T$ and satisfying the group fair constraint introduced in \cref{def:fairNotion}.

\begin{definition}[Fairness-aware $T$-Invariance]
\label{def:T-invariance}
Given a transformation model $T$, a fairness-aware classifier $f\in\mathcal{F}$ is domain invariant if it holds for all $e_i,e_j\in\mathcal{E}$.
\begin{align}
    f(\mathbf{x}^{e_i}) = f(\mathbf{x}^{e_j}), \: \text{ and } \: \rho(f(X^{e_i}),Z^{e_i})=\rho(f(X^{e_j}),Z^{e_j})=0
\end{align}
almost surely when $(\mathbf{x}^{e_j},z^{e_j})=T(\mathbf{x}^{e_i},z^{e_i},e_j)$, $\mathbf{x}^{e_i}\sim\mathbb{P}^{e_i}_X$, $\mathbf{x}^{e_j}\sim\mathbb{P}^{e_j}_X$.
\end{definition}

\cref{def:T-invariance} is crafted to enforce invariance on the predictions generated by $f$ directly. We expect a prediction to remain consistent across various data realizations $T$ while considering group fairness.

\begin{problem}[Fair Disentanglement Domain Generalization]
\label{prob:our-problem}
Under \cref{def:T-invariance} and \cref{assump:FDS}, if we restrict $\mathcal{F}$ of \cref{prob:FDG} to the set of invariant fairness-aware classifiers, the \cref{prob:FDG} is equivalent to the following problem
\begin{align}
\label{eq:our-problem}
    P^\star \triangleq \min_{f\in\mathcal{F}} \:\: R(f)\triangleq\mathbb{E}_{\mathbb{P}^{s_i}_{XZY}} \ell(f(X^{s_i}), Y^{s_i}) \quad
    \text{s.t.}\:\: f(X^{s_i}) = f(X^{s_j}),\:\rho(f(X^{s_i}),Z^{s_i})=\rho(f(X^{s_j}),Z^{s_j})=0  
\end{align}
where $(X^{s_j},Z^{s_j})=T(X^{s_i},Z,s_j)$, $\forall s_i,s_j\in\mathcal{E}^s$, $i\neq j$.
\end{problem}

Similar to \cite{robey2021model}, \cref{prob:our-problem} is not a composite optimization problem. Moreover, acquiring domain labels is often expensive or even unattainable, primarily due to privacy concerns. Consequently, under the assumptions of disentanglement-based invariance and domain shift, \cref{prob:FDG} can be approximated to \cref{prob:our-problem} by removing the max operator over $\mathcal{E}$.


In addition, \cref{prob:our-problem} offers a new and theoretically-principled perspective on \cref{prob:FDG}, when data varies from domain to domain with respect to $T$. To optimize \cref{prob:our-problem} is challenging because 
(1) The strict equality constraints in \cref{prob:our-problem} are difficult to enforce in practice; 
(2) Enforcing constraints on deep networks is known to be a challenging problem due to non-convexity. Simply transforming them to regularization cannot guarantee satisfaction for constrained problems; 
and (3) As we have incomplete access to all domains, it limits the ability to enforce fairness-aware $T$-invariance and further makes it hard to estimate $R(f)$. 

Due to such challenges, we develop a tractable method for approximately solving \cref{prob:our-problem} with optimality guarantees. To address the first challenge, we relax constraints in \cref{prob:our-problem}
\begin{align}
\label{eq:our-problem-relax}
    P^\star(\gamma_1,\gamma_2)\triangleq \min_{f\in\mathcal{F}} R(f)\quad
    \text{s.t.}\quad
    \delta^{s_i,s_j}(f)\leq\gamma_1, \:\rho^{s_i}(f)\leq\frac{\gamma_2}{2},\:\text{and}\:\rho^{s_j}(f)\leq\frac{\gamma_2}{2} 
\end{align}
where 
\begin{align}
    \delta^{s_i,s_j}(f)&\triangleq\mathbb{E}_{\mathbb{P}^{s_i}_{XZ}}
    d\big[f(X^{s_i}),f(X^{s_j}=T(X^{s_i},Z^{s_i},s_j))\big], \label{eq:delta}\\
    \rho^{s_i}(f)&\triangleq\rho(f(X^{s_i}),Z^{s_i}),\quad \rho^{s_j}(f)\triangleq\rho(f(X^{s_j}),Z^{s_j}) \label{eq:rho}
\end{align}
and $\forall s_i,s_j\in\mathcal{E}^s$, $i\neq j$. Here, $\gamma_1,\gamma_2>0$ are constants controlling the extent of relaxation and $d[\cdot]$ is a distance metric, \textit{e.g.,} KL-divergence. When $\gamma_1=\gamma_2=0$, \cref{eq:our-problem,eq:our-problem-relax} are equivalent. 

Since it is unrealistic to have access to the full distribution and we only have access to source domains, given data sampled from $\mathcal{E}_{s}$, we consider the empirical dual problem
\begin{align}
\label{eq:empirical-dual}
    D^\star_{\xi,N,\mathcal{E}_{s}}(\gamma_1,\gamma_2) \triangleq &\max_{\lambda_1({s_i,s_j}),\lambda_2({s_i,s_j})} \min_{\boldsymbol{\theta}\in\Theta} \:\hat{R}(\boldsymbol{\theta}) \nonumber\\
    &+\frac{1}{|\mathcal{E}^s|}\sum_{{s_i,s_j}\in\mathcal{E}_{s}}\Big[\lambda_1{(s_i,s_j)}\big(\hat{\delta}^{s_i,s_j}(\boldsymbol{\theta})-\gamma_1\big) 
    + \lambda_2{(s_i,s_j)}\big(\hat{\rho}^{s_i}(\boldsymbol{\theta})+\hat{\rho}^{s_j}(\boldsymbol{\theta})-\gamma_2\big) \Big] 
\end{align}
where $\xi=\mathbb{E}_{\mathbb{P}_X}||f(\mathbf{x})-\hat{f}(\mathbf{x},\boldsymbol{\theta})||_{\infty}>0$ is a constant bounding the difference between $f$ and its parameterized counterpart $\hat{f}:\mathcal{X}\times\Theta\rightarrow\mathbb{R}$ defined in the Defn. 5.1 of \cite{robey2021model}. 
$N$ is the number of samples drawn from $\mathbb{P}_{XZY}$ and it can be empirically replaced by $\sum_{s\in\mathcal{E}^s}|\mathcal{D}^s|$.
$\lambda_1{(s_i,s_j)},\lambda_2{(s_i,s_j)}>0$ are dual variables.
$\hat{R}(\boldsymbol{\theta})$, $\hat{\delta}^{s_i,s_j}(\boldsymbol{\theta})$, $\hat{\rho}^{s_i}(\boldsymbol{\theta})$ and $\hat{\rho}^{s_j}(\boldsymbol{\theta})$ are the empirical counterparts of $R(\hat{f}(\cdot,\boldsymbol{\theta}))$, $\delta^{s_i,s_j}(\hat{f}(\cdot,\boldsymbol{\theta}))$, $\rho^{s_i}(\hat{f}(\cdot,\boldsymbol{\theta}))$ and $\rho^{s_j}(\hat{f}(\cdot,\boldsymbol{\theta}))$, respectively.

\begin{algorithm}[t]
    \small
    \caption{\sysname{}: Fair Disentangled Domain Generalization.}
    \label{alg:algor-s2}
    \begin{flushleft}
        \textbf{Require}: Encoders $E=\{E_c,E_a,E_s\}$, decoder $G$ and sensitive classifier $h$. \\
        \textbf{Initialize}: primal and dual learning rate $\eta_p, \eta_d$, empirical constant $\gamma_1, \gamma_2$.
    \end{flushleft}
    \begin{algorithmic}[1]
    \Repeat
        \For{minibatch $\mathcal{B}=\{(\mathbf{x}_i, z_i, y_i)\}_{i=1}^{m} \subset\mathcal{D}_{s}$}
            \State $\mathcal{L}_{cls}(\boldsymbol{\theta})=\frac{1}{m}\sum_{i=1}^m\ell(y_i, \hat{f}(\mathbf{x}_i,\boldsymbol{\theta}))$
            \State Initialize $\mathcal{L}_{inv}(\boldsymbol{\theta})=0$ and $\mathcal{B}'=[\:]$
            \For{each $(\mathbf{x}_i, z_i, y_i)$ in the minibatch}
                \State \multiline{%
                Generate $(\mathbf{x}_j, z_j, y_j)=T(\mathbf{x}_i, z_i, y_i)$ and add to $\mathcal{B}'$}
                \State \multiline{%
                $\mathcal{L}_{inv}(\boldsymbol{\theta}) += \frac{1}{m}d[\hat{f}(\mathbf{x}_i,\boldsymbol{\theta}), \hat{f}(\mathbf{x}_j,\boldsymbol{\theta})]$}
            \EndFor
            \State \multiline{%
            $\mathcal{L}_{fair}(\boldsymbol{\theta}) = \big|\frac{1}{m}\sum_{(\mathbf{x}_i,z_i)\in\mathcal{B}}g(\hat{f}(\mathbf{x}_i,\boldsymbol{\theta}),z_i)\big| + \big|\frac{1}{m}\sum_{(\mathbf{x}_j,z_j)\in\mathcal{B}'}g(\hat{f}(\mathbf{x}_j,\boldsymbol{\theta}),z_j)\big|$}
            \State $\mathcal{L}(\boldsymbol{\theta})=\mathcal{L}_{cls}(\boldsymbol{\theta}) + \lambda_1\cdot\mathcal{L}_{inv}(\boldsymbol{\theta}) + \lambda_2\cdot\mathcal{L}_{fair}(\boldsymbol{\theta})$
            \State $\boldsymbol{\theta}\leftarrow\text{Adam}(\mathcal{L}(\boldsymbol{\theta}),\boldsymbol{\theta}, \eta_p)$
            \State \multiline{%
            $\lambda_1\leftarrow \text{max}\{ [\lambda_1+\eta_d\cdot(\mathcal{L}_{inv}(\boldsymbol{\theta})-\gamma_1)], 0\}$, $\lambda_2\leftarrow \text{max}\{ [\lambda_2+\eta_d\cdot(\mathcal{L}_{fair}(\boldsymbol{\theta})-\gamma_2)], 0\}$
            }
        \EndFor
    \Until{convergence}
    \Procedure{$T$}{$\mathbf{x}, z, y$}
        \State $\mathbf{c},\mathbf{a},\mathbf{s}= E(\mathbf{x})$
        \State Sample $\mathbf{a}'\sim\mathcal{N}(0,I_a)$, $\mathbf{s}'\sim\mathcal{N}(0,I_s)$
        \State $\mathbf{x}'= G(\mathbf{c}, \mathbf{a}', \mathbf{s}')$, $z'= h(\mathbf{a}')$
        \State \textbf{return} $(\mathbf{x}', z', y)$
    \EndProcedure
    \end{algorithmic}
\end{algorithm}

\subsection{The \sysname{} Algorithm}
In practice, we propose a simple but effective algorithm, given in \cref{alg:algor-s2}, which is co-trained with the transformation model $T$.
The detailed training process of $T$ is provided in \cref{alg:algor-s1} of \cref{app:expdetails}.
In \cref{alg:algor-s2}, we harness the power of $T$ to address the unconstrained dual optimization problem outlined in \cref{eq:empirical-dual} through a series of primal-dual iterations.

Given a finite number of observed source domains, to enhance the generalization performance for unseen target domains, the invariant classifier $\hat{f}$ is trained by expanding the dataset with synthetic domains generated by $T$. 
These synthetic domains are created by introducing random sample style and random sensitive factors, hence a random sensitive attribute, resulting in an arbitrary fair dependence within such domains. 
As described in \cref{fig:framework}, the sensitive factor $\mathbf{a}^{s'}$ and the style factor $\mathbf{s}^{s'}$ are randomly sampled from their prior distributions $\mathcal{N}(0,\mathbf{I}_a)$ and $\mathcal{N}(0,\mathbf{I}_s)$, respectively. 
A sensitive attribute $z^{s'}$ is further predicted from $\mathbf{a}^{s'}$ through $h$.
Along with the unchanged semantic factor $\mathbf{c}$ encoded by $(\mathbf{x}^s,z^s,y)$, they are further passed through $G$ to generate $(\mathbf{x}^{s'},z^{s'},y)$ with the unchanged class labels in an augmented synthetic domain. 
Under \cref{assump:FDS,def:T-invariance}, according to \cref{eq:delta,eq:rho}, data augmented in synthetic domains are required to maintain invariance in terms of accuracy and fairness with the data in the corresponding original domains.

Specifically, in lines 15-20 of \cref{alg:algor-s2}, we describe the transformation procedure that takes an example $(\mathbf{x}, z, y)$ as INPUT and returns an augmented example $(\mathbf{x}^\prime, z^\prime, y)$ from a new synthetic domain as OUTPUT. The augmented example has the same semantic factor as the input example but has different sensitive and style factors sampled from their associated prior distributions that encode a new synthetic domain.
Lines 1-14 show the main training loop for \sysname{}. In line 6, for each example in the minibatch $\mathcal{B}$, we apply the procedure $T$ to generate an augmented example from a new synthetic domain described above. In line 7, we consider KL-divergence as the distance metric for $d[\cdot]$.
All the augmented examples are stored in the set $\mathcal{B}'$. The Lagrangian dual loss function is defined based on $\mathcal{B}$ and $\mathcal{B}'$ in line 10. The primal parameters $\boldsymbol{\theta}$ and the dual parameters $\lambda_1$ and $\lambda_2$ are updated in lines 11-12. 

%% file: theory.tex
With the approximation on the dual problem in \cref{eq:empirical-dual}, the duality gap between $P^\star$ in \cref{eq:our-problem-relax} and $D^\star_{\xi,N,\mathcal{E}_{s}}(\gamma_1,\gamma_2)$ in \cref{eq:empirical-dual} can be explicitly bounded.
\begin{theorem}[Fairness-aware Data-dependent Duality Gap]
\label{the:duality-gap-theorem}
Given $\xi>0$, assuming $\{\hat{f}(\cdot,\boldsymbol{\theta}):\boldsymbol{\theta}\in\Theta\}\subseteq\mathcal{F}$ has finite VC-dimension, with $M$ datapoints sampled from $\mathbb{P}_{XZY}$ we have 
\begin{align*}
    |P^\star-D^\star_{\xi,N,\mathcal{E}_{s}}(\boldsymbol{\gamma})| \leq L ||\boldsymbol{\gamma}||_{1} + \xi k (1+||\boldsymbol{\lambda}^\star_{p}||_{1}) + O(\sqrt{\log(M)/M})
\end{align*}
where $\boldsymbol{\gamma}=[\gamma_1,\gamma_2]^T$; $L$ is the Lipschitz constant of $P^\star(\gamma_1,\gamma_2)$; $k$ is a small universal constant defined in \cref{prop:gap2} of \cref{app:proofs}; and $\boldsymbol{\lambda}^\star_p$ is the optimal dual variable for a perturbed version of \cref{eq:our-problem-relax}.
\end{theorem}
The duality gap that arises when solving the empirical problem presented in \cref{eq:empirical-dual} is minimal when the fairness-aware $T$-invariance in \cref{def:T-invariance} margin $\boldsymbol{\gamma}$ is narrow, and the parametric space closely approximates $\mathcal{F}$.

Furthermore, we present the following theorem to establish an upper bound on fairness within an unseen target domain.
\begin{theorem}[Fairness Upper Bound of the Unseen Target Domain]
\label{the:fairUpperBound}
    In accordance with \cref{def:fairNotion,eq:rho}, for any domain $e\in\mathcal{E}$, the fairness dependence under instance distribution $\mathbb{P}^e_{XZY}$ with respect to the classifier $f\in\mathcal{F}$ is defined as:
    \begin{align*}
        \rho^e(f)=\big|\mathbb{E}_{\mathbb{P}^e_{XZ}} g(f(X^e),Z^e) \big|
    \end{align*}
    With observed source domains $\mathcal{E}_s$, the dependence at any unseen target domain $t\in\mathcal{E}\backslash\mathcal{E}_{s}$ is upper bounded. $dist[\cdot]$ is the Jensen-Shannon distance metric \cite{endres2003new}.
    \begin{align*}
        \rho^{t}(f)\leq 
        \frac{1}{|\mathcal{E}_s|}\sum_{s_i\in\mathcal{E}_s}\rho^{s_i}(f) 
        +\sqrt{2}\min_{s_i\in\mathcal{E}_s} dist\big[\mathbb{P}^t_{XZY},\mathbb{P}^{s_i}_{XZY} \big] 
        +\sqrt{2}\max_{s_i,s_j\in\mathcal{E}_s} dist\big[\mathbb{P}^{s_i}_{XZY},\mathbb{P}^{s_j}_{XZY}\big]
    \end{align*}
    where $dist[\mathbb{P}_1,\mathbb{P}_2]=\sqrt{\frac{1}{2}KL(\mathbb{P}_1||\frac{\mathbb{P}_1+\mathbb{P}_2}{2})+\frac{1}{2}KL(\mathbb{P}_2||\frac{\mathbb{P}_1+\mathbb{P}_2}{2})}$ is JS divergence defined based on KL divergence.
\end{theorem}

Notice that the second term in \cref{the:fairUpperBound} becomes uncontrollable during training as it relies on the unseen target domain.
Therefore, to preserve fairness across target domains, we aim to learn semantic factors that map the transformation mode $T$, ensuring that $\mathbb{P}^{s_i}_{C|XZY},\forall s_i\in\mathcal{E}_s$ remains invariant across source domains. 
Simultaneously, we strive for the classifier $f$ to achieve high fairness within the source domains.
Proofs of \cref{the:fairUpperBound,the:duality-gap-theorem} are provided in \cref{app:proofs}.

%% file: experiments.tex
\begin{table}[!t]
\footnotesize
    \centering
    \caption{Statistics summary of all datasets.}
    \begin{tabular}{lllll}
        \toprule
        \textbf{Datasets} & \textbf{Domains} & \textbf{Sensitive Attr.} & \textbf{Labels} & ($s$, $\rho^s$), $\forall s \in\mathcal{E}_s$\\
        \midrule
        \texttt{ccMNIST} & digit color & background color & digit label & (R, 0.11), (G, 0.43), (B, 0.87)\\
        \midrule
        \multirow{2}{*}{\texttt{FairFace}} & \multirow{2}{*}{race} & \multirow{2}{*}{gender} & \multirow{2}{*}{age} & (B, 0.91), (E, 0.87), (I, 0.58),  \\
         & & & & (L, 0.48), (M, 0.87), (S, 0.39), (W, 0.49)\\
        \midrule
        \texttt{YFCC100M-FDG} & year & location & in-,outdoor & ($d_0$, 0.73), ($d_1$, 0.84), ($d_2$, 0.72) \\
        \midrule
        \multirow{2}{*}{\texttt{NYSF}} & \multirow{2}{*}{city} & \multirow{2}{*}{race} & \multirow{2}{*}{stop record} & (R, 0.93), (B, 0.85), (M, 0.81),  \\
         & & & & (Q, 0.98), (S, 0.88) \\
        \bottomrule
    \end{tabular}
    \label{tab:proportions}
\end{table}

\subsection{Settings}

\textbf{Datasets.} 
We evaluate the performance of our \sysname{} on four benchmarks. To highlight each source data and its fair dependence score $\rho^s$ defined in \cref{assump:multiple-latent-spaces}, we summarize the statistics in \cref{tab:proportions}. 

\textbf{(1)} \texttt{ccMNIST} is a domain generalization benchmark created by colorizing digits and the backgrounds of the \texttt{MNIST} dataset \cite{lecun1998gradient}. \texttt{ccMNIST} consists of images of handwritten digits from 0 to 9. Similar to \texttt{ColoredMNIST} \cite{arjovsky2019invariant}, for binary classification, digits are labeled with 0 and 1 for digits from 0-4 and 5-9, respectively. \texttt{ccMNIST} contains 70,000 images divided into three data domains, each characterized by a different digit color (\textit{i.e.,} red, green, blue) and followed by a different correlation between the class label and sensitive attribute (digit background colors).
\textbf{(2)} \texttt{FairFace} \cite{Karkkainen_2021_WACV} is a dataset that contains a balanced representation of different racial groups. It includes 108,501 images from seven racial categories: Black (B), East Asian (E), Indian (I), Latino (L), Middle Eastern (M), Southeast Asian (S), and White (W). In our experiments, we set each racial group as a domain, gender as the sensitive attributes, and age ($\geq$ or $<$ 50) as the class label. 
\textbf{(3)} \texttt{YFCC100M-FDG} is an image dataset created by \textit{Yahoo Labs} and released to the public in 2014. It is randomly selected from the \texttt{YFCC100M} \cite{thomee2016yfcc100m} dataset with a total of 90,000 images. For domain variations, \texttt{YFCC100M-FDG} is divided into three domains. Each contains 30,000 images from different year ranges, before 1999 ($d_0$), 2000 to 2009 ($d_1$), and 2010 to 2014 ($d_2$). The outdoor or indoor tag is used as the binary class label for each image. Latitude and longitude coordinates, representing where images were taken, are translated into different continents. The North American or non-North American continent is the sensitive attribute (related to spatial disparity).
\textbf{(4)} \texttt{NYSF} \cite{Koh-icml-2021} is a real-world dataset on policing in New York City in 2011. It documents whether a pedestrian who was stopped on suspicion of weapon possession would, in fact, possess a weapon. \texttt{NYSF} consists of records collected in five different regions: Manhattan (M), Brooklyn (B), Queens (Q), Bronx (R), and Staten (S). We use regions as different domains. This data had a pronounced racial bias against African Americans, so we consider race (black or non-black) as the sensitive attribute.


\textbf{Baselines.}
We compare the performance of \sysname{} with 19 baseline methods that fall into two main categories:
\textbf{(1)} 12 state-of-the-art domain generalizations methods, specifically designed to address covariate shifts: ColorJitter, ERM \cite{vapnik1999nature}, IRM \cite{arjovsky2019invariant}, GDRO \cite{sagawa2019distributionally}, Mixup \cite{yan2020improve}, MLDG \cite{li2018learning}, CORAL \cite{sun2016deep}, MMD \cite{li2018domain}, DANN \cite{ganin2016domain}, CDANN \cite{li2018deep}, DDG \cite{zhang2022towards}, and MBDG \cite{robey2021model}, where ColorJitter is a naive function in \textit{PyTorch} that randomly changes the brightness, contrast, saturation and hue of images; and 
\textbf{(2)} 7 state-of-the-art fairness-aware domain generalizations methods, specifically designed to address either covariate or dependence shifts: DDG-FC, MBDG-FC, EIIL \cite{creager2021environment}, FarconVAE \cite{oh2022learning}, FCR \cite{an2022transferring}, FTCS \cite{roh2023improving}, and FATDM \cite{pham2023fairness}, where DDG-FC and MBDG-FC are two baselines that built upon DDG \cite{zhang2022towards} and MBDG \cite{robey2021model}, respectively by straightforwardly adding fairness constraints defined in \cref{def:fairNotion} to the loss functions of the original models.
Due to space limits, we present results for 14 baselines in the main paper. Comprehensive results for all baselines can be found in \cref{app:addtional-results}.


\textbf{Evaluation metrics.}
Three metrics are used for evaluation. Two of them are for fairness quantification, Demographic Parity ($DP$) \cite{Dwork-2011-CoRR} and the Area Under the ROC Curve ($AUC_{fair}$) between predictions of sensitive subgroups \cite{ling2003auc}. 
Notice that the $AUC_{fair}$ is not the same as the one commonly used in classification based on TPR and FPR. The intuition behind this $AUC_{fair}$ is based on the nonparametric \textit{Mann-Whitney U} test, in which a fair condition is defined as the classifier's prediction probability of a randomly selected sample $\mathbf{x}_{-1}$ from one sensitive subgroup being greater than a randomly selected sample $\mathbf{x}_1$ from the other sensitive subgroup is equal to the probability of $\mathbf{x}_1$ being greater than $\mathbf{x}_{-1}$ \cite{Zhao-ICDM-2019,Calders-ICDM-2013}.
A value of $DP$ closer to $1$ indicates fairness, and $0.5$ of $AUC_{fair}$ represents zero bias effect on predictions.


\textbf{Model selection.} 
The model selection in domain generalization is intrinsically a learning problem, followed by \cite{robey2021model}, we use leave-one-domain-out validation criteria, which is one of the three selection methods stated in \cite{gulrajani2020search}. Specifically, we evaluate \sysname{} on the held-out source domain and average the performance of $|\mathcal{E}_{s}|-1$ domains over the held-out one. 

We defer a detailed description of the experimental settings (including evaluation metrics, architectures, and hyperparameter search) in \cref{app:expdetails} and additional results in \cref{app:addtional-results}.

\begin{figure*}[t]
    \begin{minipage}{0.675\textwidth}
        \centering
        \includegraphics[width=\textwidth]{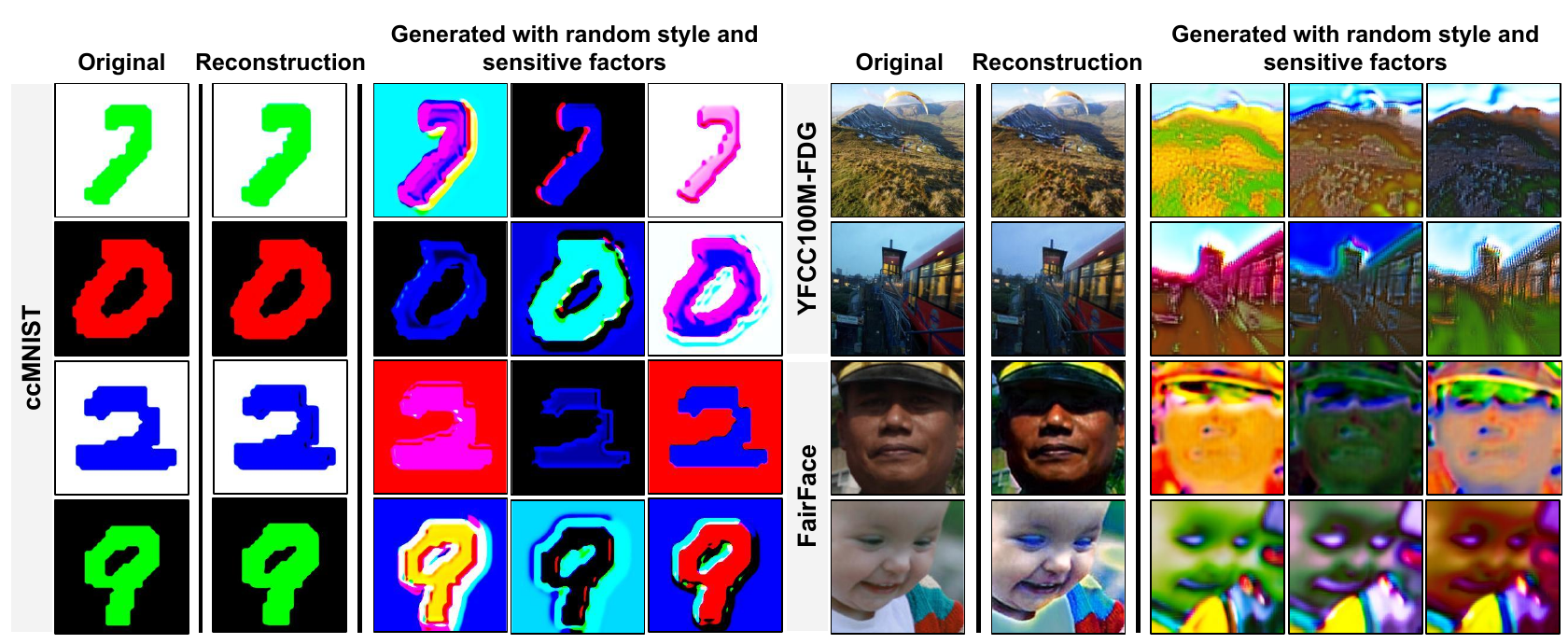}
        \caption{Visualizations for images under reconstruction and the transformation model $T$ with random style and sensitive factors.}
        \label{fig:T-vis}
    \end{minipage}
    \hfill
    \begin{minipage}{0.3\textwidth}
        \centering
        \centering
        \includegraphics[width=\textwidth]{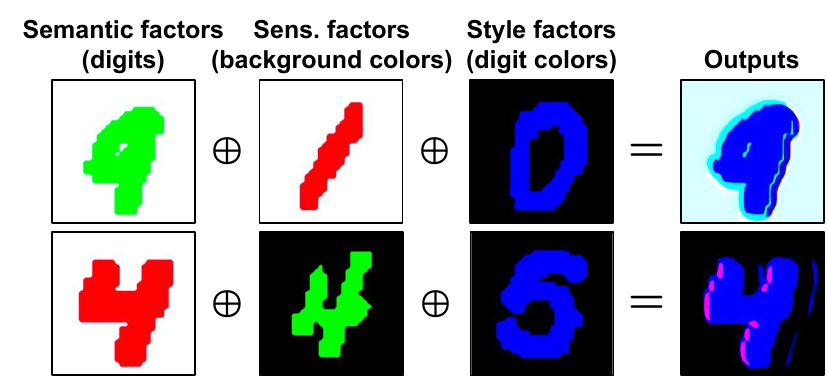}
        \caption{Example results of generating images using latent factors encoded from three images. }
        \label{fig:T-vis0}
    \end{minipage}
\end{figure*}

\begin{figure}[t!]
    \centering
    \begin{minipage}{0.485\linewidth}
        \includegraphics[width=.8\linewidth]{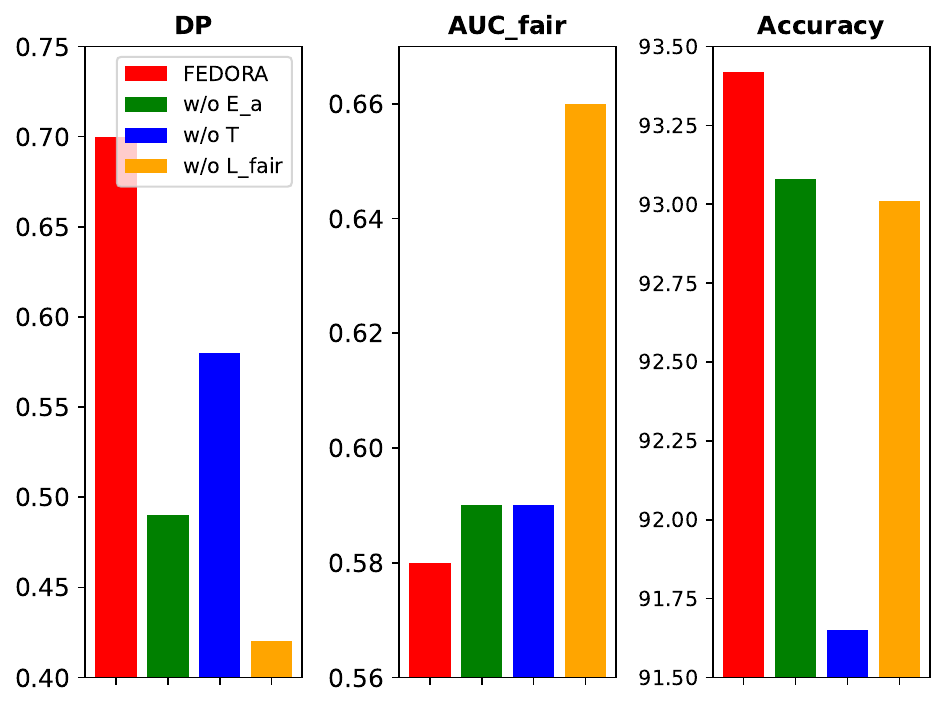}
        \caption{Ablation study on \texttt{FairFace}. Averaged results are plotted across all domains.}
    \label{fig:abs-fairface}
    \end{minipage}
    \hfill
    \begin{minipage}{0.485\linewidth}
        \centering
        \begin{subfigure}{0.5\linewidth}
            \includegraphics[width=\linewidth]{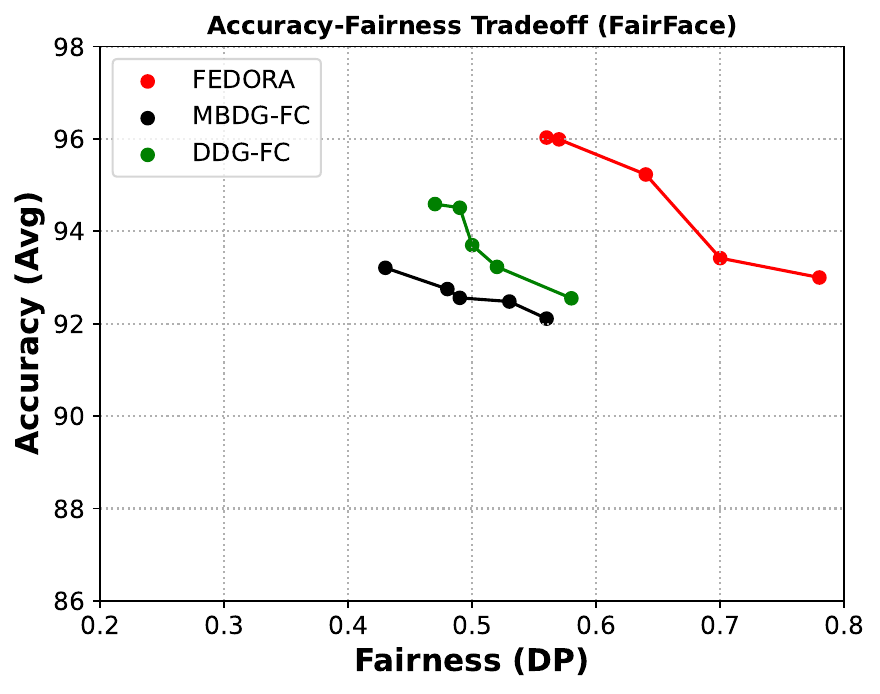}
        \end{subfigure}%
        \begin{subfigure}{0.5\linewidth}
            \includegraphics[width=\linewidth]{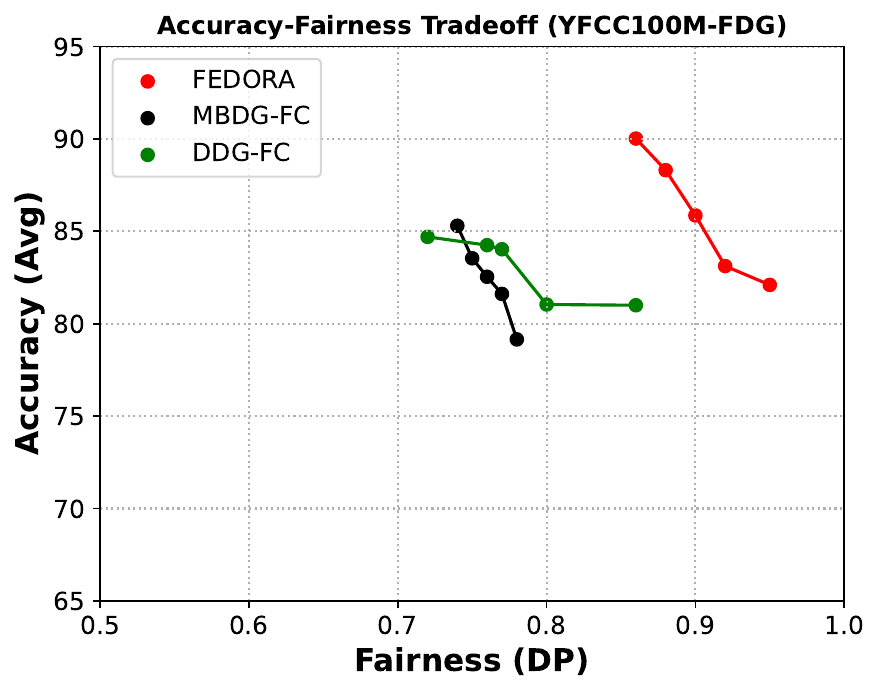}
        \end{subfigure}
          \caption{Results of accuracy-fairness tradeoff on \texttt{Fairface} (left) and \texttt{YFCC100M-FDG} (right) sweeping over a range of $\lambda_2$. }
          \label{fig:tradeoff}
    \end{minipage}
\end{figure}

\begin{table}[!t]
\scriptsize
    \centering
    \setlength\tabcolsep{4.8pt}
    \caption{Performance on \texttt{FairFace}. (bold is the best; underline is the second best).}
    \label{tab:result-fairface}
    \begin{tabular}{l|c|c|c |c}
        \toprule
         & \multicolumn{4}{c}{$DP$ $\uparrow$ / $AUC_{fair}$ $\downarrow$ / \textbf{Accuracy} $\uparrow$} \\ 
        \cmidrule(lr){2-5}
        \multirow{-2}{*}{\textbf{Methods}} & (B, 0.91)  & (W, 0.49) & (L, 0.48) & \textbf{Avg}\\
        \cmidrule(lr){1-1} \cmidrule(lr){2-2} \cmidrule(lr){3-3} \cmidrule(lr){4-4} \cmidrule(lr){5-5} 
        ColorJitter & 0.64$\pm$0.26 / 0.64$\pm$0.15 / \underline{93.47}$\pm$1.56  
        & 0.34$\pm$0.09 / 0.64$\pm$0.02 / 92.07$\pm$0.55 
        & 0.39$\pm$0.10 / 0.70$\pm$0.02 / 91.77$\pm$0.61 & 0.42 / 0.66 / 92.94
        \\
        ERM & 0.67$\pm$0.17 / 0.58$\pm$0.02 / 91.89$\pm$1.10 
        & 0.39$\pm$0.09 / 0.61$\pm$0.01 / 92.82$\pm$0.38
        & \underline{0.57}$\pm$0.15 / 0.62$\pm$0.01 / 91.96$\pm$0.51 & 0.51 / 0.61 / 93.08
        \\
        IRM & 0.63$\pm$0.12 / 0.58$\pm$0.01 / 93.39$\pm$1.03 
        & 0.32$\pm$0.19 / 0.66$\pm$0.01 / 90.54$\pm$1.56
        & 0.41$\pm$.021 / 0.63$\pm$0.05 / 92.06$\pm$1.89 & 0.43 / 0.62 / 92.48
        \\
        GDRO & 0.71$\pm$0.16 / 0.57$\pm$0.02 / 89.81$\pm$1.10 
        & 0.48$\pm$0.09 / \underline{0.60}$\pm$0.01 / 92.50$\pm$0.38
        & 0.54$\pm$0.15 / 0.62$\pm$0.01 / 91.59$\pm$0.51 & 0.55 / \underline{0.60} / 92.55
        \\
        Mixup & 0.58$\pm$0.19 / 0.59$\pm$0.02 / 92.46$\pm$0.69 
        & 0.43$\pm$0.19 / 0.61$\pm$0.01 / \underline{92.98}$\pm$0.03
        & 0.55$\pm$0.22 / 0.61$\pm$0.02 / 93.43$\pm$2.02 & 0.51 / \underline{0.60} / 93.19
        \\
        DDG & 0.60$\pm$0.20 / 0.59$\pm$0.02 / 91.76$\pm$1.03 
        & \underline{0.51}$\pm$0.07 / \underline{0.60}$\pm$0.01 / 91.34$\pm$0.80
        & 0.44$\pm$0.17 / 0.62$\pm$0.02 / 93.46$\pm$0.32 & 0.49 / 0.61 / 92.74
        \\
        MBDG & 0.60$\pm$0.15 / 0.58$\pm$0.01 / 91.29$\pm$1.41 
        & 0.30$\pm$0.04 / 0.62$\pm$0.01 / 91.05$\pm$0.53 
        & 0.56$\pm$0.09 / 0.61$\pm$0.01 / \underline{93.49}$\pm$0.97 & 0.50 / \underline{0.60} / 92.71
        \\
        \cmidrule(lr){1-5}
        DDG-FC  & 0.61$\pm$0.06 / 0.58$\pm$0.03 / 92.27$\pm$1.65 
        & 0.48$\pm$0.15 / 0.62$\pm$0.02 / 92.45$\pm$1.55
        &  0.50$\pm$0.25 / 0.62$\pm$0.03 / 92.42$\pm$0.30 & 0.52 / 0.61 / \underline{93.23}
        \\
        MBDG-FC  & 0.70$\pm$0.15 / 0.56$\pm$0.03 / 92.12$\pm$0.43 
        & 0.32$\pm$0.07 / \underline{0.60}$\pm$0.03 / 91.50$\pm$0.57
        & \underline{0.57}$\pm$0.23 / 0.62$\pm$0.02 / 91.89$\pm$0.81 & 0.53 / \underline{0.60} / 92.48
        \\
        EIIL & 0.88$\pm$0.07 / 0.59$\pm$0.05 / 84.75$\pm$2.16 
        & 0.46$\pm$0.05 / 0.65$\pm$0.03 / 86.53$\pm$1.02
        & 0.49$\pm$0.07 / \textbf{0.59}$\pm$0.01 / 88.39$\pm$1.25 & 0.64 / 0.61 / 87.78
        \\
        FarconVAE & \underline{0.93}$\pm$0.03 / \textbf{0.54}$\pm$0.01 / 89.61$\pm$0.64 
        & \underline{0.51}$\pm$0.07 / \underline{0.60}$\pm$0.01 / 86.40$\pm$0.42
        & \textbf{0.58}$\pm$0.05 / \underline{0.60}$\pm$0.05 / 88.70$\pm$0.71 & 0.66 / \textbf{0.58} / 88.46
        \\
        FCR & 0.81$\pm$0.05 / 0.59$\pm$0.02 / 79.66$\pm$0.25 
        & 0.39$\pm$0.06 / 0.63$\pm$0.02 / 82.33$\pm$0.89 & 0.38$\pm$0.12 / 0.66$\pm$0.02 / 85.22$\pm$2.33 & 0.54 / 0.63 / 83.68  \\
        
        FTCS & 0.75$\pm$0.10 / 0.60$\pm$0.02 / 80.00$\pm$0.20 
        & 0.40$\pm$0.06 / \underline{0.60}$\pm$0.02 / 79.66$\pm$1.05  & 0.42$\pm$0.23 / 0.65$\pm$0.03 / 79.64$\pm$1.00  & 0.57 / 0.64 / 80.91  \\
        
        FATDM & \underline{0.93}$\pm$0.03 / 0.57$\pm$0.02 / 92.20$\pm$0.36 
        & 0.46$\pm$0.05 / 0.63$\pm$0.01 / 92.56$\pm$0.31
        & 0.51$\pm$0.16 / 0.63$\pm$0.02 / 93.33$\pm$0.20 & \underline{0.67} / 0.61 / 92.54
        \\
        \cmidrule(lr){1-5}
        \sysname{} & \textbf{0.94}$\pm$0.05 / \underline{0.55}$\pm$0.02 / \textbf{93.91}$\pm$0.33 
        & \textbf{0.52}$\pm$0.17 / \textbf{0.58}$\pm$0.03 / \textbf{93.02}$\pm$0.50 
        & \textbf{0.58}$\pm$0.15 / \textbf{0.59}$\pm$0.01 / \textbf{93.73}$\pm$0.26 & \textbf{0.70} / \textbf{0.58} / \textbf{93.42}
        \\
        \bottomrule
    \end{tabular}
\end{table}

\begin{table}[!t]
\scriptsize
    \centering
    \setlength\tabcolsep{4.8pt}
    \caption{Performance on \texttt{YFCC100M-FDG}. (Bold is the best; \underline{underline} is the second best.)}
    \label{tab:result-YFCC100M}
    \begin{tabular}{l|c|c|c|c}
        \toprule
         & \multicolumn{4}{c}{\textbf{$DP$} $\uparrow$ / \textbf{$AUC_{fair}$} $\downarrow$ / \textbf{Accuracy} $\uparrow$} \\ 
        \cmidrule(lr){2-5}
        \multirow{-2}{*}{\textbf{Methods}} & ($d_0$, 0.73) & ($d_1$, 0.84) & ($d_2$, 0.72) & \textbf{Avg}\\
        \cmidrule(lr){1-1} \cmidrule(lr){2-2} \cmidrule(lr){3-3} \cmidrule(lr){4-4} \cmidrule(lr){5-5}
        ColorJitter & 0.67$\pm$0.06 / 0.57$\pm$0.02 / 57.47$\pm$1.20 & 0.67$\pm$0.34 / 0.61$\pm$0.01 / 82.43$\pm$1.25 & 0.65$\pm$0.21 / 0.64$\pm$0.02 / 87.88$\pm$0.35 & 0.66 / 0.61 / 75.93\\
        ERM & 0.81$\pm$0.09 / 0.58$\pm$0.01 / 40.51$\pm$0.23 & 0.71$\pm$0.18 / 0.66$\pm$0.03 / 83.91$\pm$0.33 & \underline{0.89}$\pm$0.08 / 0.59$\pm$0.01 / 82.06$\pm$0.33 & 0.80 / 0.61 / 68.83 \\
        IRM & 0.76$\pm$0.10 / 0.58$\pm$0.02 / 50.51$\pm$2.44 & 0.87$\pm$0.08 / 0.60$\pm$0.02 / 73.26$\pm$0.03 & 0.70$\pm$0.24 / 0.57$\pm$0.02 / 82.78$\pm$2.19 & 0.78 / 0.58 / 68.85 \\
        GDRO & 0.80$\pm$0.05 / 0.59$\pm$0.01 / 53.43$\pm$2.29 & 0.73$\pm$0.22 / 0.60$\pm$0.01 / 87.56$\pm$2.20 & 0.79$\pm$0.13 / 0.65$\pm$0.02 / 83.10$\pm$0.64 & 0.78 / 0.62 / 74.70 \\
        Mixup & \underline{0.82}$\pm$0.07 / 0.57$\pm$0.03 / 61.15$\pm$0.28 & 0.79$\pm$0.14 / 0.63$\pm$0.03 / 78.63$\pm$0.97 & \underline{0.89}$\pm$0.05 / 0.60$\pm$0.01 / 85.18$\pm$0.80 & \underline{0.84} / 0.60 / 74.99 \\
        DDG & 0.81$\pm$0.14 / 0.57$\pm$0.03 / 60.08$\pm$1.08 & 0.74$\pm$0.12 / 0.66$\pm$0.03 / 92.53$\pm$0.91 & 0.71$\pm$0.21 / 0.59$\pm$0.03 / \textbf{95.02}$\pm$1.92 & 0.75 / 0.61 / 82.54 \\
        MBDG & 0.79$\pm$0.15 / 0.58$\pm$0.01 / 60.46$\pm$1.90 & 0.73$\pm$0.07 / 0.67$\pm$0.01 / 94.36$\pm$0.23 & 0.71$\pm$0.11 / 0.59$\pm$0.03 / \underline{93.48}$\pm$0.65 & 0.74 / 0.61 / \underline{82.77} \\
        \cmidrule(lr){1-5}
        DDG-FC & 0.76$\pm$0.06 / 0.58$\pm$0.03 / 59.96$\pm$2.36 & 0.83$\pm$0.06 / 0.58$\pm$0.01 / \textbf{96.80}$\pm$1.28 & 0.82$\pm$0.09 / 0.59$\pm$0.01 / 86.38$\pm$2.45 & 0.80 / 0.58 / 81.04 \\
        MBDG-FC & 0.80$\pm$0.13 / 0.58$\pm$0.01 / \underline{62.31}$\pm$0.13 & 0.72$\pm$0.09 / 0.63$\pm$0.01 / \underline{94.73}$\pm$2.09 & 0.80$\pm$0.07 / \textbf{0.53}$\pm$0.01 / 87.78$\pm$2.11 & 0.77 / 0.58 / 81.61 \\
        EIIL & \textbf{0.87}$\pm$0.11 / \underline{0.55}$\pm$0.02 / 56.74$\pm$0.60 & 0.76$\pm$0.05 / \underline{0.54}$\pm$0.03 / 68.99$\pm$0.91 & 0.87$\pm$0.06 / 0.78$\pm$0.03 / 72.19$\pm$0.75 & 0.83 / 0.62 / 65.98  \\
        FarconVAE & 0.67$\pm$0.06 / 0.61$\pm$0.03 / 51.21$\pm$0.61 & \underline{0.90}$\pm$0.06 / 0.59$\pm$0.01 / 72.40$\pm$2.13 & 0.85$\pm$0.12 / \underline{0.55}$\pm$0.01 / 74.20$\pm$2.46 & 0.81 / 0.58 / 65.93 \\
        FCR & 0.62$\pm$0.21 / 0.70$\pm$0.01 / 55.32$\pm$0.04 & 0.63$\pm$0.14 / 0.66$\pm$0.10 / 70.89$\pm$0.22 & 0.66$\pm$0.30 / 0.78$\pm$0.02 / 70.58$\pm$0.23 & 0.64 / 0.71 / 65.60 \\
        FTCS & 0.72$\pm$0.03 / 0.60$\pm$0.01 / 60.21$\pm$0.10 & 0.79$\pm$0.02 / 0.59$\pm$0.01 / 79.96$\pm$0.05 & 0.69$\pm$0.10 / 0.60$\pm$0.06 / 72.99$\pm$0.50 & 0.73 / 0.60 / 71.05 \\
        FATDM & 0.80$\pm$0.10 / \underline{0.55}$\pm$0.01 / 61.56$\pm$0.89 & 0.88$\pm$0.08 / 0.56$\pm$0.01 / 90.00$\pm$0.66 & 0.86$\pm$0.10 / 0.60$\pm$0.02 / 89.12$\pm$1.30 & \underline{0.84} / \underline{0.57} / 80.22 \\
        \cmidrule(lr){1-5}
        \sysname{} & \textbf{0.87}$\pm$0.09 / \textbf{0.53}$\pm$0.01 / \textbf{62.56}$\pm$2.25 & \textbf{0.94}$\pm$0.05 / \textbf{0.52}$\pm$0.01 / 93.36$\pm$1.70 & \textbf{0.93}$\pm$0.03 / \textbf{0.53}$\pm$0.02 / 93.43$\pm$0.73 & \textbf{0.92} / \textbf{0.53} / \textbf{83.12} \\
        \bottomrule
    \end{tabular}
\end{table}

\subsection{Results}

\textbf{Data augmentation in synthetic domains via $T$.}
We visualize the augmented samples with random variations in \cref{fig:T-vis}. 
The first column (Original) shows the images sampled from the datasets. 
In the second column (Reconstruction), we display images generated from latent factors encoded from the images in the first column.
The images in the second column closely resemble those in the first column.
Images in the last three columns are generated using the semantic factors encoded from images in the first column, associated with style and sensitive factors randomly sampled from their respective Gaussian distributions.
The images in the last three columns preserve the fundamental semantic information of the corresponding samples in the first column. However, their style and sensitive attributes undergo significant changes at random. 
The generated images within synthetic domains enhance the classifier's generalization ($f$) to unseen source domains.
This demonstrates that the transformation model $T$ effectively extracts latent factors and produces diverse transformations of the provided data domains.

\textbf{Effectiveness of $T$.}
To further validate the effectiveness of $T$, drawing inspiration from \cite{huang2018multimodal}, we train a separate transformation model for each domain. 
Subsequently, we generate an output image by utilizing distinct latent factors from each domain.
Using \texttt{ccMNIST} as an example, we individually train three transformation models $\{T^i\}_{i=1}^3$ within each domain. Each $T^i$ includes unique encoders $E_c^i$, $E_a^i$, and $E_s^i$. 
As shown in \cref{fig:T-vis0}, an output image is generated through $G$ using a semantic factor (digit class, $E_c^1(\mathbf{x}^1)$), a sensitive factor (background color, $E_a^2(\mathbf{x}^2)$), and a style factor (digit color, $E_s^3(\mathbf{x}^3)$) from images in different domains.
As a result, the output image is constructed from the digit of $\mathbf{x}^1$, the background color of $\mathbf{x}^2$, and the digit color of $\mathbf{x}^3$, with given variations.
This suggests that the augmented data with random variations in \cref{fig:T-vis} for the synthetic domain are not merely altering colors; instead, they are precisely generated with unchanged semantics and random sensitive and style factors.

\textbf{The effectiveness of \sysname{} across domains in terms of predicted fairness and accuracy.}
Comprehensive experiments showcase that \sysname{} consistently outperforms baselines by a considerable margin. For all tables in the main paper and Appendix, results shown in each column represent performance on the target domain, using the rest as source domains. Due to space limit, selected results for three domains of \texttt{FairFace} are shown in \cref{tab:result-fairface}, but the average results are based on all domains. Complete performance for all domains of datasets refers to \cref{app:addtional-results}.
As shown in \cref{tab:result-fairface}, for the \texttt{FairFace} dataset, our method has the best accuracy and fairness level for the average DG performance over all the domains. 
More specifically, our method has better fairness metrics (3\% for $DP$, 2\% for $AUC_{fair}$) and comparable accuracy (0.19\% better) than the best of the baselines for individual metrics. As shown in \cref{tab:result-YFCC100M}, for \texttt{YFCC100M-FDG}, our method excels in fairness metrics (8\% for $DP$, 4\% for $AUC_{fair}$) and comparable accuracy (0.35\% better) compared to the best baselines.

\textbf{Ablation studies.}
We conduct three ablation studies to study the robustness of \sysname{} on \texttt{FairFace}. In-depth descriptions and the pseudocodes for these studies can be found in \cref{app:abs}. More results can be found in \cref{app:addtional-results}. (1) In \sysnameabs{}, we modify the encoder within $T$ by restricting its output to only latent semantic and style factors. (2) \sysnameabss{} skips data augmentation in synthetic domains via $T$ and results are conducted only based $f$ constrained by fair notions outlined in \cref{def:fairNotion}. (3) In \sysnameabsss{}, the fair constraint on $f$ is not included, and we eliminate the $\mathcal{L}_{fair}$ in line 9 of \cref{alg:algor-s2}. We include the performance of such ablation studies in \cref{fig:abs-fairface}. The results illustrate that when data is disentangled into three factors, and the model is designed accordingly, it can enhance generalization performance due to covariate and dependence shifts. Generating data in synthetic domains with random fairness dependence patterns proves to be an effective approach for ensuring fairness invariance across domains.

\textbf{Fairness-accuracy tradeoff.} In our \cref{alg:algor-s2}, because $\lambda_2$ (lines 10 and 12) is the parameter that regularizes the fair loss, we conduct additional experiments to show the change of tradeoffs between accuracy and fairness sweeping over a range of $\lambda_2\in[0.01, 0.05, 0.1, 1, 10]$. Our results show that the larger (small) $\lambda_2$, the better(worse) model fairness for each domain as well as in average, but it gives worse (better) model accuracy.
Evaluation on \texttt{FairFace} and \texttt{YFCC100M-FDG} is given in \cref{fig:tradeoff}. Results in the top-right of the figure indicate good performance. This result is plotted on the average performance over all target domains.

%% file: conclusions.tex

In this paper, we introduce a novel approach designed to tackle the challenges of domain generalization when confronted with covariate shift and dependence shift simultaneously. 
We present a tractable algorithm and showcase its effectiveness through comprehensive analyses and exhaustive empirical studies.




%% file: app_notations.tex
For clear interpretation, we list the notations used in this paper and their corresponding explanation, as shown in \cref{tab:detailed_notations}.

\begin{table}[h!]
    \caption{Important notations and corresponding descriptions.}
    \centering
    \begin{tabular}{c|p{6.5cm}}
        \toprule
        \textbf{Notations} & \textbf{Descriptions} \\
        \cmidrule(lr){1-2}
        $\mathcal{X}$ & input feature space\\
        $\mathcal{Z}$ & sensitive space\\
        $\mathcal{Y}$ & output space \\
        $\mathcal{C}$ & latent space for semantic factors\\
        $\mathcal{S}$ & latent space for style factors\\
        $\mathcal{A}$ & latent space for sensitive factors\\
        $\mathbf{c}$ & semantic factor\\
        $\mathbf{s}$ & style factor\\
        $\mathbf{a}$ & sensitive factor\\
        $d[\cdot]$ & distance metric over outputs\\
        $dist[\cdot]$ & distance metric over distributions\\
        $\mathcal{D}$ & data set\\
        $\mathbf{x}$ & data features\\
        $y$ & class label\\
        $z$ & sensitive attribute\\
        $f$ & classifier\\
        $\mathcal{F}$ & classifier space\\
        $\hat{f}$ & $\xi$-parameterization of $\mathcal{F}$\\
        $\hat{y}$ & predicted class label\\
        $\Theta$ & parameter space\\
        $g(,)$ & fairness function\\
        $|\cdot|$ & absolute function\\
        $p_1$ & empirical estimate of the proportion of samples in the group $z=1$\\
        $e$ & domain labels\\
        $s$ & source domain labels\\
        $\mathcal{E}$ & set of data labels\\
        $\mathcal{B}$ & sampled data batch\\
        $T$ & domain transformation model\\
        $E$ & encoder network \\
        $G$ & decoder network\\
        $\mathcal{L}$ & loss function\\
        $\delta$ & expectation of the relaxed constraint\\
        $h$ & sensitive classifier\\
        $\hat{z}$ & sensitive attributes predicted by $h$\\
        $\eta_p,\eta_d$ & primal and dual learning rate\\
        $\lambda$ & dual variable\\
        $\gamma$ & empirical constant\\
         \bottomrule
    \end{tabular}
    \label{tab:detailed_notations}
\end{table}

%% file: app_expdetails.tex
\subsection{Evaluation Metrics.}
\label{sec:evalMetrics}
Two fair metrics are used for evaluation. 
\begin{itemize}[leftmargin=*]
    \item \textit{Demographic Parity} ($DP$) \cite{Dwork-2011-CoRR} is formalized as
    \begin{align*}
        \text{DP}=k, \:\text{if DP}\leq 1; \text{DP}=1/k, \:\text{otherwise}
    \end{align*}
    where $k=\mathbb{P}(\hat{Y}=1|Z=-1) / \mathbb{P}(\hat{Y}=1|Z=1)$
    This is also known as a lack of disparate impact \cite{Feldman-KDD-2015}. A value closer to 1 indicates fairness. 

    \item \textit{The Area Under the ROC Curve} ($AUC_{fair}$) \cite{Calders-ICDM-2013} varies from zero to one, and it is symmetric around 0.5, which represents random predictability or zero bias effect on predictions.
    \begin{align*}
        \frac{\sum_{(\mathbf{x}_i,z=-1,y_i)\in \mathcal{D}_{-1}}\sum_{(\mathbf{x}_j,z=1,y_j)\in \mathcal{D}_1}I\big(\mathbb{P}(\hat{y}_i=1)>\mathbb{P}(\hat{y}_j=1)\big)}{|\mathcal{D}_{-1}|\times |\mathcal{D}_{1}|}
    \end{align*}
    where $|\mathcal{D}_{-1}|$ and $|\mathcal{D}_{1}|$ represent sample size of subgroups $z=-1$ and $z=1$, respectively. $I(\cdot)$ is the indicator function that returns 1 when its argument is true and 0 otherwise. 
\end{itemize}

\subsection{Details of Learning the Transformation Model}
\begin{figure}[t]
    \centering
    \includegraphics[width=0.8\linewidth]{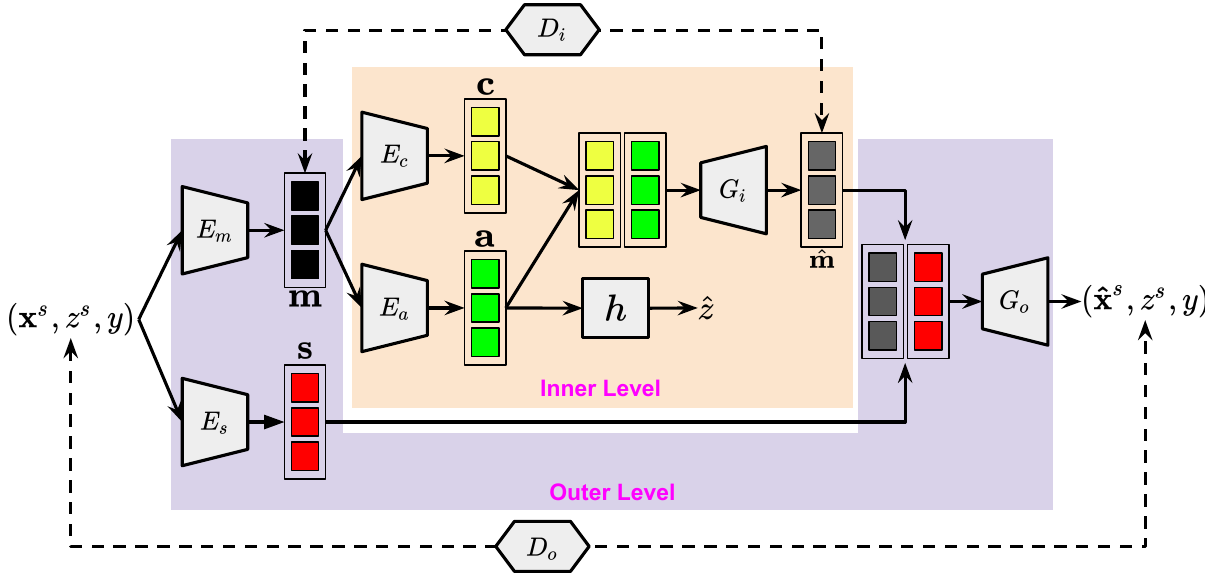}
    \caption{A two-level approach for leaning the transformation model $T$.}
    \label{fig:learningT}
\end{figure}

For simplicity, we denote the transformation model $T$ consisting of three encoders $E_c,E_a,E_s$, and a decoder $G$.
However, in practice, we consider a bi-level auto-encoder (see \cref{fig:learningT}), wherein an additional content encoder $E_m:\mathcal{X}\rightarrow\mathcal{M}$ takes data as input and outputs a content factor. 
Furthermore, the decoder $G$ used in the main paper is renamed $G_o$. Specifically, the inner level decoder is denoted as $G_i:\mathcal{C}\times\mathcal{A}\rightarrow\mathcal{M}$.
As a consequence, the transformation model $T$ consists of encoders $E=\{E_m,E_s,E_c,E_a\}$ and decoders $G= \{G_i,G_o\}$.



Specifically, in the outer level, an instance is first encoded to a content factor $\mathbf{m}\in\mathcal{M}$ and a style factor $\mathbf{s}\in\mathcal{S}$ through the corresponding encoders $E_m$ and $E_s$, respectively. In the inner level, the content factor $\mathbf{m}$ is further encoded to a content factor $\mathbf{c}\in\mathcal{C}$ and a sensitive factor $\mathbf{a}\in\mathcal{A}$, through encoders $E_c$ and $E_a$. 
Therefore, the bidirectional reconstruction loss and the sensitiveness loss stated in \cref{sec:method} are reformulated.
\begin{align*}
    \mathcal{L}_{recon}^{data} = 
    \mathbb{E}_{\mathbf{x}^s\sim \mathbb{P}^s_X} \big[\big\lVert G_o\big(\hat{\mathbf{m}}, E_s(\mathbf{x}^s)\big)-\mathbf{x}^s \big\rVert_1\big] 
    + \mathbb{E}_{\mathbf{m}\sim \mathbb{P}_M} \big[\big\lVert G_i\big(E_c(\mathbf{m}), E_a(\mathbf{m})\big)-\mathbf{m} \big\rVert_1\big]
\end{align*}
where $\hat{\mathbf{m}}=G_i(\mathbf{c,a})=G_i\big(E_c(E_m(\mathbf{x}^s)), E_a(E_m(\mathbf{x}^s))\big)$; $\mathbb{P}_M$ is given by $\mathbf{m}=E_m(\mathbf{x}^s)$. 
\begin{equation}
\begin{aligned}
    \mathcal{L}_{recon}^{factor} =
    &\mathbb{E}_{\mathbf{c}\sim \mathbb{P}_C,\mathbf{a}\sim \mathcal{N}(0,\mathbf{I}_a)} \big[\big\lVert E_c\big(G_i(\mathbf{c,a}) \big)-\mathbf{c} \big\rVert_1\big]
    + 
    \mathbb{E}_{\mathbf{c}\sim \mathbb{P}_C,\mathbf{a}\sim \mathcal{N}(0,\mathbf{I}_a)} \big[\big\lVert E_a\big(G_i(\mathbf{c,a}) \big)-\mathbf{a} \big\rVert_1\big] \nonumber\\
    &+\mathbb{E}_{\mathbf{m}\sim \mathbb{P}_M, \mathbf{s}^s\sim \mathcal{N}(0,\mathbf{I}_s)} \big[\big\lVert E_s(G_o(\mathbf{m,s}))-\mathbf{s}\big\rVert_1 \big] 
    +
    \mathbb{E}_{\mathbf{c}\sim \mathbb{P}_C,\mathbf{s}^s\sim \mathcal{N}(0,\mathbf{I}_s), \mathbf{a}\sim \mathcal{N}(\mathbf{0},\mathbf{I}_a)} \big[\big\lVert E_s\big(G_o(G_i(\mathbf{c,a}), \mathbf{s}) \big)-\mathbf{s} \big\rVert_1\big] \nonumber\\
    &+\mathbb{E}_{\mathbf{m}\sim \mathbb{P}_M,\mathbf{s}^s\sim \mathcal{N}(0,\mathbf{I}_s)} \big[\big\lVert E_m\big(G_o(\mathbf{m,s}) \big)-\mathbf{m} \big\rVert_1\big]
\end{aligned}
\end{equation}
where $\mathbb{P}_C$ and $\mathbb{P}_M$ are given by $\mathbf{c}=E_c(E_m(\mathbf{x}^s))$ and $\mathbf{m}=E_m(\mathbf{x}^s)$. $\mathbf{a}=E_a(E_m(\mathbf{x}^s))$, and $\mathbf{s}=E_s(\mathbf{x}^s)$.
\begin{align*}
    \mathcal{L}_{sens} = CrossEntropy(z^s,h(E_a(E_m(\mathbf{x}^s))))
\end{align*}
Additionally, motivated by the observation that GANs \cite{goodfellow2020generative} can improve data quality for evaluating the disentanglement effect in the latent spaces, we use GANs to match the distribution of reconstructed data to the same distribution. 
Followed by \cite{huang2018multimodal}, data and semantic factors generated through encoders and decoders should be indistinguishable from the given ones in the same domain.
\begin{align*}
    \mathcal{L}_{adv} 
    = &\mathbb{E}_{\mathbf{c}\sim \mathbb{P}_C,\mathbf{s}^s\sim \mathcal{N}(0,\mathbf{I}_s), \mathbf{a}\sim \mathcal{N}(\mathbf{0},\mathbf{I}_a)} \big[\log\big(1-D_o(G_o(\hat{\mathbf{m}},\mathbf{s}^s)) \big) \big]
    +\mathbb{E}_{\mathbf{x}^s\sim \mathbb{P}^s_X} \big[\log D_o(\mathbf{x}^s) \big]\\
    &+ \mathbb{E}_{\mathbf{c}\sim \mathbb{P}_C, \mathbf{a}\sim \mathcal{N}(\mathbf{0},\mathbf{I}_a)} \big[\log\big(1-D_i(G_i(\mathbf{c,a})) \big) \big]
    +\mathbb{E}_{\mathbf{m}\sim \mathbb{P}_M} \big[\log D_i(\mathbf{m}) \big]
\end{align*}
where $D_o:\mathcal{X}\rightarrow\mathbb{R}$ and $D_i:\mathcal{M}\rightarrow\mathbb{R}$ are the discriminators for the outer and inner levels, respectively.

\textbf{Total Loss.} We jointly train the encoders, decoders, and discriminators to optimize the final objective, a weighted sum of the three loss terms.
\begin{align}
\label{eq:total-loss}
    \min_{E_m,E_s,E_c,E_a,G_i,G_o}\: \max_{D_i,D_o} \: \beta_1\mathcal{L}^{data}_{recon} + \beta_2\mathcal{L}^{factor}_{recon} + \beta_3\mathcal{L}_{sens} + \beta_4\mathcal{L}_{adv}
\end{align}
where $\beta_1, \beta_2, \beta_3, \beta_4 >0$ are hyperparameters that control the importance of each loss term. To optimize, the learning algorithm is given in \cref{alg:algor-s1}.

\begin{algorithm}[t]
\caption{Learning the Transformation Model $T$.}
\label{alg:algor-s1}
\begin{flushleft}
    \textbf{Require}: learning rate $\alpha_1,\alpha_2,\alpha_3$, initial coefficients $\beta_1, \beta_2, \beta_3, \beta_4$.\\
    \textbf{Initialize}: Parameter of encoders $\{\boldsymbol{\theta}_m, \boldsymbol{\theta}_s, \boldsymbol{\theta}_c, \boldsymbol{\theta}_a$\}, decoders $\{\boldsymbol{\phi}_i, \boldsymbol{\phi}_o\}$, sensitive classifier $\boldsymbol{\theta}_z$, and discriminators $\{\boldsymbol{\psi}_i, \boldsymbol{\psi}_o\}$.
\end{flushleft}
\begin{algorithmic}[1]
\Repeat
    \For{minibatch $\{(\mathbf{x}_i, y_i, z_i)\}_{i=1}^q \in\mathcal{D}_{s}$}
        \State Compute $\mathcal{L}_{total}$ for Stage 1 using \cref{eq:total-loss}.
        \State $\boldsymbol{\psi}_o,\boldsymbol{\psi}_i\leftarrow\text{Adam}(\beta_4\mathcal{L}_{adv},\boldsymbol{\psi}_o, \boldsymbol{\psi}_i, \alpha_1)$
        \State \multiline{%
            $\boldsymbol{\theta}_m, \boldsymbol{\theta}_c, \boldsymbol{\theta}_s, \boldsymbol{\theta}_a, \boldsymbol{\phi}_o, \boldsymbol{\phi}_i \leftarrow\text{Adam}\big(\beta_1\mathcal{L}^{data}_{recon}+\beta_2\mathcal{L}^{factor}_{recon}, \boldsymbol{\theta}_m, \boldsymbol{\theta}_c, \boldsymbol{\theta}_s, \boldsymbol{\theta}_a, \boldsymbol{\phi}_o, \boldsymbol{\phi}_i, \alpha_2\big)$}
        \State $\boldsymbol{\theta}_z\leftarrow\text{Adam}(\beta_3\mathcal{L}_{sens}, \boldsymbol{\theta}_z, \alpha_3)$
    \EndFor
\Until{convergence}
\State \textbf{Return} $\{\boldsymbol{\theta}_m, \boldsymbol{\theta}_s, \boldsymbol{\theta}_c, \boldsymbol{\theta}_a, \boldsymbol{\theta}_z, \boldsymbol{\phi}_{i}, \boldsymbol{\phi}_o\}$
\end{algorithmic}
\end{algorithm}

\subsection{Architecture Details}
\label{sec:app_arch_details}

We have two sets of networks. One is for \texttt{ccMNIST}, \texttt{FairFace}, and \texttt{YFCC100M-FDG}, and the other one is for the \texttt{NYSF} dataset.

For \texttt{ccMNIST}, \texttt{FairFace}, and \texttt{YFCC100M-FDG} datasets: All the images are resized to $224\times224$. 
$E_m$ and $E_c$'s structures are the same. Each of them is made of four convolution layers. The first one has $64$ filters, and each of the others has $128$ filters. The kernel sizes are $(7,7), (4,4), (3,3), (3,3)$ for layers $1$ to $4$, respectively. The stride of the second layer is $(2,2)$, and the stride of all the other layers is $(1,1)$. The activation function of the first three layers is ReLU. The last convolution layer does not have an activation function.
$E_s$ and $E_a$'s structures are the same. Each of them is made of $6$ convolution layers, and there is an adaptive average pooling layer with output size $1$ between the last two convolution layers. The numbers of filters are $64, 128, 256, 256, 256$, and $2$ for the convolution layers, respectively. The kernel sizes are $(7,7), (4,4), (4,4), (4,4), (4,4), (1,1)$. And the strides are $(1,1), (2,2), (2,2), (2,2), (2,2), (1,1)$. The activation function of the first five layers is ReLU. The last convolution layer does not have an activation function.
$G_o$ and $G_i$'s structures are almost the same. The only difference between them is the output size, $3$ for $G_o$ and $128$ for $G_i$. Each of them has two parts. 
The first part is made of $4$ convolution layers, and there is an upsampling layer with a scale factor $2.0$ between the second convolution layer and the third convolution layer. The numbers of filters are $128, 128, 64,$ and $3$ for the convolution layers, respectively. The kernel sizes are $(3,3), (3,3), (5,5), (7,7)$. The strides are $(1,1)$ for all the convolution layers. The first and the third convolution layers' activation functions are ReLU. The fourth convolution layer's activation function is Tanh. The second convolution layer does not have an activation function.
The second part is made of three fully connected layers. The number of neurons is $256$ and $256$, respectively, and the output size is $512$. The activation function of the first two layers is ReLU, and there is no activation function on the output.
$D_o$ comprises $4$ convolution layers followed by an average pooling layer whose kernel size is $3$, stride is $2$, and padding is $[1,1]$. The numbers of filters of the convolution layers are $64, 128, 256, 1$, respectively. The kernel sizes are $(4,4)$ for the first three convolution layers and $(1,1)$ for the fourth convolution layer. The strides are $(2,2)$ for the first three convolution layers and $(1,1)$ for the fourth convolution layer. The first three convolution layers' activation functions are LeakyReLU. The other layers do not have activation functions.
$D_i$ is made of one fully connected layer whose input size is $112$, and the output size is $64$ with the activation function ReLU.
$h$ comprises one fully connected layer with input size $2$, output size $1$, and activation function Sigmoid.
$f$ has two parts. The first part is Resnet-50 \cite{he2016deep}, and the second is one fully connected layer with input size $2048$ and output size $2$.

For the \texttt{NYSF} dataset:
$E_m$ is made of two fully connected layers. The number of neurons is $32$, and the output size is $16$. The activation function of the first layer is ReLU, and there is no activation function on the output.
$E_s$ is made of two fully connected layers. The number of neurons is $32$, and the output size is $2$. The activation function of the first layer is ReLU, and there is no activation function on the output.
$G_o$ is made of two fully connected layers. The number of neurons is $32$, and the output size is $51$. The activation function of the first layer is ReLU, and there is no activation function on the output.
$D_o$ is made of two fully connected layers. The number of neurons is $32$, and the output size is $16$. The activation function of the first layer is ReLU, and there is no activation function on the output.
$E_c$ is made of two fully connected layers. The number of neurons is $16$, and the output size is $8$. The activation function of the first layer is ReLU, and there is no activation function on the output.
$E_a$ is made of two fully connected layers. The number of neurons is $8$, and the output size is $2$. The activation function of the first layer is ReLU, and there is no activation function on the output.
$G_i$ is made of two fully connected layers. The number of neurons is $16$, and the output size is $16$. The activation function of the first layer is ReLU, and there is no activation function on the output.
$D_i$ is made of two fully connected layers. The number of neurons is $8$, and the output size is $8$. The activation function of the first layer is ReLU, and there is no activation function on the output.
$h$ comprises one fully connected layer with input size $2$ and output size $1$. The activation function is Sigmoid.
$f$ has two parts. The first part is made of $3$ fully connected layers. The number of neurons is $32$, and the output size is $32$. The activation function of the first two layers is ReLU, and there is no activation function on the output. The second part is made of one fully connected layer whose input size is $32$, the output size is $32$, and it does not have an activation function.

\subsection{Hyperparameter Search}
\label{sec:app_hyper_search}
We follow the same set of the MUNIT \cite{huang2018multimodal} for the hyperparameters. More specifically, the learning rate is $0.0001$, the number of iterations is $600000$, and the batch size is $1$. The loss weights in learning $T$ are chosen from $\{1, 5, 10\}$. The selected best ones are $\beta_1=10, \beta_2=1, \beta_3=1, \beta_4=1$. We monitor the loss of the validation set and choose the $\beta$ with the lowest validation loss. 

For the hyperparameters in learning the classifier $f$, the learning rate is chosen from $\{0.000005, 0.00001, 0.00005, 0.0001, 0.0005\}$. $\eta$ is chosen from $\{0.01, 0.05, 0.1\}$. $\gamma$ is chosen from $\{0.01,0.025,0.05\}$. $\lambda$ is chosen from $\{0.1, 1, 10, 20\}$. The batch size is chosen from $\{22,64,80,128,512,1024,2048\}$. The numbers of iterations is chosen from $\{500,1000,...,8000\}$ on the \texttt{ccMNIST} and \texttt{NYSF} datasets. The number of iterations are chosen from $\{300,600,...,7800,8000\}$ on the \texttt{FairFace} and \texttt{YFCC100M-FDG} datasets.  The selected best ones are: the learning rate is $0.00005$, $\eta_1=\eta_2=0.05$, $\gamma_1=\gamma_2=0.025$, $\lambda_1=\lambda_2=1$. The batch size on the \texttt{ccMNIST} and \texttt{YFCC100M-FDG} datasets is $64$, and it is $22$ on the \texttt{FairFace} dataset and $1024$ on the \texttt{NYSF} dataset. The number of iterations on the \texttt{ccMNIST} dataset is $3000, 500, 7000$ for domains R, G, B, respectively. The number of iterations on the \texttt{FairFace} dataset is $7200, 7200, 7800, 8000, 6600, 7200, 6900$ for domains B, E, I, L, M, S, W, respectively. The number of iterations on the \texttt{YFCC100M-FDG} dataset is $7200, 6000, 6900$ for $d_0, d_1, d_2$, respectively. The number of iterations on the \texttt{NYSF} dataset is $500, 3500, 4000, 1500, 8000$ for domains R, B, M, Q, S, respectively. We monitor the accuracy and the value of fairness metrics from the validation set and select the best ones.
The grid space of the grid search on all the baselines is the same as for our method.


%% file: app_abs.tex
We conduct three ablation studies, and detailed algorithms of designed ablation studies are given in \cref{alg:abs1,alg:abs2,alg:abs3}. For additional ablation study results on \texttt{ccMNIST}, \texttt{YFCC100M-FDG}, and \texttt{NYSF}, refer to \cref{app:addtional-results}. 
\begin{enumerate}[leftmargin=*]
    \item The difference between the full \sysname{} and the first ablation study (\sysnameabs{}) is that the latter does not have the inner level when learning $T$. Since the inner level is used to extract the content and sensitive factors from the semantic one, the same sensitive label of the generated images will remain due to the absence of $h(\cdot)$. Therefore, \sysnameabs{} is expected to have a lower level of fairness in the experiments. Results shown in the tables indicate that \sysnameabs{} has a significantly lower performance on fairness metrics.
    
    \item The second study (\sysnameabss{}) does not train the auto-encoders to generate images. All losses are computed only based on the sampled images. Similar to \sysnameabs{}, it is much harder to train a good classifier without the generated images in synthetic domains.  Our results demonstrate that \sysnameabss{} performs worse on all the datasets.
    
    \item The difference between \sysname{} and the third study (\sysnameabsss{}) is that \sysnameabsss{} does not have the fairness loss $\mathcal{L}_{fair}$ in line 9 of \cref{alg:algor-s2}. Therefore, this algorithm only focuses on accuracy without considering fairness. Results based on \sysnameabsss{} show that it has a good level of accuracy but a poor level of fairness.
\end{enumerate}

\begin{algorithm}[!h]
\caption{\sysnameabs{} (Ablation Study 1)}
\label{alg:abs1}
\begin{algorithmic}[1]
\Repeat
    \For{minibatch $\mathcal{B}=\{(\mathbf{x}_i, z_i, y_i)\}_{i=1}^m \in\mathcal{D}_{s}$}
        \State $\mathcal{L}_{cls}(\boldsymbol{\theta})=(1/m)\sum_{i=1}^m\ell(y_i, \hat{f}(\mathbf{x}_i,\boldsymbol{\theta}))$
        \State \multiline{%
        $\mathcal{L}_{fair}(\boldsymbol{\theta}) = (1/m)\sum_{i=1}^{m}(\frac{1}{p_1(1-p_1)}(\frac{z_i+1}{2}-p_1)\hat{f}(\mathbf{x}_i,\boldsymbol{\theta})$}
        \For{each $(\mathbf{x}_i, z_i, y_i)$ in the minibatch}
            \State \multiline{%
            $(\mathbf{x}'_i,z_i, y_i)=\textsc{T}(\mathbf{x}_i, x_i, y_i)$}
            \State $\mathcal{L}'_{cls}(\boldsymbol{\theta})=(1/m)\sum_{i=1}^m\ell(y_i, \hat{f}(\mathbf{x}'_i,\boldsymbol{\theta}))$
        \EndFor
        \State \multiline{%
        $\mathcal{L}_{cls}(\boldsymbol{\theta}) = \mathcal{L}_{cls}(\boldsymbol{\theta}) +\mathcal{L}'_{cls}(\boldsymbol{\theta})$}
        \State $\mathcal{L}(\boldsymbol{\theta})=\mathcal{L}_{cls}(\boldsymbol{\theta}) + \lambda_2\cdot\mathcal{L}_{fair}(\boldsymbol{\theta})$
        \State $\boldsymbol{\theta}\leftarrow\boldsymbol{\theta}-\eta_p\cdot\nabla_{\boldsymbol{\theta}}\mathcal{L}(\boldsymbol{\theta})$
        \State $\lambda_2\leftarrow \text{max}\{ [\lambda_2+\eta_d\cdot(\mathcal{L}_{fair}(\boldsymbol{\theta})-\gamma_2)], 0\}$
    \EndFor
\Until{convergence}
\Procedure{\textsc{T}}{$\mathbf{x}, z, y$}
    \State $\mathbf{c}= E^{m}(\mathbf{x}, \boldsymbol{\theta}^{m})$
    \State Sample $\mathbf{s}'\sim\mathcal{N}(0,I_s)$
    \State $\mathbf{x}'= G^o(\mathbf{c}, \mathbf{s}', \boldsymbol{\phi}_o)$
    \State \textbf{return} $(\mathbf{x}', z, y)$
\EndProcedure
\end{algorithmic}
\end{algorithm}

\begin{algorithm}[!h]
\caption{\sysnameabss{} (Ablation Study 2)}
\label{alg:abs2}
\begin{algorithmic}[1]
\Repeat
    \For{minibatch $\mathcal{B}=\{(\mathbf{x}_i, z_i, y_i)\}_{i=1}^m \in \mathcal{D}_{s}$}
        \State $\mathcal{L}_{cls}(\boldsymbol{\theta})=(1/m)\sum_{i=1}^m\ell(y_i, \hat{f}(\mathbf{x}_i,\boldsymbol{\theta}))$        
        \State \multiline{%
        $\mathcal{L}_{fair}(\boldsymbol{\theta}) = (1/m)\sum_{i=1}^{m}(\frac{1}{p_1(1-p_1)}(\frac{z_i+1}{2}-p_1)\hat{f}(\mathbf{x}_i,\boldsymbol{\theta})$}
        \State $\mathcal{L}(\boldsymbol{\theta})=\mathcal{L}_{cls}(\boldsymbol{\theta}) + \lambda_2\cdot\mathcal{L}_{fair}(\boldsymbol{\theta})$
        \State $\boldsymbol{\theta}\leftarrow\boldsymbol{\theta}-\eta_p\cdot\nabla_{\boldsymbol{\theta}}\mathcal{L}(\boldsymbol{\theta})$
        \State $\lambda_2\leftarrow \text{max}\{ [\lambda_2+\eta_d\cdot(\mathcal{L}_{fair}(\boldsymbol{\theta})-\gamma_2)], 0\}$
    \EndFor
\Until{convergence}
\end{algorithmic}
\end{algorithm}

\begin{algorithm}[!h]
\caption{\sysnameabsss{} (Ablation Study 3)}
\label{alg:abs3}
\begin{algorithmic}[1]
\Repeat
    \For{minibatch $\mathcal{B}=\{(\mathbf{x}_i, z_i, y_i)\}_{i=1}^m \in\mathcal{D}_{s}$}
        \State $\mathcal{L}_{cls}(\boldsymbol{\theta})=(1/m)\sum_{i=1}^m\ell(y_i, \hat{f}(\mathbf{x}_i,\boldsymbol{\theta}))$
        \State Initialize $\mathcal{L}'_{inv}(\boldsymbol{\theta})=0$
        \For{each $(\mathbf{x}_i, z_i, y_i)$ in the minibatch}
            \State \multiline{%
            $(\mathbf{x}'_i, y_i)=\textsc{T}(\mathbf{x}_i, z_i, y_i)$}
            \State \multiline{%
            $\mathcal{L}'_{inv}(\boldsymbol{\theta}) \mathrel{+}= d[\hat{f}(\mathbf{x}_i,\boldsymbol{\theta}), \hat{f}(\mathbf{x}'_i,\boldsymbol{\theta})]$}
        \EndFor
        \State $\mathcal{L}_{inv}(\boldsymbol{\theta}) = \mathcal{L}'_{inv}(\boldsymbol{\theta})/m$

        \State $\mathcal{L}(\boldsymbol{\theta})=\mathcal{L}_{cls}(\boldsymbol{\theta}) + \lambda_1\cdot\mathcal{L}_{inv}(\boldsymbol{\theta})$
        \State $\boldsymbol{\theta}\leftarrow\boldsymbol{\theta}-\eta_p\cdot\nabla_{\boldsymbol{\theta}}\mathcal{L}(\boldsymbol{\theta})$
        \State $\lambda_1\leftarrow \text{max}\{ [\lambda_1+\eta_d\cdot(\mathcal{L}_{inv}(\boldsymbol{\theta})-\gamma_1)], 0\}$
    \EndFor
\Until{convergence}
\Procedure{\textsc{T}}{$\mathbf{x},z, y$}
    \State $\mathbf{c}= E^{c}(E^{m}(\mathbf{x}, \boldsymbol{\theta}^{m}) ,\boldsymbol{\theta}^{c})$
    \State Sample $\mathbf{a}'\sim\mathcal{N}(0,I_a)$
    \State Sample $\mathbf{s}'\sim\mathcal{N}(0,I_s)$
    \State $\mathbf{x}'= G^o(G^i(\mathbf{c}, \mathbf{a}', \boldsymbol{\phi}_{i}), \mathbf{s}', \boldsymbol{\phi}_o)$
    \State \textbf{return} $(\mathbf{x}', z, y)$
\EndProcedure
\end{algorithmic}
\end{algorithm}

%% file: app_proofs.tex
\subsection{Sketch Proof of \cref{the:duality-gap-theorem}}

Before we prove \cref{the:duality-gap-theorem}, we first make the following propositions and assumptions.
\begin{proposition}
\label{B.1}
    Let $d'$ be a distance metric between probability measures for which it holds that $d'[\mathbb{P,T}]=0$ for two distributions $\mathbb{P}$ and $\mathbb{T}$ if and only if $\mathbb{P}=\mathbb{T}$ almost surely. Then $P^\star(0,0)=P^\star$
\end{proposition}

\begin{proposition}
\label{prop:gap1}
Assuming the perturbation function $P^\star(\gamma_1,\gamma_2)$ is $L$-lipschitz continuous in $\gamma_1,\gamma_2$. Then given \cref{B.1}, it follows that $|P^\star-P^\star(\gamma_1,\gamma_2)|\leq L||\boldsymbol{\gamma}||_{1}$, where $\boldsymbol{\gamma}=[\gamma_1,\gamma_2]^T$.
\end{proposition}

\begin{definition}
Let $\Theta\subseteqq \mathbb{R}^p$ be a finite-dimensional parameter space. For $\xi>0$, a function $\hat{f}:\mathcal{X}\times\Theta\rightarrow\mathcal{Y}$ is said to be an $\xi$-parameterization of $\mathcal{F}$ if it holds that for each $f\in\mathcal{F}$, there exists a parameter $\boldsymbol{\theta}\in\Theta$ such that $\mathbb{E}_{\mathbb{P}_X}\Arrowvert \hat{f}(\mathbf{x},\boldsymbol{\theta})-f(\mathbf{x})\Arrowvert_\infty\leq\xi$.
Given an $\xi$-parameterization $\hat{f}$ of $\mathcal{F}$, consider the following saddle-point problem:
\begin{align*}
\label{saddle}
    D_\xi^\star(\gamma_1,\gamma_2)
    \triangleq\max_{\lambda_1(s_i,s_j),\lambda_2(s_i,s_j)}\min_{\boldsymbol{\theta}\in\Theta} \: &R(\boldsymbol{\theta}) +\int_{s_i,s_j\in\mathcal{E}_s}[\delta^{s_i,s_j}(\boldsymbol{\theta})-\gamma_1]\mathrm{d}\lambda_1(s_i,s_j)\\
    &+\int_{s_i,s_j\in\mathcal{E}_s}[\rho^{s_i}(\boldsymbol{\theta})+\rho^{s_j}(\boldsymbol{\theta})-\gamma_2]\mathrm{d}\lambda_2(s_i,s_j)
\end{align*}
where $R(\boldsymbol{\theta})=R(\hat{f}(\cdot,\boldsymbol{\theta}))$ and $\mathcal{L}^{s_i,s_j}(\boldsymbol{\theta})=\mathcal{L}^{s_i,s_j}(\hat{f}(\cdot,\boldsymbol{\theta}))$.
\end{definition}

\begin{assumption}
The loss function $\ell$ is non-negative, convex, and $L_\ell$-Lipschitz continuous in its first argument,
\begin{align*}
    |\ell(f_1(\mathbf{x}),y)-\ell(f_2(\mathbf{x}),y)|\leq\Arrowvert f_1(\mathbf{x})-f_2(\mathbf{x})\Arrowvert_\infty
\end{align*}
\end{assumption}

\begin{assumption}
The distance metric $d$ is non-negative, convex, and satisfies the following uniform Lipschitz-like inequality for some constant $L_d>0$:
\begin{align*}
    |d[f_1(\mathbf{x}),f_1(\mathbf{x}^\prime=T(\mathbf{x},z,s))]-d[f_2(\mathbf{x}),f_2(\mathbf{x}^\prime=T(\mathbf{x},z,s))]| 
    \leq L_d\Arrowvert f_1(\mathbf{x})-f_2(\mathbf{x})\Arrowvert_\infty, \quad \forall s\in \mathcal{E}_s
\end{align*}
\end{assumption}

\begin{assumption}
The fairness metric $g$ is non-negative, convex, and satisfies the following uniform Lipschitz-like inequality for some constant $L_g>0$:
\begin{align*}
    |(g\circ f_1)(\mathbf{x},z)-(g\circ f_2)(\mathbf{x},z)|\leq L_g\Arrowvert f_1(\mathbf{x})-f_2(\mathbf{x})\Arrowvert_\infty
\end{align*}
\end{assumption}

\begin{assumption}
    There exists a parameter $\boldsymbol{\theta}\in\Theta$ such that $\delta^{s_i,s_j}(\boldsymbol{\theta})<\gamma_1-\xi\cdot \max\{L_\ell,L_d\}$ and $\rho^{s_i}(\boldsymbol{\theta})+\rho^{s_j}(\boldsymbol{\theta})<\gamma_2-\xi\cdot \max\{L_\ell,L_g\}, \forall s_i,s_j\in \mathcal{E}_s$
\end{assumption}


\begin{proposition}
\label{prop:gap2}
Let $\gamma_1,\gamma_2>0$ be given. With the assumptions above, it holds that
\begin{align*}
    P^\star(\gamma_1,\gamma_2)\leq D_\xi^\star(\gamma_1,\gamma_2)
    \leq P^\star(\gamma_1,\gamma_2)+\xi(1+\Arrowvert \lambda^\star_{p} \Arrowvert_{1})\cdot k
\end{align*}
where $\lambda^\star_{p}$ is the optimal dual variable for a perturbed version of \cref{eq:our-problem-relax} in which the constraints are tightened to hold with margin $\gamma-\xi\cdot k$, $k=\max\{L_\ell,L_d,L_g\}$. In particular, this result implies that
\begin{align*}
    |P^\star(\gamma_1,\gamma_2)-D_\xi^\star(\gamma_1,\gamma_2)|\leq\xi k(1+\Arrowvert \lambda^\star_{p} \Arrowvert_{L_1})
\end{align*}
\end{proposition}

\begin{proposition}[Empirical gap]
\label{prop:gap3}
    Assume $\ell$ and $d$ are non-negative and bounded in $[-B,B]$ and let $d_{\mathrm{VC}}$ denote the VC-dimension of the hypothesis class $\mathcal{A}_\xi=\{\hat{f}(\cdot,\boldsymbol{\theta}):\boldsymbol{\theta}\in\Theta\}\subseteq\mathcal{F}$. Then it holds with probability $1-\omega$ over the $N$ samples from each domain that
\begin{align*}
    |D^\star_\xi(\gamma_1,\gamma_2)-D^\star_{\xi,N,\mathcal{E}_\mathrm{s}}(\gamma_1,\gamma_2)|\leq2B\sqrt{\frac{1}{N}[1+\log(\frac{4(2N)^{d_{\mathrm{VC}}}}{\omega})]}
\end{align*}
\end{proposition}

Let $\xi>0$ be given, and let $\hat{f}$ be an $\xi$-parameterization of $\mathcal{F}$. Let the assumptions hold, and further assume that $\ell$, $d$, and $g$ are $[0,B]$-bounded and that $d[\mathbb{P,T}]=0$ if and only if $\mathbb{P=T}$ almost surely, and that $P^\star(\gamma_1,\gamma_2)$ is $L$-Lipschitz. Then assuming that $\mathcal{A}_\xi=\{\hat{f}(\cdot,\theta):\theta\in\Theta\}\subseteq\mathcal{F}$ has finite VC-dimension, it holds with probability $1-\omega$ over the $N$ samples that
\begin{align*}
    |P^\star-D^\star_{\xi,N,\mathcal{E}_{s}}(\boldsymbol{\gamma})| \leq L ||\boldsymbol{\gamma}||_{1} + \xi k (1+||\boldsymbol{\lambda}^\star_{p}||_{1}) + O(\sqrt{\log(M)/M})
\end{align*}

Now we prove \cref{the:duality-gap-theorem}.
\begin{proof}
    The proof of this theorem is a simple consequence of the triangle inequality. Indeed, by combining \cref{prop:gap1,prop:gap2,prop:gap3}, we find that
\begin{align*}
    &|P^\star-D^\star_{\xi,N,\mathcal{E}_{s}}(\gamma_1,\gamma_2)|\\
    =&|P^\star+P^\star(\gamma_1,\gamma_2)-P^\star(\gamma_1,\gamma_2)+D_\xi^\star(\gamma_1,\gamma_2)
    -D_\xi^\star(\gamma_1,\gamma_2)-D^\star_{\xi,N,\mathcal{E}_{s}}(\gamma_1,\gamma_2)|\\
    \leq&|P^\star-P^\star(\gamma_1,\gamma_2)|+|P^\star(\gamma_1,\gamma_2)-D_\xi^\star(\gamma_1,\gamma_2)|
    +|D_\xi^\star(\gamma_1,\gamma_2)-D^\star_{\xi,N,\mathcal{E}_{s}}(\gamma_1,\gamma_2)|\\
    \leq &L\Arrowvert\gamma\Arrowvert_1+\xi k(1+\Arrowvert \lambda^\star_{p} \Arrowvert_{1})+2B\sqrt{\frac{1}{N}[1+\log(\frac{4(2N)^{d_{\mathrm{VC}}}}{\omega})]}
\end{align*}
\end{proof}

\subsection{Sketch Proof of \cref{the:fairUpperBound}}
\begin{lemma}
\label{lemma:ebound_no_abs}
    Given two domains $e_i,e_j\in\mathcal{E}$, $\mathbb{E}_{\mathbb{P}^{e_j}_{XZ}} g(f(X^{e_j}),Z^{e_j})$ can be bounded by $\mathbb{E}_{\mathbb{P}^{e_i}_{XZ}}g(f(X^{e_i}),Z^{e_i})$ as follows:
    \begin{align*}
        \mathbb{E}_{\mathbb{P}^{e_j}_{XZ}}g(f(X^{e_j}),Z^{e_j}) \leq
        \mathbb{E}_{\mathbb{P}^{e_i}_{XZ}}g(f(X^{e_i}),Z^{e_i}) + \sqrt{2}dist[\mathbb{P}^{e_j}_{XZY}, \mathbb{P}^{e_i}_{XZY}]
    \end{align*}
\end{lemma}

\begin{lemma}
\label{lemma:ebound_abs}
    Given two domains $e_i,e_j\in\mathcal{E}$, under \cref{lemma:ebound_no_abs}, $\rho^{e_j}(f)$ can be bounded by $\rho^{e_i}(f)$ as follows:
    \begin{align*}
        \rho^{e_j}(f)\leq \rho^{e_i}(f)+ \sqrt{2}dist[\mathbb{P}^{e_j}_{XZY}, \mathbb{P}^{e_i}_{XZY}]
    \end{align*}
\end{lemma}

Under \cref{lemma:ebound_no_abs,lemma:ebound_abs}, we now prove \cref{the:fairUpperBound}
\begin{proof}
    Let $s_\star\in\mathcal{E}_s$ be the source domain nearest to the target domain $t\in\mathcal{E}\backslash\mathcal{E}_s$. Under \cref{lemma:ebound_abs}, we have
    \begin{align*}
        \rho^{t}(f)\leq \rho^{s_i}(f)+ \sqrt{2}dist[\mathbb{P}^{t}_{XZY}, \mathbb{P}^{e_i}_{XZY}]
    \end{align*}
    where $s_i\in\mathcal{E}_s$. Taking the average of upper bounds based on all source domains, we have:
    \begin{align*}
        \rho^{t}(f) 
        \leq &\frac{1}{|\mathcal{E}_s|}\sum_{s_i\in\mathcal{E}_s}\rho^{s_i}(f)+ \frac{\sqrt{2}}{|\mathcal{E}_s|}\sum_{s_i\in\mathcal{E}_s} dist[\mathbb{P}^{t}_{XZY}, \mathbb{P}^{e_i}_{XZY}] \\
        \leq &\frac{1}{|\mathcal{E}_s|}\sum_{s_i\in\mathcal{E}_s}\rho^{s_i}(f)+ \frac{\sqrt{2}}{|\mathcal{E}_s|}|\mathcal{E}_s| dist[\mathbb{P}^{t}_{XZY}, \mathbb{P}^{s_\star}_{XZY}]
        +\frac{\sqrt{2}}{|\mathcal{E}_s|}\sum_{s_i\in\mathcal{E}_s} dist[\mathbb{P}^{s_\star}_{XZY}, \mathbb{P}^{s_i}_{XZY}] \\
        \leq &\frac{1}{|\mathcal{E}_s|}\sum_{s_i\in\mathcal{E}_s}\rho^{s_i}(f)+ \sqrt{2}\min_{s_i\in\mathcal{E}_s} dist[\mathbb{P}^{t}_{XZY}, \mathbb{P}^{e_i}_{XZY}]
        +\sqrt{2}\max_{s_i,s_j\in\mathcal{E}_s} dist[\mathbb{P}^{e_i}_{XZY}, \mathbb{P}^{e_j}_{XZY}]
    \end{align*}
\end{proof}

%% file: app_additional_results.tex
\begin{figure}
    \centering
        \begin{subfigure}[b]{0.325\linewidth}
            \includegraphics[width=\linewidth]{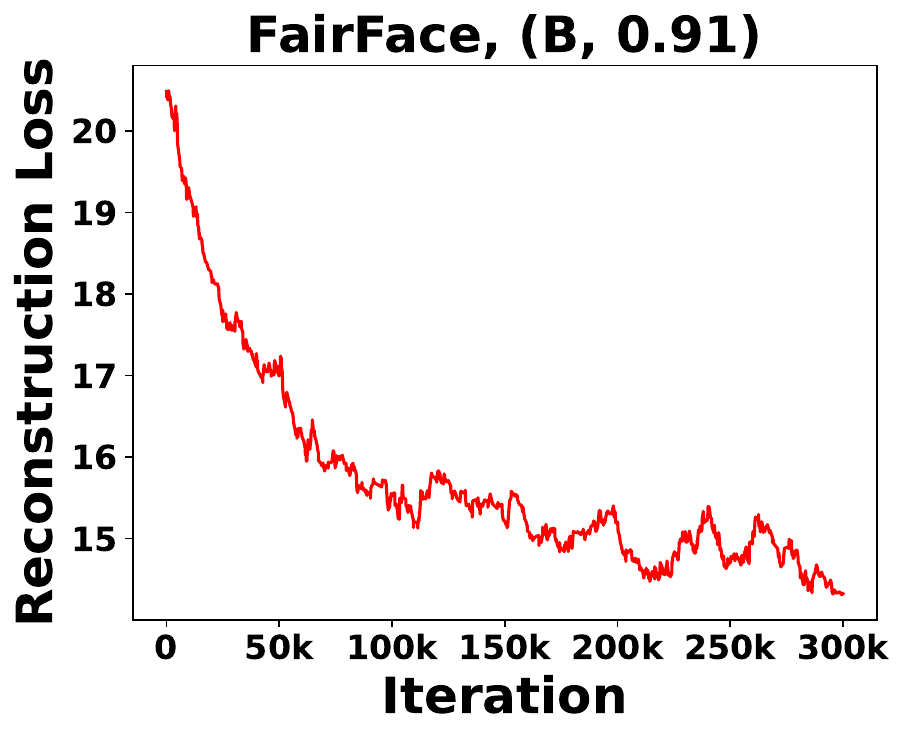}
        \end{subfigure}
        \begin{subfigure}[b]{0.325\linewidth}
            \includegraphics[width=\linewidth]{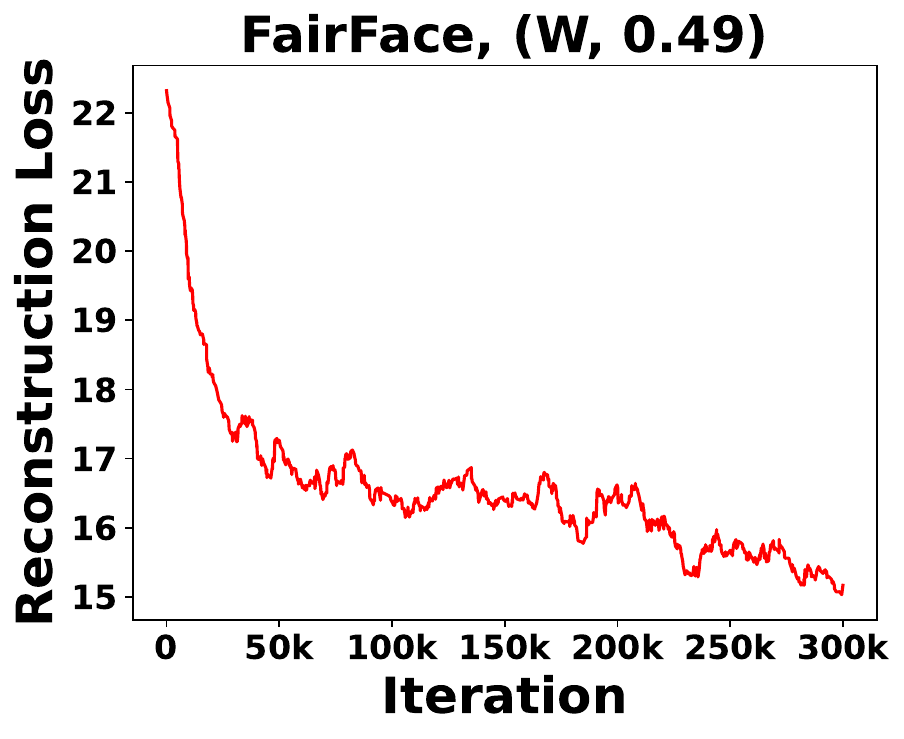}
        \end{subfigure}
        \begin{subfigure}[b]{0.325\linewidth}
            \includegraphics[width=\linewidth]{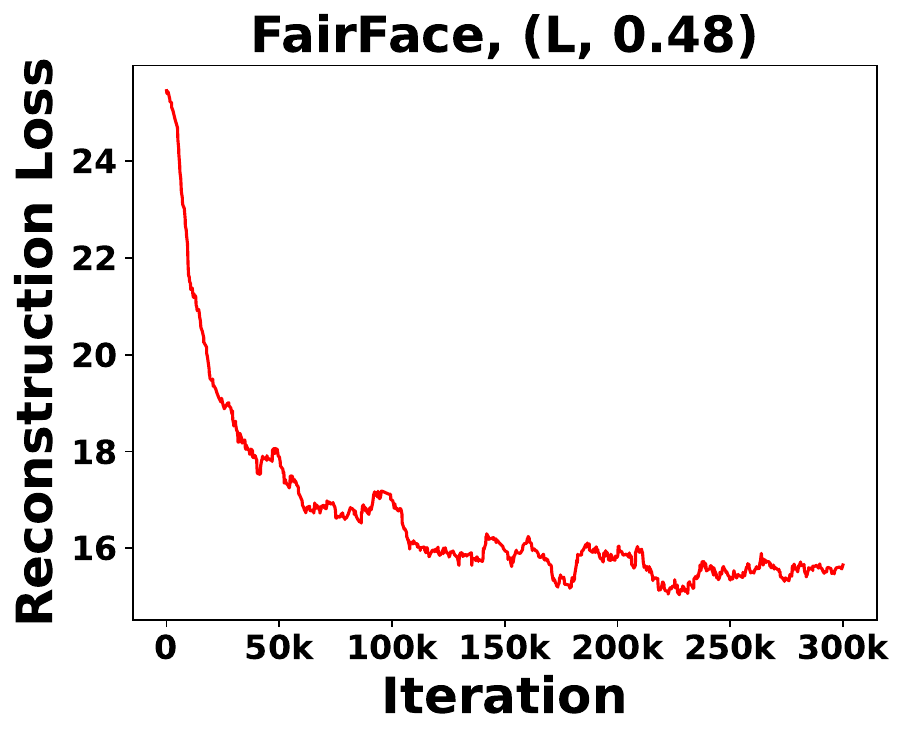}
        \end{subfigure}
        \caption{Tracking the change of reconstruction loss using the B/W/L domains of the \texttt{FairFace} dataset. }
        \label{fig:loss_curve}
\end{figure}

Additional results including complete results with all domains and baselines on \texttt{ccMNIST} (\cref{tab:result-ccMNIST-full}), \texttt{FairFace} (\cref{tab:result-fairface-full}), \texttt{FairFace} (\cref{tab:result-YFCC100M-full}), and \texttt{NYSF} (\cref{tab:result-NY-full}) are provided. 
We showcase the reconstruction loss using the \texttt{FairFace} data in \cref{fig:loss_curve}.
Additional ablation study results are in \cref{tab:abs-ccmnist,tab:abs-fairface,tab:abs-yfcc,tab:abs-nysf}.


\begin{table*}[!t]
\tiny
    \centering
    \setlength\tabcolsep{4.8pt}
    \caption{Full performance on \texttt{ccMNIST}. (bold is the best; underline is the second best).}
    \label{tab:result-ccMNIST-full}
    \begin{tabular}{l|c|c|c|c}
        \toprule
         & \multicolumn{4}{c}{$DP$ $\uparrow$  / $AUC_{fair}$ $\downarrow$ / \textbf{Accuracy} $\uparrow$} \\ 
        \cmidrule(lr){2-5}
        \multirow{-2}{*}{\textbf{Methods}} & (R, 0.11) & (G, 0.43) & (B, 0.87) & \textbf{Avg}\\
        \cmidrule(lr){1-1} \cmidrule(lr){2-2} \cmidrule(lr){3-3} \cmidrule(lr){4-4} \cmidrule(lr){5-5}
        ColorJitter & 0.11$\pm$0.05 / 0.95$\pm$0.02 / 90.59$\pm$0.23 & 0.44$\pm$0.01 / 0.71$\pm$0.03 / 87.62$\pm$0.22 & 0.87$\pm$0.03 / 0.66$\pm$0.01 / 86.33$\pm$1.50 &  0.47 / 0.77 / 88.18 \\
        ERM & 0.12$\pm$0.25 / 0.91$\pm$0.03 / 98.00$\pm$1.14 & 0.43$\pm$0.23 / 0.78$\pm$0.01 / 98.07$\pm$0.35  & 0.89$\pm$0.06 / 0.64$\pm$0.01 / 95.64$\pm$1.75 & 0.48 / 0.78 / 97.24\\
        IRM  & \underline{0.21}$\pm$0.15 / 0.97$\pm$0.02 / 75.50$\pm$2.11 & 0.28$\pm$0.10 / \underline{0.64}$\pm$0.01 / 92.74$\pm$0.27 & 0.76$\pm$0.12 / 0.63$\pm$0.03 / 80.05$\pm$2.34 & 0.42 / 0.75 / 82.76\\
        GDRO  & 0.12$\pm$0.09 / 0.92$\pm$0.03 / 98.19$\pm$0.93 & 0.43$\pm$0.06 / 0.75$\pm$0.03 / 98.17$\pm$0.87 & 0.90$\pm$0.07 / 0.65$\pm$0.01 / 95.03$\pm$0.12 & 0.48 / 0.77 / 97.13\\
        Mixup & 0.12$\pm$0.21 / 0.92$\pm$0.02 / 97.89$\pm$1.97 & 0.41$\pm$0.06 / 0.79$\pm$0.02 / 98.00$\pm$1.36 & 0.93$\pm$0.04 / 0.65$\pm$0.01 / 96.09$\pm$1.07 & 0.49 / 0.79 / 97.32\\
        MLDG  & 0.11$\pm$0.12 / 0.91$\pm$0.03 / \underline{98.52}$\pm$0.94 & 0.43$\pm$0.22 / 0.77$\pm$0.02 / \textbf{98.67}$\pm$0.61 & 0.87$\pm$0.09 / 0.62$\pm$0.03 / 93.76$\pm$1.50 & 0.46 / 0.77 / 96.98\\
        CORAL & 0.11$\pm$0.08 / 0.91$\pm$0.03 / \textbf{98.69}$\pm$0.76 & 0.42$\pm$0.20 / 0.79$\pm$0.02 / \underline{98.30}$\pm$0.98 & 0.87$\pm$0.07 / 0.64$\pm$0.01 / 93.74$\pm$1.54 & 0.47 / 0.78 / 96.91\\
        MMD & 0.11$\pm$0.08 / 0.92$\pm$0.01 / \textbf{98.69}$\pm$1.07 & 0.41$\pm$0.21 / 0.73$\pm$0.03 / 97.72$\pm$1.31 & 0.93$\pm$0.04 / 0.59$\pm$0.01 / 95.37$\pm$1.56 & 0.48 / 0.75 / 97.26\\
        DANN & 0.14$\pm$0.08 / 0.87$\pm$0.03 / 85.94$\pm$1.76 & 0.17$\pm$0.13 / 0.90$\pm$0.03 / 84.93$\pm$0.67 & 0.76$\pm$0.17 / 0.63$\pm$0.03 / 84.04$\pm$1.75 & 0.36 / 0.80 / 84.97\\
        CDANN & 0.19$\pm$0.13 / 0.90$\pm$0.01 / 93.03$\pm$2.18 & 0.60$\pm$0.17 / 0.89$\pm$0.03 / 71.92$\pm$1.03 & 0.77$\pm$0.14 / 0.63$\pm$0.02 / 84.03$\pm$1.96 & 0.52 / 0.81 / 82.99\\
        DDG  & 0.11$\pm$0.07 / 0.91$\pm$0.01 / 98.26$\pm$2.38 & 0.42$\pm$0.14 / 0.77$\pm$0.02 / 98.14$\pm$0.11 & \underline{0.96}$\pm$0.03 / 0.60$\pm$0.01 / \textbf{97.02}$\pm$1.70 & 0.50 / 0.76 / \textbf{97.81}\\
        MBDG & 0.12$\pm$0.04 / 0.93$\pm$0.01 / 98.47$\pm$0.94 & 0.42$\pm$0.08 / 0.81$\pm$0.03 / 97.62$\pm$1.87 & 0.90$\pm$0.08 / 0.64$\pm$0.03 / 96.01$\pm$2.26 & 0.48 / 0.79 / \underline{97.37}\\
        \cmidrule(lr){1-5}
        DDG-FC & 0.11$\pm$0.04 / 0.91$\pm$0.03 / 96.69$\pm$1.12 & 0.42$\pm$0.05 / 0.75$\pm$0.01 / 96.09$\pm$1.86 & \textbf{0.97}$\pm$0.02 / \underline{0.58}$\pm$0.01 / 95.66$\pm$2.17 & 0.50 / 0.75 / 96.14\\
        MBDG-FC & 0.13$\pm$0.08 / 0.91$\pm$0.02 / 98.07$\pm$1.06 & 0.45$\pm$0.20 / 0.76$\pm$0.03 / 96.09$\pm$0.61 & 0.94$\pm$0.04 / 0.64$\pm$0.01 / 95.42$\pm$1.13 & 0.50 / 0.77 / 96.52\\
        EIIL & 0.15$\pm$0.08 / 0.94$\pm$0.03 / 81.00$\pm$0.31 & 0.26$\pm$0.06 / 0.98$\pm$0.01 / 82.67$\pm$2.44 & 0.62$\pm$0.16 / 0.98$\pm$0.01 / 71.68$\pm$0.51 & 0.34 / 0.97 / 78.45 \\
        FarconVAE & 0.11$\pm$0.08 / 0.94$\pm$0.01 / 94.40$\pm$2.35 & 0.43$\pm$0.21 / 0.77$\pm$0.03 / 82.61$\pm$1.90 & \textbf{0.97}$\pm$0.02 / 0.59$\pm$0.01 / 76.22$\pm$0.45 & 0.50 / 0.77 / 84.41 \\
        FCR & 0.12$\pm$0.05 / 0.90$\pm$0.01 / 79.00$\pm$1.00  & 0.56$\pm$0.22 / 0.75$\pm$0.03 / 77.65$\pm$2.06 & 0.82$\pm$0.15 / 0.67$\pm$0.02 / 70.35$\pm$1.36 & 0.50 / 0.77 / 75.67   \\
        FTCS & 0.10$\pm$0.02 / 0.93$\pm$0.01 / 83.16$\pm$2.45 & 0.52$\pm$0.32 / 0.69$\pm$0.02 / 78.65$\pm$1.23 & 0.80$\pm$0.09 / 0.66$\pm$0.02 / 72.51$\pm$1.02 & \underline{0.66} / 0.69 / 95.59   \\
        FATDM & 0.17$\pm$0.03 / \underline{0.86}$\pm$0.02 / 96.00$\pm$0.23 & \underline{0.92}$\pm$0.02 / \underline{0.64}$\pm$0.01 / 95.55$\pm$1.10 & 0.90$\pm$0.06 / \textbf{0.57}$\pm$0.03 / 95.23$\pm$0.55 & \underline{0.66} / \underline{0.67} / 95.59 \\
        \cmidrule(lr){1-5}
        \sysname{} & \textbf{0.23}$\pm$0.09 / \textbf{0.84}$\pm$0.01 / 96.15$\pm$0.50 & \textbf{0.98}$\pm$0.01 / \textbf{0.58}$\pm$0.01 / 97.94$\pm$0.30 & 0.92$\pm$0.05 / \textbf{0.57}$\pm$0.03 / \underline{96.19}$\pm$1.33 & \textbf{0.71} / \textbf{0.66} / 96.76\\
        \bottomrule
    \end{tabular}
\end{table*}

\begin{table*}[!t]
\tiny
    \centering
    \setlength\tabcolsep{1pt}
    \caption{Full performance on \texttt{FairFace}. (bold is the best; underline is the second best).}
    \label{tab:result-fairface-full}
    \begin{tabular}{l|c|c|c |c|c}
        \toprule
         & \multicolumn{5}{c}{$DP$ $\uparrow$ / $AUC_{fair}$ $\downarrow$ / \textbf{Accuracy} $\uparrow$} \\ 
        \cmidrule(lr){2-6}
        \multirow{-2}{*}{\textbf{Methods}} & (B, 0.91) & (E, 0.87) & (I, 0.58) & (W, 0.49) & (L, 0.48) \\
        \cmidrule(lr){1-1} \cmidrule(lr){2-2} \cmidrule(lr){3-3} \cmidrule(lr){4-4} \cmidrule(lr){5-5} \cmidrule(lr){6-6} 
        ColorJitter & 0.64$\pm$0.26 / 0.64$\pm$0.15 / 93.47$\pm$1.56  & 0.41$\pm$0.34 / 0.68$\pm$0.09 / 95.62$\pm$1.96 
        & 0.44$\pm$0.21 / 0.63$\pm$0.05 / 92.99$\pm$1.00 
        & 0.34$\pm$0.09 / 0.64$\pm$0.02 / 92.07$\pm$0.55 
        & 0.39$\pm$0.10 / 0.70$\pm$0.02 / 91.77$\pm$0.61 
        \\
        ERM & 0.67$\pm$0.17 / 0.58$\pm$0.02 / 91.89$\pm$1.10 & 0.43$\pm$0.21 / 0.64$\pm$0.02 / \underline{95.69}$\pm$2.19 
        & 0.50$\pm$0.19 / 0.59$\pm$0.03 / 93.28$\pm$1.61 
        & 0.39$\pm$0.09 / 0.61$\pm$0.01 / 92.82$\pm$0.38
        & \underline{0.57}$\pm$0.15 / 0.62$\pm$0.01 / 91.96$\pm$0.51 
        \\
        IRM & 0.63$\pm$0.12 / 0.58$\pm$0.01 / 93.39$\pm$1.03 & 0.32$\pm$0.23 / 0.63$\pm$0.03 / 95.12$\pm$0.49 & 0.45$\pm$0.06 / 0.59$\pm$0.02 / 92.01$\pm$1.13  & 0.32$\pm$0.19 / 0.66$\pm$0.01 / 90.54$\pm$1.56
        & 0.41$\pm$.021 / 0.63$\pm$0.05 / 92.06$\pm$1.89 
        \\
        GDRO & 0.71$\pm$0.16 / 0.57$\pm$0.02 / 89.81$\pm$1.10 & 0.46$\pm$0.16 / 0.61$\pm$0.02 / 95.26$\pm$1.53 & 0.50$\pm$0.14 / 0.59$\pm$0.01 / 93.27$\pm$1.27  & 0.48$\pm$0.09 / 0.60$\pm$0.01 / 92.50$\pm$0.38
        & 0.54$\pm$0.15 / 0.62$\pm$0.01 / 91.59$\pm$0.51 
        \\
        Mixup & 0.58$\pm$0.19 / 0.59$\pm$0.02 / 92.46$\pm$0.69 & 0.40$\pm$0.04 / 0.61$\pm$0.02 / 93.31$\pm$1.42 & 0.42$\pm$0.09 / 0.59$\pm$0.02 / \textbf{93.42}$\pm$2.43  & 0.43$\pm$0.19 / 0.61$\pm$0.01 / \underline{92.98}$\pm$0.03
        & 0.55$\pm$0.22 / 0.61$\pm$0.02 / 93.43$\pm$2.02 
        \\
        MLDG & 0.63$\pm$0.25 / 0.58$\pm$0.02 / 92.71$\pm$2.36 & 0.41$\pm$0.15 / 0.62$\pm$0.03 / 95.59$\pm$0.87 & 0.51$\pm$0.15 / 0.60$\pm$0.02 / 93.35$\pm$1.87 & 0.47$\pm$0.20 / \underline{0.59}$\pm$0.01 / 92.82$\pm$1.65
        &  0.53$\pm$0.18 / 0.62$\pm$0.03 / 92.99$\pm$0.86  
        \\
        CORAL & 0.69$\pm$0.19 / 0.58$\pm$0.01 / 92.09$\pm$2.03 & 0.34$\pm$0.24 / 0.64$\pm$0.01 / \textbf{95.91}$\pm$1.44 & \underline{0.53}$\pm$0.05 / 0.59$\pm$0.02 / 93.35$\pm$0.26  & 0.50$\pm$0.14 / 0.60$\pm$0.02 / 92.47$\pm$2.04 
        & 0.56$\pm$0.23 / \textbf{0.59}$\pm$0.03 / 92.62$\pm$1.11 
        \\
        MMD & 0.69$\pm$0.25 / 0.56$\pm$0.01 / \underline{93.87}$\pm$0.14 & 0.45$\pm$0.22 / \textbf{0.57}$\pm$0.02 / 94.68$\pm$0.20 & 0.27$\pm$0.18 / \textbf{0.57}$\pm$0.03 / 89.88$\pm$0.22  & 0.39$\pm$0.20 / 0.68$\pm$0.02 / 91.75$\pm$1.37 
        & 0.55$\pm$0.16 / 0.61$\pm$0.02 / 92.53$\pm$1.41 
        \\
        DANN & 0.46$\pm$0.07 / 0.61$\pm$0.02 / 91.80$\pm$0.64 & 0.53$\pm$0.18 / 0.85$\pm$0.03 / 91.54$\pm$2.24 & 0.38$\pm$0.18 / 0.63$\pm$0.01 / 90.09$\pm$0.60  & 0.11$\pm$0.09 / 0.66$\pm$0.01 / 86.80$\pm$1.18 
        & 0.39$\pm$0.21 / 0.67$\pm$0.01 / 90.82$\pm$2.44 
        \\
        CDANN & 0.62$\pm$0.24 / 0.59$\pm$0.03 / 91.22$\pm$0.33 & 0.43$\pm$0.10 / 0.66$\pm$0.02 / 94.75$\pm$2.23 & 0.43$\pm$0.18 / 0.61$\pm$0.01 / 92.41$\pm$1.68  & 0.35$\pm$0.17 / 0.67$\pm$0.02 / 90.19$\pm$0.60
        & 0.42$\pm$0.23 / 0.61$\pm$0.03 / 92.42$\pm$2.19 
        \\
        DDG & 0.60$\pm$0.20 / 0.59$\pm$0.02 / 91.76$\pm$1.03 & 0.36$\pm$0.15 / 0.63$\pm$0.02 / 95.52$\pm$2.35 & 0.49$\pm$0.17 / 0.59$\pm$0.01 / 92.35$\pm$2.04  & \underline{0.51}$\pm$0.07 / 0.60$\pm$0.01 / 91.34$\pm$0.80
        & 0.44$\pm$0.17 / 0.62$\pm$0.02 / 93.46$\pm$0.32 
        \\
        MBDG & 0.60$\pm$0.15 / 0.58$\pm$0.01 / 91.29$\pm$1.41 & 0.46$\pm$0.19 / 0.63$\pm$0.01 / 95.01$\pm$1.39 & 0.52$\pm$0.14 / \underline{0.58}$\pm$0.02 / 92.77$\pm$2.07  & 0.30$\pm$0.04 / 0.62$\pm$0.01 / 91.05$\pm$0.53 
        & 0.56$\pm$0.09 / 0.61$\pm$0.01 / \underline{93.49}$\pm$0.97 
        \\
        \cmidrule(lr){1-6}
        DDG-FC  & 0.61$\pm$0.06 / 0.58$\pm$0.03 / 92.27$\pm$1.65 & 0.39$\pm$0.18 / 0.64$\pm$0.03 / 95.51$\pm$2.36 & 0.45$\pm$0.17 / \underline{0.58}$\pm$0.03 / \underline{93.38}$\pm$0.52 & 0.48$\pm$0.15 / 0.62$\pm$0.02 / 92.45$\pm$1.55
        &  0.50$\pm$0.25 / 0.62$\pm$0.03 / 92.42$\pm$0.30 
        \\
        MBDG-FC  & 0.70$\pm$0.15 / 0.56$\pm$0.03 / 92.12$\pm$0.43 & 0.35$\pm$0.07 / \underline{0.60}$\pm$0.01 / 95.54$\pm$1.80 & \textbf{0.56}$\pm$0.07 / \textbf{0.57}$\pm$0.01 / 92.41$\pm$1.61  & 0.32$\pm$0.07 / 0.60$\pm$0.03 / 91.50$\pm$0.57
        & \underline{0.57}$\pm$0.23 / 0.62$\pm$0.02 / 91.89$\pm$0.81 
        \\
        EIIL & 0.88$\pm$0.07 / 0.59$\pm$0.05 / 84.75$\pm$2.16 & 0.69$\pm$0.12 / 0.71$\pm$0.01 / 92.86$\pm$1.70 & 0.47$\pm$0.08 / \textbf{0.57}$\pm$0.01 / 86.93$\pm$0.89  & 0.46$\pm$0.05 / 0.65$\pm$0.03 / 86.53$\pm$1.02
        & 0.49$\pm$0.07 / \textbf{0.59}$\pm$0.01 / 88.39$\pm$1.25
        \\
        FarconVAE & \underline{0.93}$\pm$0.03 / \textbf{0.54}$\pm$0.01 / 89.61$\pm$0.64 & 0.72$\pm$0.17 / 0.63$\pm$0.01 / 91.50$\pm$1.89 & 0.42$\pm$0.24 / \underline{0.58}$\pm$0.03 / 87.42$\pm$2.14  & \underline{0.51}$\pm$0.07 / 0.60$\pm$0.01 / 86.40$\pm$0.42
        & \textbf{0.58}$\pm$0.05 / \underline{0.60}$\pm$0.05 / 88.70$\pm$0.71 
        \\
        FCR & 0.81$\pm$0.05 / 0.59$\pm$0.02 / 79.66$\pm$0.25  & 0.60$\pm$0.09 / 0.69$\pm$0.02 / 89.22$\pm$1.30  & 0.40$\pm$0.06 / 0.62$\pm$0.02 / 79.15$\pm$0.56  & 0.39$\pm$0.06 / 0.63$\pm$0.02 / 82.33$\pm$0.89 & 0.38$\pm$0.12 / 0.66$\pm$0.02 / 85.22$\pm$2.33   \\

        FTCS & 0.75$\pm$0.10 / 0.60$\pm$0.02 / 80.00$\pm$0.20 & 0.66$\pm$0.18 / 0.65$\pm$0.01 / 88.11$\pm$1.09 & 0.49$\pm$0.05 / 0.65$\pm$0.01 / 82.15$\pm$0.64  & 0.40$\pm$0.06 / 0.60$\pm$0.02 / 79.66$\pm$1.05 & 0.42$\pm$0.23 / 0.65$\pm$0.03 / 79.64$\pm$1.00   \\
        
        FATDM & \underline{0.93}$\pm$0.03 / 0.57$\pm$0.02 / 92.20$\pm$0.36 & \underline{0.80}$\pm$0.02 / 0.65$\pm$0.02 / 92.89$\pm$1.00 & 0.52$\pm$0.10 / 0.60$\pm$0.01 / 92.22$\pm$1.60 & 0.46$\pm$0.05 / 0.63$\pm$0.01 / 92.56$\pm$0.31
        & 0.51$\pm$0.16 / 0.63$\pm$0.02 / 93.33$\pm$0.20 
        \\
        \cmidrule(lr){1-6}
        \sysname{} & \textbf{0.94}$\pm$0.05 / \underline{0.55}$\pm$0.02 / \textbf{93.91}$\pm$0.33 & \textbf{0.87}$\pm$0.05 / \underline{0.60}$\pm$0.01 / \textbf{95.91}$\pm$1.06 & 0.48$\pm$0.06 / \textbf{0.57}$\pm$0.02 / 92.55$\pm$1.45  & \textbf{0.52}$\pm$0.17 / \textbf{0.58}$\pm$0.03 / \textbf{93.02}$\pm$0.50 
        & \textbf{0.58}$\pm$0.15 / \textbf{0.59}$\pm$0.01 / \textbf{93.73}$\pm$0.26 
        \\
        \bottomrule
    \end{tabular}
\end{table*}
\begin{table*}[!t]
\tiny
    \centering
    \setlength\tabcolsep{4.8pt}
    \begin{tabular}{l|c|c|c}
        \toprule
         & \multicolumn{3}{c}{$DP$ $\uparrow$ / $AUC_{fair}$ $\downarrow$ / \textbf{Accuracy} $\uparrow$} \\ 
        \cmidrule(lr){2-4}
        \multirow{-2}{*}{\textbf{Methods}} & (M, 0.87) & (S, 0.39) & \textbf{Avg} \\
        \cmidrule(lr){1-1} \cmidrule(lr){2-2} \cmidrule(lr){3-3} \cmidrule(lr){4-4}
        ColorJitter & 0.36$\pm$0.12 / 0.65$\pm$0.05 / \underline{92.79}$\pm$1.22 & 0.35$\pm$0.20 / 0.69$\pm$0.06 / 91.89$\pm$1.02  
        & 0.42 / 0.66 / 92.94
        \\
        ERM & 0.34$\pm$0.08 / \underline{0.62}$\pm$0.01 / 92.51$\pm$1.45 & 0.68$\pm$0.14 / 0.59$\pm$0.03 / 93.48$\pm$0.94 
        & 0.51 / 0.61 / 93.08
        \\
        IRM & 0.34$\pm$0.11 / 0.65$\pm$0.02 / 92.47$\pm$2.42 & 0.55$\pm$0.23 / 0.59$\pm$0.01 / 91.81$\pm$0.66 
        & 0.43 / 0.62 / 92.48  
        \\
        GDRO & 0.45$\pm$0.14 / 0.63$\pm$0.02 / 91.75$\pm$1.11 & 0.72$\pm$0.14 / 0.59$\pm$0.01 / 93.65$\pm$0.67 
        & 0.55 / \underline{0.60} / 92.55 
        \\
        Mixup  & 0.31$\pm$0.11 / \underline{0.62}$\pm$0.02 / \textbf{93.52}$\pm$0.79 & 0.91$\pm$0.04 / 0.58$\pm$0.02 / 93.20$\pm$0.33 
        & 0.51 / \underline{0.60} / 93.19 
        \\
        MLDG & 0.35$\pm$0.20 / \underline{0.62}$\pm$0.01 / 92.45$\pm$0.07 & 0.71$\pm$0.22 / 0.57$\pm$0.01 / \textbf{93.85}$\pm$0.40 
        & 0.51 / \underline{0.60} / \underline{93.39} 
        \\
        CORAL & 0.43$\pm$0.08 / 0.63$\pm$0.01 / 92.23$\pm$0.06 & 0.74$\pm$0.10 / 0.58$\pm$0.01 / \underline{93.77}$\pm$1.99 
        & 0.54 / \underline{0.60} / 93.21 
        \\
        MMD & 0.48$\pm$0.25 / \underline{0.62}$\pm$0.02 / 91.07$\pm$2.00 & 0.66$\pm$0.18 / 0.59$\pm$0.03 / 92.58$\pm$1.63 
        & 0.50 / \underline{0.60} / 92.34 
        \\
        DANN & \textbf{0.65}$\pm$0.14 / 0.88$\pm$0.01 / 91.46$\pm$0.50 & 0.80$\pm$0.14 / 0.57$\pm$0.02 / 88.20$\pm$1.65 
        & 0.47 / 0.70 / 90.10  
        \\
        CDANN & 0.27$\pm$0.12 / 0.67$\pm$0.01 / 91.07$\pm$0.97 & 0.52$\pm$0.12 / 0.82$\pm$0.02 / 88.32$\pm$0.37 
        & 0.43 / 0.66 / 91.48  
        \\
        DDG & 0.37$\pm$0.14 / 0.64$\pm$0.01 / 91.36$\pm$0.65 & 0.63$\pm$0.22 / 0.58$\pm$0.01 / 93.40$\pm$0.37 
        & 0.49 / 0.61 / 92.74 
        \\
        MBDG  & 0.38$\pm$0.14 / 0.64$\pm$0.02 / 92.23$\pm$1.15 & 0.67$\pm$0.06 / \underline{0.56}$\pm$0.03 / 93.12$\pm$0.70 
        & 0.50 / \underline{0.60} / 92.71  
        \\
        \cmidrule(lr){1-4}
        DDG-FC  & 0.42$\pm$0.09 / 0.95$\pm$0.03 / 92.70$\pm$1.49 & 0.76$\pm$0.21 / 0.59$\pm$0.02 / \textbf{93.85}$\pm$1.79 
        & 0.52 / 0.61 / 93.23
        \\
        MBDG-FC  & 0.49$\pm$0.19 / 0.63$\pm$0.03 / 90.67$\pm$0.42 & 0.74$\pm$0.23 / 0.57$\pm$0.01 / 93.24$\pm$0.32 
        & 0.53 / \underline{0.60} / 92.48
        \\
        EIIL & 0.52$\pm$0.09 / 0.63$\pm$0.03 / 84.96$\pm$1.37 & \textbf{0.98}$\pm$0.01 / \textbf{0.55}$\pm$0.02 / 89.99$\pm$2.27 
        & 0.64 / 0.61 / 87.78 
        \\
        FarconVAE & 0.54$\pm$0.22 / \textbf{0.58}$\pm$0.02 / 85.62$\pm$1.49 & \underline{0.92}$\pm$0.06 / \underline{0.56}$\pm$0.10 / 90.00$\pm$0.05 
        & 0.66 / \textbf{0.58} / 88.46 
        \\
        FRC & 0.51$\pm$0.08 / 0.66$\pm$0.02 / 82.16$\pm$0.78  & 0.72$\pm$0.10 / 0.60$\pm$0.01 / 88.01$\pm$1.00  & 0.54 / 0.63 / 83.68  \\

        FTCS & 0.49$\pm$0.10 / 0.68$\pm$0.01 / 81.15$\pm$1.25  & 0.75$\pm$0.21 / 0.62$\pm$0.02 / 75.69$\pm$2.07  & 0.57 / 0.64 / 80.91  \\

        FATDM & \underline{0.55}$\pm$0.12 / 0.65$\pm$0.01 / 92.23$\pm$1.56 & \underline{0.92}$\pm$0.10 / 0.57$\pm$0.02 / 92.36$\pm$0.99 
        & \underline{0.67} / 0.61 / 92.54 
        \\
        \cmidrule(lr){1-4}
        \sysname{} & 0.54$\pm$0.08 / \underline{0.62}$\pm$0.02 / 92.61$\pm$1.84 & \textbf{0.98}$\pm$0.01 / \textbf{0.55}$\pm$0.01 / 92.26$\pm$2.48 
        & \textbf{0.70} / \textbf{0.58} / \textbf{93.42}
        \\
        \bottomrule
    \end{tabular}
\end{table*}

\begin{table*}[!t]
\tiny
    \centering
    \setlength\tabcolsep{4.8pt}
    \caption{Full performance on \texttt{YFCC100M-FDG}. (bold is the best; underline is the second best).}
    \label{tab:result-YFCC100M-full}
    \begin{tabular}{l|c|c|c|c}
        \toprule
         & \multicolumn{4}{c}{$DP$ $\uparrow$ / $AUC_{fair}$ $\downarrow$ / \textbf{Accuracy} $\uparrow$} \\ 
        \cmidrule(lr){2-5}
        \multirow{-2}{*}{\textbf{Methods}} & ($d_0$, 0.73) & ($d_1$, 0.84) & ($d_2$, 0.72) & \textbf{Avg}\\
        \cmidrule(lr){1-1} \cmidrule(lr){2-2} \cmidrule(lr){3-3} \cmidrule(lr){4-4} \cmidrule(lr){5-5}
        ColorJitter & 0.67$\pm$0.06 / 0.57$\pm$0.02 / 57.47$\pm$1.20 & 0.67$\pm$0.34 / 0.61$\pm$0.01 / 82.43$\pm$1.25 & 0.65$\pm$0.21 / 0.64$\pm$0.02 / 87.88$\pm$0.35 & 0.66 / 0.61 / 75.93\\
        ERM & 0.81$\pm$0.09 / 0.58$\pm$0.01 / 40.51$\pm$0.23 & 0.71$\pm$0.18 / 0.66$\pm$0.03 / 83.91$\pm$0.33 & 0.89$\pm$0.08 / 0.59$\pm$0.01 / 82.06$\pm$0.33 & 0.80 / 0.61 / 68.83 \\
        IRM & 0.76$\pm$0.10 / 0.58$\pm$0.02 / 50.51$\pm$2.44 & 0.87$\pm$0.08 / 0.60$\pm$0.02 / 73.26$\pm$0.03 & 0.70$\pm$0.24 / 0.57$\pm$0.02 / 82.78$\pm$2.19 & 0.78 / 0.58 / 68.85 \\
        GDRO & 0.80$\pm$0.05 / 0.59$\pm$0.01 / 53.43$\pm$2.29 & 0.73$\pm$0.22 / 0.60$\pm$0.01 / 87.56$\pm$2.20 & 0.79$\pm$0.13 / 0.65$\pm$0.02 / 83.10$\pm$0.64 & 0.78 / 0.62 / 74.70 \\
        Mixup & \underline{0.82}$\pm$0.07 / 0.57$\pm$0.03 / 61.15$\pm$0.28 & 0.79$\pm$0.14 / 0.63$\pm$0.03 / 78.63$\pm$0.97 & \underline{0.89}$\pm$0.05 / 0.60$\pm$0.01 / 85.18$\pm$0.80 & \underline{0.84} / 0.60 / 74.99 \\
        MLDG & 0.75$\pm$0.13 / 0.67$\pm$0.01 / 49.56$\pm$0.69 & 0.71$\pm$0.19 / 0.57$\pm$0.02 / 89.45$\pm$0.44 & 0.71$\pm$0.14 / 0.57$\pm$0.03 / 87.51$\pm$0.18 & 0.72 / 0.60 / 75.51 \\
        CORAL & 0.80$\pm$0.11 / 0.58$\pm$0.02 / 58.96$\pm$2.34 & 0.72$\pm$0.11 / 0.64$\pm$0.03 / 91.66$\pm$0.85 & 0.70$\pm$0.07 / 0.64$\pm$0.03 / 89.28$\pm$1.77 & 0.74 / 0.62 / 79.97 \\
        MMD & 0.79$\pm$0.11 / 0.59$\pm$0.02 / 61.51$\pm$1.79 & 0.71$\pm$0.15 / 0.64$\pm$0.03 / 91.15$\pm$2.33 & 0.79$\pm$0.17 / 0.60$\pm$0.01 / 86.69$\pm$0.19 & 0.76 / 0.61 / 79.87 \\
        DANN & 0.70$\pm$0.13 / 0.78$\pm$0.02 / 47.71$\pm$1.56 & 0.79$\pm$0.12 / \underline{0.53}$\pm$0.01 / 84.80$\pm$1.14 & 0.77$\pm$0.17 / 0.59$\pm$0.02 / 58.50$\pm$1.74 & 0.75 / 0.64 / 63.67 \\
        CDANN & 0.74$\pm$0.13 / 0.58$\pm$0.02 / 55.87$\pm$2.09 & 0.70$\pm$0.22 / 0.65$\pm$0.02 / 87.06$\pm$2.43 & 0.72$\pm$0.13 / 0.63$\pm$0.02 / 85.76$\pm$2.43 & 0.72 / 0.62 / 76.23 \\
        DDG & 0.81$\pm$0.14 / 0.57$\pm$0.03 / 60.08$\pm$1.08 & 0.74$\pm$0.12 / 0.66$\pm$0.03 / 92.53$\pm$0.91 & 0.71$\pm$0.21 / 0.59$\pm$0.03 / \textbf{95.02}$\pm$1.92 & 0.75 / 0.61 / 82.54 \\
        MBDG & 0.79$\pm$0.15 / 0.58$\pm$0.01 / 60.46$\pm$1.90 & 0.73$\pm$0.07 / 0.67$\pm$0.01 / 94.36$\pm$0.23 & 0.71$\pm$0.11 / 0.59$\pm$0.03 / \underline{93.48}$\pm$0.65 & 0.74 / 0.61 / \underline{82.77} \\
        \cmidrule(lr){1-5}
        DDG-FC & 0.76$\pm$0.06 / 0.58$\pm$0.03 / 59.96$\pm$2.36 & 0.83$\pm$0.06 / 0.58$\pm$0.01 / \textbf{96.80}$\pm$1.28 & 0.82$\pm$0.09 / 0.59$\pm$0.01 / 86.38$\pm$2.45 & 0.80 / 0.58 / 81.04 \\
        MBDG-FC & 0.80$\pm$0.13 / 0.58$\pm$0.01 / \underline{62.31}$\pm$0.13 & 0.72$\pm$0.09 / 0.63$\pm$0.01 / \underline{94.73}$\pm$2.09 & 0.80$\pm$0.07 / \textbf{0.53}$\pm$0.01 / 87.78$\pm$2.11 & 0.77 / 0.58 / 81.61 \\
        EIIL & \textbf{0.87}$\pm$0.11 / \underline{0.55}$\pm$0.02 / 56.74$\pm$0.60 & 0.76$\pm$0.05 / 0.54$\pm$0.03 / 68.99$\pm$0.91 & 0.87$\pm$0.06 / 0.78$\pm$0.03 / 72.19$\pm$0.75 & 0.83 / 0.62 / 65.98  \\
        FarconVAE & 0.67$\pm$0.06 / 0.61$\pm$0.03 / 51.21$\pm$0.61 & \underline{0.90}$\pm$0.06 / 0.59$\pm$0.01 / 72.40$\pm$2.13 & 0.85$\pm$0.12 / \underline{0.55}$\pm$0.01 / 74.20$\pm$2.46 & 0.81 / 0.58 / 65.93 \\
        FCR & 0.62$\pm$0.21 / 0.70$\pm$0.01 / 55.32$\pm$0.04 & 0.63$\pm$0.14 / 0.66$\pm$0.10 / 70.89$\pm$0.22 & 0.66$\pm$0.30 / 0.78$\pm$0.02 / 70.58$\pm$0.23 & 0.64 / 0.71 / 65.60 \\
        FTCS & 0.72$\pm$0.03 / 0.60$\pm$0.01 / 60.21$\pm$0.10 & 0.79$\pm$0.02 / 0.59$\pm$0.01 / 79.96$\pm$0.05 & 0.69$\pm$0.10 / 0.60$\pm$0.06 / 72.99$\pm$0.50 & 0.73 / 0.60 / 71.05 \\
        FATDM & 0.80$\pm$0.10 / \underline{0.55}$\pm$0.01 / 61.56$\pm$0.89 & 0.88$\pm$0.08 / 0.56$\pm$0.01 / 90.00$\pm$0.66 & 0.86$\pm$0.10 / 0.60$\pm$0.02 / 89.12$\pm$1.30 & \underline{0.84} / \underline{0.57} / 80.22 \\
        \cmidrule(lr){1-5}
        \sysname{} & \textbf{0.87}$\pm$0.09 / \textbf{0.53}$\pm$0.01 / \textbf{62.56}$\pm$2.25 & \textbf{0.94}$\pm$0.05 / \textbf{0.52}$\pm$0.01 / 93.36$\pm$1.70 & \textbf{0.93}$\pm$0.03 / \textbf{0.53}$\pm$0.02 / 93.43$\pm$0.73 & \textbf{0.92} / \textbf{0.53} / \textbf{83.12} \\
        \bottomrule
    \end{tabular}
\end{table*}

\begin{table*}[!t]
\tiny
    \centering
    \setlength\tabcolsep{0.2pt}
    \caption{Full performance on \texttt{NYSF}. (bold is the best; underline is the second best).}
    \label{tab:result-NY-full}
    \begin{tabular}{l|c|c|c |c|c|c}
        \toprule
         & \multicolumn{6}{c}{$DP$ $\uparrow$ / $AUC_{fair}$ $\downarrow$ / \textbf{Accuracy} $\uparrow$} \\ 
        \cmidrule(lr){2-7}
        \multirow{-2}{*}{\textbf{Methods}} & (R, 0.93) & (B, 0.85) & (M, 0.81) & (Q, 0.59) & (S, 0.62) & \textbf{Avg} \\
        \cmidrule(lr){1-1} \cmidrule(lr){2-2} \cmidrule(lr){3-3} \cmidrule(lr){4-4} \cmidrule(lr){5-5} \cmidrule(lr){6-6} \cmidrule(lr){7-7}
        ERM & 0.91$\pm$0.07 / 0.53$\pm$0.01 / 60.21$\pm$1.48 & 0.90$\pm$0.07 / 0.54$\pm$0.01 / 58.93$\pm$1.10  & 0.92$\pm$0.04 / 0.54$\pm$0.01 / 59.49$\pm$1.50 & 0.88$\pm$0.06 / 0.57$\pm$0.02 / 62.48$\pm$0.64 & 0.86$\pm$0.12 / 0.61$\pm$0.03 / 54.54$\pm$0.68 & 0.90 / 0.56 / 59.13
        \\
        IRM & \underline{0.98}$\pm$0.01 / \underline{0.52}$\pm$0.02 / 61.61$\pm$0.80 & \underline{0.94}$\pm$0.04 / \textbf{0.52}$\pm$0.02 / 56.89$\pm$0.73 & 0.92$\pm$0.02 / \underline{0.53}$\pm$0.03 / 59.64$\pm$2.33 & 0.87$\pm$0.06 / 0.54$\pm$0.01 / 55.81$\pm$1.74 & 0.89$\pm$0.07 / 0.54$\pm$0.03 / 57.00$\pm$2.01 & 0.92 / \underline{0.53} / 58.19
        \\
        GDRO & 0.81$\pm$0.18 / 0.56$\pm$0.02 / 58.73$\pm$2.23 & 0.89$\pm$0.07 / 0.55$\pm$0.03 / \underline{59.44}$\pm$1.66 & 0.87$\pm$0.08 / 0.55$\pm$0.02 / \textbf{62.57}$\pm$0.91 & 0.86$\pm$0.05 / 0.57$\pm$0.01 / \underline{62.92}$\pm$1.17 & 0.77$\pm$0.08 / 0.64$\pm$0.04 / 60.44$\pm$2.86 & 0.84 / 0.57 / \textbf{60.82}
        \\
        Mixup & 0.96$\pm$0.03 / 0.53$\pm$0.01 / \textbf{62.63}$\pm$1.84 & 0.90$\pm$0.06 / 0.54$\pm$0.04 / 58.96$\pm$2.89 & 0.92$\pm$0.04 / 0.54$\pm$0.03 / 58.29$\pm$0.80 & 0.93$\pm$0.04 / 0.53$\pm$0.01 / 61.34$\pm$1.60 & 0.84$\pm$0.08 / 0.61$\pm$0.02 / 53.07$\pm$3.13 & 0.91 / 0.55 / 58.86
        \\
        MLDG & 0.96$\pm$0.03 / \underline{0.52}$\pm$0.02 / 61.81$\pm$0.53 & 0.90$\pm$0.08 / 0.55$\pm$0.01 / 58.11$\pm$0.13  & 0.93$\pm$0.02 / \underline{0.53}$\pm$0.02 / 58.27$\pm$0.47 & 0.89$\pm$0.08 / 0.56$\pm$0.02 / 62.85$\pm$2.38 & 0.85$\pm$0.05 / 0.59$\pm$0.03 / 54.42$\pm$0.02 & 0.91 / 0.55 / 59.10 
        \\
        CORAL & 0.95$\pm$0.02 / \underline{0.52}$\pm$0.02 / \underline{62.17}$\pm$0.92 & 0.93$\pm$0.04 / 0.54$\pm$0.01 / 58.06$\pm$1.99 & \underline{0.95}$\pm$0.03 / \underline{0.53}$\pm$0.01 / 58.84$\pm$0.74 & 0.95$\pm$0.03 / 0.53$\pm$0.02 / 61.45$\pm$0.28 & 0.88$\pm$0.08 / 0.54$\pm$0.03 / 52.08$\pm$1.06 & \underline{0.93} / \underline{0.53} / 58.52
        \\
        MMD & 0.91$\pm$0.05 / 0.53$\pm$0.01 / 60.34$\pm$1.39 & 0.89$\pm$0.07 / 0.55$\pm$0.02 / 58.47$\pm$0.35 & 0.92$\pm$0.02 / 0.54$\pm$0.01 / 59.31$\pm$0.40 & 0.88$\pm$0.03 / 0.56$\pm$0.01 / 62.48$\pm$1.31 & 0.81$\pm$0.17 / 0.61$\pm$0.02 / 57.73$\pm$1.54 & 0.88 / 0.56 / 59.67
        \\
        DANN & 0.83$\pm$0.13 / \underline{0.52}$\pm$0.02 / 40.80$\pm$2.47 & \textbf{0.96}$\pm$0.02 / 0.55$\pm$0.03 / 54.55$\pm$0.17 & 0.88$\pm$0.04 / \textbf{0.52}$\pm$0.01 / 59.19$\pm$1.21 & 0.96$\pm$0.02 / 0.53$\pm$0.02 / 63.60$\pm$0.34 & 0.86$\pm$0.05 / 0.56$\pm$0.03 / 58.96$\pm$0.98 & 0.90 / 0.54 / 55.42
        \\
        CDANN & 0.95$\pm$0.03 / \underline{0.52}$\pm$0.01 / 57.61$\pm$0.68 & \underline{0.94}$\pm$0.03 / 0.54$\pm$0.02 / 56.97$\pm$1.29 & 0.87$\pm$0.09 / \textbf{0.52}$\pm$0.02 / 59.59$\pm$1.74 & \underline{0.97}$\pm$0.02 / 0.54$\pm$0.03 / \textbf{64.25}$\pm$1.25 & 0.74$\pm$0.16 / 0.60$\pm$0.01 / 57.73$\pm$1.89 & 0.89 / 0.54 / 59.23
        \\
        DDG & 0.92$\pm$0.03 / \underline{0.52}$\pm$0.01 / 56.52$\pm$0.71 & 0.92$\pm$0.04 / 0.54$\pm$0.04 / 58.21$\pm$1.40 & 0.92$\pm$0.07 / \underline{0.53}$\pm$0.02 / \underline{60.91}$\pm$2.47 & 0.89$\pm$0.07 / 0.55$\pm$0.01 / 56.68$\pm$0.87 & 0.84$\pm$0.07 / 0.58$\pm$0.03 / 54.91$\pm$1.33 & 0.90 / 0.54 / 57.44
        \\
        MBDG & 0.96$\pm$0.02 / \underline{0.52}$\pm$0.01 / 55.96$\pm$1.37 & 0.90$\pm$0.07 / 0.70$\pm$0.01 / 51.52$\pm$1.55 & \textbf{0.96}$\pm$0.02 / \underline{0.53}$\pm$0.03 / 58.74$\pm$2.46 & 0.96$\pm$0.03 / \underline{0.52}$\pm$0.01 / 60.73$\pm$1.56 & 0.90$\pm$0.04 / \textbf{0.52}$\pm$0.02 / 52.45$\pm$1.98 & \underline{0.93} / 0.56 / 55.88
        \\
        \cmidrule(lr){1-7}
        DDG-FC & 0.95$\pm$0.03 / \underline{0.52}$\pm$0.01 / 54.53$\pm$1.44 & 0.93$\pm$0.02 / \underline{0.53}$\pm$0.03 / 59.32$\pm$0.59 & 0.92$\pm$0.04 / 0.52$\pm$0.01 / 60.08$\pm$1.31 & 0.92$\pm$0.02 / 0.54$\pm$0.02 / 59.90$\pm$1.75 & 0.90$\pm$0.05 / 0.57$\pm$0.02 / 57.45$\pm$0.08 & 0.92 / \underline{0.53} / 58.26
        \\
        MBDG-FC & 0.96$\pm$0.02 / 0.55$\pm$0.02 / 55.93$\pm$1.98 & 0.91$\pm$0.07 / 0.54$\pm$0.03 / 55.50$\pm$0.55 & 0.90$\pm$0.06 / \underline{0.53}$\pm$0.02 / 57.37$\pm$2.39 & 0.94$\pm$0.04 / \underline{0.52}$\pm$0.01 / 61.04$\pm$2.31 & 0.91$\pm$0.06 / \underline{0.53}$\pm$0.03 / 52.57$\pm$0.92 & 0.92 / \underline{0.53} / 56.48
        \\
        EIIL & 0.95$\pm$0.02 / \underline{0.52}$\pm$0.02 / 58.28$\pm$3.23 & 0.92$\pm$0.03 / 0.54$\pm$0.02 / 56.76$\pm$3.87 & 0.83$\pm$0.11 / 0.54$\pm$0.02 / 59.47$\pm$1.69 & 0.84$\pm$0.12 / 0.55$\pm$0.02 / 52.18$\pm$0.26 & \underline{0.95}$\pm$0.03 / 0.59$\pm$0.02 / 55.74$\pm$0.12 & 0.90 / 0.54 / 56.49
        \\
        FarconVAE & 0.90$\pm$0.07 / 0.53$\pm$0.03 / 60.52$\pm$0.14 & 0.89$\pm$0.05 / 0.55$\pm$0.04 / \textbf{60.30}$\pm$0.64 & 0.82$\pm$0.07 / 0.56$\pm$0.01 / 60.31$\pm$0.40 & \underline{0.97}$\pm$0.02 / 0.56$\pm$0.03 / 61.30$\pm$1.14 & 0.86$\pm$0.10 / 0.58$\pm$0.02 / \underline{60.70}$\pm$1.48 & 0.89 / 0.56 / \underline{60.62}
        \\
        FCR & 0.91$\pm$0.05 / 0.55$\pm$0.02 / 56.21$\pm$1.15 & 0.88$\pm$0.05 / 0.56$\pm$0.02 / 55.12$\pm$1.02 & 0.77$\pm$0.05 / 0.55$\pm$0.01 / 58.15$\pm$0.95  & 0.85$\pm$0.15 / 0.59$\pm$0.02 / 59.15$\pm$1.11 & 0.82$\pm$0.09 / 0.55$\pm$0.02 / 51.45$\pm$1.05   & 0.83 / 0.56 / 56.02   \\

        FTCS & 0.92$\pm$0.06 / 0.54$\pm$0.01 / 58.23$\pm$1.15 & 0.84$\pm$0.02 / 0.56$\pm$0.03 / 49.23$\pm$1.21 & 0.80$\pm$0.07 / 0.56$\pm$0.02 / 55.48$\pm$0.48 & 0.89$\pm$0.15 / 0.58$\pm$0.00 / 57.15$\pm$0.51 & 0.84$\pm$0.05 / 0.56$\pm$0.01 / 56.15$\pm$0.15   & 0.86 / 0.56 / 55.25   \\
        
        FATDM & 0.93$\pm$0.05 / \underline{0.52}$\pm$0.01 / 59.32$\pm$1.00 & 0.86$\pm$0.05 / 0.58$\pm$0.02 / 59.01$\pm$0.32 & 0.85$\pm$0.08 / \underline{0.53}$\pm$0.02 / 60.45$\pm$0.87  & 0.85$\pm$0.05 / \underline{0.52}$\pm$0.01 / 60.35$\pm$0.44 & 0.88$\pm$0.03 / \textbf{0.52}$\pm$0.01 / 59.22$\pm$0.09 & 0.87 / \underline{0.53} / 59.67
        \\
        \cmidrule(lr){1-7}
        \sysname{} & \textbf{0.99}$\pm$0.00 / \textbf{0.50}$\pm$0.00 / 62.01$\pm$1.87 & \textbf{0.96}$\pm$0.01 / \textbf{0.52}$\pm$0.02 / 58.37$\pm$0.67 & 0.92$\pm$0.02 / \textbf{0.52}$\pm$0.02 / 59.49$\pm$1.93 & \textbf{0.99}$\pm$0.01 / \textbf{0.50}$\pm$0.00 / 59.11$\pm$0.94 & \textbf{0.98}$\pm$0.02 / \underline{0.53}$\pm$0.01 / \textbf{60.77}$\pm$0.23 & \textbf{0.97} / \textbf{0.51} / 59.95
        \\
        \bottomrule
    \end{tabular}
\end{table*}

\begin{table*}[!t]
\tiny
    \centering
    \setlength\tabcolsep{4.8pt}
    \caption{Ablation studies results on \texttt{ccMNIST}.}
    \label{tab:abs-ccmnist}
    \begin{tabular}{l|c|c|c|c}
        \toprule
         & \multicolumn{4}{c}{$DP$ $\uparrow$ / $AUC_{fair}$ $\downarrow$ / \textbf{Accuracy} $\uparrow$} \\ 
        \cmidrule(lr){2-5}
        \multirow{-2}{*}{\textbf{Methods}} & (R, 0.11) & (G, 0.43) & (B, 0.87) & \textbf{Avg}\\
        \cmidrule(lr){1-1} \cmidrule(lr){2-2} \cmidrule(lr){3-3} \cmidrule(lr){4-4} \cmidrule(lr){5-5}
        \sysnameabs{} & 0.23$\pm$0.05 / 0.98$\pm$0.01 / 94.89$\pm$1.72 & 0.11$\pm$0.06 / 0.92$\pm$0.02 / 98.19$\pm$1.39 & 0.42$\pm$0.06 / 0.72$\pm$0.03 / 95.28$\pm$0.22 & 0.25 / 0.87 / 96.12 \\
        \sysnameabss{} & 0.21$\pm$0.12 / 0.92$\pm$0.01 / 96.74$\pm$1.15 & 0.15$\pm$0.08 / 0.86$\pm$0.02 / 96.95$\pm$0.93 & 0.48$\pm$0.06 / 0.57$\pm$0.02 / 96.05$\pm$1.17 & 0.28 / 0.79 / 96.58 \\
        \sysnameabsss{} & 0.22$\pm$0.08 / 0.91$\pm$0.02 / 96.63$\pm$0.63 & 0.44$\pm$0.16 / 0.75$\pm$0.01 / 97.90$\pm$0.40 & 0.97$\pm$0.02 / 0.61$\pm$0.02 / 96.01$\pm$0.20 & 0.54 / 0.76 / 96.85 \\
        \bottomrule
    \end{tabular}
\end{table*}

\begin{table*}[!t]
\tiny
    \centering
    \setlength\tabcolsep{4.8pt}
    \caption{Ablation studies results on \texttt{FairFace}.}
    \label{tab:abs-fairface}
    \begin{tabular}{l|c|c|c}
        \toprule
         & \multicolumn{3}{c}{$DP$ $\uparrow$ / $AUC_{fair}$ $\downarrow$ / \textbf{Accuracy} $\uparrow$} \\ 
        \cmidrule(lr){2-4}
        \multirow{-2}{*}{\textbf{Methods}} & (B, 0.91) & (E, 0.87) & (I, 0.58) \\
        \cmidrule(lr){1-1} \cmidrule(lr){2-2} \cmidrule(lr){3-3} \cmidrule(lr){4-4} 
        \sysnameabs{} & 0.68$\pm$0.18 / 0.57$\pm$0.02 / 93.07$\pm$0.68 & 0.43$\pm$0.20 / 0.60$\pm$0.03 / 95.55$\pm$2.09 & 0.37$\pm$0.09 / 0.59$\pm$0.03 / 92.26$\pm$0.37  \\
        \sysnameabss{} & 0.83$\pm$0.08 / 0.56$\pm$0.01 / 92.81$\pm$0.81 & 0.50$\pm$0.22 / 0.56$\pm$0.01 / 95.12$\pm$0.73 & 0.42$\pm$0.17 / 0.59$\pm$0.02 / 92.34$\pm$0.14  \\
        \sysnameabsss{} & 0.59$\pm$0.16 / 0.58$\pm$0.01 / 92.92$\pm$1.35 & 0.36$\pm$0.08 / 0.62$\pm$0.03 / 95.55$\pm$1.84 & 0.42$\pm$0.20 / 0.62$\pm$0.02 / 93.35$\pm$0.83 \\
        \bottomrule
    \end{tabular}
\end{table*}
\begin{table*}[!t]
\tiny
    \centering
    \setlength\tabcolsep{4.8pt}
    \begin{tabular}{l|c|c|c}
        \toprule
         & \multicolumn{3}{c}{$DP$ $\uparrow$ / $AUC_{fair}$ $\downarrow$ / \textbf{Accuracy} $\uparrow$} \\ 
        \cmidrule(lr){2-4}
        \multirow{-2}{*}{\textbf{Methods}} & (M, 0.87) & (S, 0.39) & (W, 0.49) \\
        \cmidrule(lr){1-1} \cmidrule(lr){2-2} \cmidrule(lr){3-3} \cmidrule(lr){4-4} 
        \sysnameabs{} & 0.49$\pm$0.13 / 0.62$\pm$0.03 / 92.61$\pm$2.32 & 0.69$\pm$0.22 / 0.56$\pm$0.01 / 93.28$\pm$2.31 & 0.35$\pm$0.26 / 0.58$\pm$0.01 / 92.18$\pm$0.46  \\
        \sysnameabss{} & 0.39$\pm$0.07 / 0.68$\pm$0.01 / 91.46$\pm$2.05 & 0.92$\pm$0.06 / 0.56$\pm$0.01 / 87.87$\pm$1.25 & 0.52$\pm$0.23 / 0.59$\pm$0.01 / 90.78$\pm$0.31  \\
        \sysnameabsss{} & 0.38$\pm$0.15 / 0.72$\pm$0.03 / 92.27$\pm$0.02 & 0.42$\pm$0.16 / 0.67$\pm$0.03 / 92.17$\pm$0.99 & 0.34$\pm$0.08 / 0.72$\pm$0.03 / 91.88$\pm$0.67  \\
        \bottomrule
    \end{tabular}
\end{table*}
\begin{table*}[!t]
\tiny
    \centering
    \setlength\tabcolsep{4.8pt}
    \begin{tabular}{l|c|c}
        \toprule
         & \multicolumn{2}{c}{$DP$ $\uparrow$ / $AUC_{fair}$ $\downarrow$ / \textbf{Accuracy} $\uparrow$} \\ 
        \cmidrule(lr){2-3}
        \multirow{-2}{*}{\textbf{Methods}} & (L, 0.48) & \textbf{Avg} \\
        \cmidrule(lr){1-1} \cmidrule(lr){2-2} \cmidrule(lr){3-3}
        \sysnameabs{} & 0.47$\pm$0.07 / 0.63$\pm$0.01 / 92.62$\pm$0.93 & 0.49 / 0.59 / 93.08 \\
        \sysnameabss{} & 0.53$\pm$0.03 / 0.59$\pm$0.01 / 91.19$\pm$0.57 & 0.58 / 0.59 / 91.65 \\
        \sysnameabsss{}& 0.40$\pm$0.07 / 0.70$\pm$0.02 / 92.96$\pm$0.85 & 0.42 / 0.66 / 93.01 \\
        \bottomrule
    \end{tabular}
\end{table*}

\begin{table*}[!t]
\tiny
    \centering
    \setlength\tabcolsep{4.8pt}
    \caption{Ablation studies results on \texttt{YFCC100M-FDG}.}
    \label{tab:abs-yfcc}
    \begin{tabular}{l|c|c|c|c}
        \toprule
         & \multicolumn{4}{c}{$DP$ $\uparrow$ / $AUC_{fair}$ $\downarrow$ / \textbf{Accuracy} $\uparrow$} \\ 
        \cmidrule(lr){2-5}
        \multirow{-2}{*}{\textbf{Methods}} & ($d_0$, 0.73) & ($d_1$, 0.84) & ($d_2$, 0.72) & \textbf{Avg}\\
        \cmidrule(lr){1-1} \cmidrule(lr){2-2} \cmidrule(lr){3-3} \cmidrule(lr){4-4} \cmidrule(lr){5-5}
        \sysnameabs{} & 0.69$\pm$0.13 / 0.57$\pm$0.02 / 43.09$\pm$1.45 & 0.83$\pm$0.08 / 0.63$\pm$0.02 / 89.68$\pm$0.60 & 0.89$\pm$0.05 / 0.54$\pm$0.03 / 87.70$\pm$1.69 & 0.80 / 0.58 / 73.49 \\
        \sysnameabss{} & 0.82$\pm$0.12 / 0.56$\pm$0.03 / 47.21$\pm$1.17 & 0.83$\pm$0.05 / 0.63$\pm$0.01 / 73.10$\pm$0.26 & 0.82$\pm$0.08 / 0.53$\pm$0.02 / 72.95$\pm$2.25 & 0.82 / 0.57 / 64.42  \\
        \sysnameabsss{} & 0.72$\pm$0.17 / 0.69$\pm$0.03 / 54.24$\pm$1.75 & 0.92$\pm$0.02 / 0.64$\pm$0.03 / 94.35$\pm$2.35 & 0.92$\pm$0.07 / 0.64$\pm$0.03 / 93.20$\pm$2.17 & 0.86 / 0.66 / 80.59  \\
        \bottomrule
    \end{tabular}
\end{table*}

\begin{table*}[!t]
\tiny
    \centering
    \setlength\tabcolsep{0.2pt}
    \caption{Ablation studies results on \texttt{NYSF}.}
    \label{tab:abs-nysf}
    \begin{tabular}{l|c|c|c |c|c|c}
        \toprule
         & \multicolumn{6}{c}{$DP$ $\uparrow$ / $AUC_{fair}$ $\downarrow$ / \textbf{Accuracy} $\uparrow$} \\ 
        \cmidrule(lr){2-7}
        \multirow{-2}{*}{\textbf{Methods}} & (R, 0.93) & (B, 0.85) & (M, 0.81) & (Q, 0.59) & (S, 0.62) & \textbf{Avg}\\
        \cmidrule(lr){1-1} \cmidrule(lr){2-2} \cmidrule(lr){3-3} \cmidrule(lr){4-4} \cmidrule(lr){5-5} \cmidrule(lr){6-6} \cmidrule(lr){7-7} 
        \sysnameabs{} & 0.95$\pm$0.02 / 0.52$\pm$0.01 / 55.78$\pm$1.01 & 0.97$\pm$0.01 / 0.51$\pm$0.01 / 55.30$\pm$1.08 & 0.95$\pm$0.03 / 0.53$\pm$0.01 / 58.29$\pm$0.80 & 0.92$\pm$0.06 / 0.54$\pm$0.02 / 57.61$\pm$1.30 & 0.90$\pm$0.02 / 0.59$\pm$0.02 / 52.82$\pm$1.20 & 0.94 / 0.53 / 55.96
        \\
        \sysnameabss{} & 0.95$\pm$0.03 / 0.52$\pm$0.01 / 61.36$\pm$0.42 & 0.91$\pm$0.06 / 0.54$\pm$0.01 / 57.67$\pm$0.82 & 0.89$\pm$0.05 / 0.55$\pm$0.01 / 60.68$\pm$0.31 & 0.97$\pm$0.02 / 0.52$\pm$0.01 / 59.33$\pm$0.17 & 0.87$\pm$0.11 / 0.57$\pm$0.01 / 55.40$\pm$0.73 & 0.92 / 0.54 / 58.89
        \\
        \sysnameabsss{} & 0.95$\pm$0.02 / 0.52$\pm$0.02 / 63.72$\pm$0.37 & 0.87$\pm$0.09 / 0.55$\pm$0.01 / 58.86$\pm$0.68 & 0.89$\pm$0.08 / 0.54$\pm$0.01 / 60.61$\pm$0.59 & 0.83$\pm$0.08 / 0.57$\pm$0.01 / 64.17$\pm$0.35 & 0.89$\pm$0.06 / 0.58$\pm$0.02 / 56.51$\pm$0.84 & 0.89 / 0.55 / 60.77
        \\
        \bottomrule
    \end{tabular}
\end{table*}


\textbf{Sensitive analysis on slacks $\gamma_1$ and $\gamma_2$}
We show additional experiment results by choosing different $\gamma_1$ and $\gamma_2$ of \cref{alg:algor-s2} in \cref{tab:slacks}. We observe that (1) by only increasing $\gamma_2$, the model towards giving unfair outcomes but higher accuracy; (2) by only increasing $\gamma_1$, performance on both model fairness and accuracy decreases. This may be due to the failure of disentanglement of factors.

\begin{table*}[!t]
\tiny
    \centering
    \caption{Sensitiveness of slacks.}
    \label{tab:slacks}
    \begin{tabular}{l|c|c|c|c}
    \toprule
     & \multicolumn{4}{c}{$DP$ $\uparrow$ / $AUC_{fair}$ $\downarrow$ / \textbf{Accuracy} $\uparrow$} \\ 
     \cmidrule(lr){2-5}
     & \texttt{ccMNIST} & \texttt{FairFace} & \texttt{YFCC100M-FDG} & \texttt{NYSF} \\
     \cmidrule(lr){1-1} \cmidrule(lr){2-2} \cmidrule(lr){3-3} \cmidrule(lr){4-4} \cmidrule(lr){5-5}
     $\gamma_1=0.025, \gamma_2=0.25$ & 0.47 / 0.79 / 97.07 & 0.53 / 0.60 / 93.99 & 0.88 / 0.55 / 88.69 & 0.86 / 0.56 / 61.71 \\
     $\gamma_1=0.25, \gamma_2=0.025$ & 0.66 / 0.75 / 88.54 & 0.62 / 0.58 / 93.06 & 0.91 / 0.54 / 81.49 & 0.86 / 0.57 / 58.03\\
     $\gamma_1=0.025, \gamma_2=0.025$ & 0.71 / 0.66 / 96.97 & 0.70 / 0.58 / 93.42 & 0.92 / 0.53 / 83.12 & 0.97 / 0.51 / 59.95\\
    \bottomrule
    \end{tabular}
\end{table*}

%% file: app_limitations.tex
In \cref{sec:experiments,app:addtional-results}, we empirically demonstrate the effectiveness of the proposed \sysname{}, wherein our method is developed based on assumptions. We assume (1) data instances can be encoded into three latent factors, (2) such factors are independent of each other, and (3) each domain shares the same semantic space. 
\sysname{} may not work well when data are generated with more than three factors or such factors are correlated to each other. 
Studies on causal learning could be a solution to address such limitations.
Moreover, our model relies on domain augmentation. While the results demonstrate its effectiveness, it might not perform optimally when semantic spaces do not completely overlap across domains. In such scenarios, a preferable approach would involve initially augmenting data by minimizing semantic gaps for each class across training domains, followed by conducting domain augmentations.